\documentclass[12pt]{article}
\usepackage[authoryear]{natbib}
\usepackage{amsfonts}
\usepackage{amsmath,amssymb,amsthm,enumerate,epsfig,graphicx}
\usepackage{ifthen,latexsym,syntonly}
\usepackage{appendix}
\usepackage{rotating}
\usepackage{lscape}
\usepackage[export]{adjustbox}
\usepackage{booktabs}
\usepackage{multirow}
\usepackage[table,xcdraw]{xcolor}
\usepackage{parskip}
\usepackage[colorlinks, citecolor=blue, urlcolor=blue, pdfborder={0 0 1}, citebordercolor={1 1 1}]{hyperref}
\usepackage{ushort}
\usepackage{natbib}
\usepackage{float}
\usepackage{hyperref}
\usepackage{comment}
\usepackage{nameref}
\usepackage{setspace}
\usepackage{zref-xr}
\usepackage[normalem]{ulem}
\usepackage{lscape}
\usepackage[strict]{changepage}
\usepackage{bm}
\usepackage[small,compact]{titlesec}
\usepackage{subfigure}

\useunder{\uline}{\ul}{}
\zxrsetup{toltxlabel}
\zexternaldocument*{supp}
\newcommand{\argmin}{\arg\!\min}

\allowdisplaybreaks

\theoremstyle{definition}
\newtheorem{remark}{Remark}[section]

\numberwithin{equation}{section}

\DeclareMathOperator{\var}{Var}

\theoremstyle{plain}
 \newtheorem{prop}{Proposition}[section]
\newtheorem{thm}{Theorem}[section]
\newtheorem{assum}{Assumption}[section]
 \newtheorem{lem}{Lemma}[section]

\begin{document}
	
	\title{Inference on Time Series Nonparametric Conditional Moment
		Restrictions Using General  Sieves}
	\author{ Xiaohong Chen\thanks{%
			Cowles Foundation for Research in Economics, Yale University, New Haven, CT
			06520, USA. \texttt{xiaohong.chen@yale.edu}. } \and Yuan Liao\thanks{%
			Department of Economics, Rutgers University, New Brunswick, NJ 08901, USA.
			\texttt{yuan.liao@rutgers.edu}} \and Weichen Wang\thanks{%
			Faculty of Business and Economics, The University of Hong Kong, Pokfulam
			Road, Hong Kong \texttt{weichenw@hku.hk}.} }
	\date{First draft: September 2020, revised \today }
	\maketitle
	
	\begin{abstract}
		General  nonlinear sieve  learnings are  classes of nonlinear sieves that can approximate
		nonlinear functions of high dimensional variables much more flexibly than
		various linear sieves (or series).  	This paper considers general nonlinear sieve quasi-likelihood		ratio (GN-QLR) based inference on expectation functionals of time series data, where the functionals of interest are based on some nonparametric function that satisfy  conditional moment restrictions and are learned using multilayer neural networks. While the asymptotic normality of the estimated functionals depends on some unknown Riesz representer of the functional space, we show that the optimally weighted GN-QLR statistic is  asymptotically 		Chi-square distributed, regardless whether the expectation functional is regular (root-$n$ estimable) or not. This holds when the data  are  weakly dependent beta-mixing condition.
		We apply our method to the off-policy evaluation in reinforcement learning, by formulating the Bellman equation into the conditional moment restriction framework, so that we can make inference about the state-specific value functional using the proposed GN-QLR method  with time series data. In addition, estimating the averaged partial means and averaged partial derivatives of nonparametric instrumental variables and quantile IV models are also presented as leading examples. Finally, a Monte Carlo study shows the finite sample performance of the procedure


	\end{abstract}
	
	
	\onehalfspacing
	
	\newpage
	
	\section{Introduction}

	Consider a conditional moment restriction model
	\begin{equation}  \label{eq2.1a}
	\mathbb{E}[\rho(Y_{t+1},\alpha_0)|\sigma_t(\mathcal{X})]=0\,,
	\end{equation}
	where $\rho$ is a scalar residual function; $\alpha_0=(\theta_0,h_0)$ contains a finite dimensional parameter $\theta_0$ and an
	infinite dimensional parameter $h_0$, which may depend on some endogenous
	variables $W_t$. The conditioning filtration $\sigma_t(\mathcal{X})$ is the
	sigma-algebra generated by variables $\{\mathcal{X}_s: s\leq t\}$, where $%
	\mathcal{X}_s$ is a vector of multivariate (finite dimensional) exogenous
	variables, including all relevant lagged variables of $Y_t$ and other
	instrumental variables. The model therefore allows for endogenous variables and weakly dependent data.

	This paper considers optimal estimation and inference for linear functionals $\phi(\alpha_0)$ of the
	infinite dimension. The functional may be either known or not. When it is unknown, it is assumed to take the form
	$$
	\phi(\alpha_0)=\mathbb El(h_0(W_t))\,,
	$$
	where $l$ is a known linear function and $h_0(W_t)$ is the nonparametric function on endogenous variable. We use 
	general nonlinear sieve learning spaces, whose complexity grows with the sample size, to estimate the infinite dimensional parameter, such as multi-layer neural networks and Gaussian radial basis. The motivation of using general nonlinear sieve learning space, besides being adaptive to high dimensional covariates, is that they  allow unbounded supports of the covariates. This is particularly desirable for models of dependent time series  data, such as nonlinear autoregressive models. 

	We formally establish inferential
	theories of these functionals learned using the general nonlinear sieve learning space, and conduct
	inference using quasi-likelihood ratio (QLR) statistics based on the
	optimally weighted  minimum distance. Of particular interest is the estimation of an expectation functional,
	such as averaged partial means, weighted average derivatives  and
	averaged squared partial derivatives, of a nonparametric conditional moment
	restriction via  nonlinear sieve learning sieves. 
	An important insight from our main theory  is that the asymptotic distribution  does not depend on the actual choice of the learning space, but is only determined by the functional  and the loss function.  Therefore, estimators produced by   either   deep neural networks,  Gaussian radial basis, or other nonlinear sieve learning basis, have the same asymptotic distribution.


		In general,  machine learning inference  often  relies on sample splitting/ cross-fitting, which does not work well in the time series setting. We propose a new time series efficient inference based on the  optimal quasi-likelihood ratio test,  without requiring cross-fitting. It is shown that  the optimally weighted QLR statistic,
	based on the general nonlinear sieve learning of $h_{0}()$, is asymptotically
	chi-square distributed regardless of whether the information bound for the
	expectation functional is singular or not, which can be used to construct confidence sets  
	without the need to compute standard errors. We present a   Monte Carlo
	study to illustrate finite sample performance of our inference procedure.

	Depending on the specific applications, our model may involve Fredholm integral equation of either the first kind (NPIV and NPQIV)  or the second kind (Bellman equations). In the former case, it is well known that estimating $h_0$ is an ill-posed problem and the rate of convergence might be slow. In the latter case, the problem can be  well-posed.  As one of the leading examples of the Fredholm integral equation of second kind, we   show that our framework implies a natural neural network-based inference in the context of \textit{Reinforcement Learning} (RL), a popular learning device behind  many successful applications of  artificial intelligence such as    AlphaGo, video games, robotics, and autonomous driving \citep{sutton2018reinforcement, silver2016mastering, vinyals2019alphastar, shalev2016safe}. Due to the dynamics of the RL model,  theoretical analysis of  reinforcement learning naturally requires  to explicitly allow time series dependency among the observed data. Earlier theoretical studies  focused  on the settings where the value function are approximated by linear functions.   More recent developments on nonlinear learning space include  \cite{farahmand2016regularized, geist2019theory,fan2020theoretical, duan2021optimal, long20212, chen2022well,shi2020statistical}, among others.  Our innovation lies in making inference 
	about the functionals (such as the value functional for specific states) of the  $Q$-function using general nonlinear sieve learning spaces.	While the reinforcement learning is based on the well known \textit{Bellman equation}, it can be formulated as the conditional moment restriction model with time series data. 	Therefore, one can apply the GN-QLR inference to estimating the state-specific value function   in the setting of the off-policy evaluation.   These applications are potentially useful for dynamic causal inference.

	
	\bigskip
	
	
In the i.i.d. case, 	existing theoretical works on neural networks have focused on deriving approximation
	theories and optimal rates of convergence for estimations.   Theoretically, deep learning has been shown to be able to approximate a broad class of highly nonlinear functions, see, e.g. \cite{mhaskar2016learning, rolnick2017power, lin2017does, shen2021neural,hsu2021approximation, schmidt2020nonparametric}.  \cite{yang1999information} obtained the minimax  $L_2$- rate of convergence for neural network models.
	Recently, 
\cite{chen2021efficient} considered NN efficient estimation of the (weighted)
	average derivatives in a NPIV model for i.i.d. data, and presented consistent
	variance estimation. In contrast, using a  general theory of Riesz
	representations, we derive the asymptotic distribution of the finite
	dimensional parameter $\theta_0$ and functionals of the infinite dimensional parameter $h_0$
	that is learned from the general learning space. The uncertainty of the general nonlinear sieve learning estimator plays
	an essential role in the asymptotic distributions.   \cite{chernozhukov2018double,chernozhukov2018generic, chernozhukov2018automatic} proposed double machine learning and debias methods to achieve valid inference; 
	\cite{dikkala2020minimax} studied  a minimax criterion  function to study the unknown functional approximated by neural networks for NPIV models. 
	In addition, the Riesz representation is playing a central role in our inferential theory. See \cite{newey1994asymptotic, shen1997methods,chen1998sieve,chernozhukov2020adversarial} for related approaches.

	In the time series setting, the   neural networks have been  applied to  economic demand estimations as in \cite{chen2009land}, and is widely applicable in financial asset pricing such as  \cite
	{guijarro2021deep,gu2020empirical,bali2021different}. These papers 
 	approximate unknown
	functions   by neural networks, but without rigorous
	theoretical justifications. All these models can be formulated as an inference problem for conditional moments.   

	The rest of the paper is organized as follows. Section \ref{sec:model} first
	introduces the model, the general nonlinear sieve space, the estimation and inference
	procedures. Section \ref{sec:rate} establishes the convergence rate of the
	nonlinear sieve estimator for the unknown function satisfying the conditional moment
	restrictions with weakly dependent data. Section \ref{sec:normality}
	provides the limiting distribution of the estimator for functionals that can be regular or
	irregular. Section \ref{sec:qlr} shows that the GN-QLR
	statistics is asymptotically Chi-square distributed for both the regular and
	irregular functionals for time series. In Section \ref%
	{sec:examples} we apply our approach to the estimation of the value function of RL and the weighted average derivative of NPIV and
	NPQIV as leading examples. Section \ref{sec:MC} contains simulation studies
	and Section \ref{sec:conclusion} briefly concludes.

	\section{The model}
	
	\label{sec:model}

	\subsection{The General sieve learning space}
		
  This paper
	studies inference with the general nonlinear sieve learning space. The unknown function is estimated on a learning space, denoted by $\mathcal H_n$, is a general approximation space that consists of either linear or nonlinear sieves, provided that the function of interest can be approximated well by the learning space.
	
	The popular feedforward neural network (NN) is one of the leading examples that fits into this context. Many theoretical studies have shown that NN can well approximate a broad class of functions and achieves nice statistical properties. The multilayer feedforward NN composites functions taking the form:
$$
h(x)= \theta_{J+1}h_{J}(x), \quad \cdots \quad h_{j}(x)= \sigma(\theta_{j} h_{j-1}(x)), \quad \cdots \quad,h_{0}(x)=x
$$
where the parameters $\theta=(\theta_1,\cdots,\theta_J)$ with $\theta_j\in\mathbb R^{d_j\times d_{j-1}}$ , $h_j(x) \in \mathbb R^{d_j}$, and $\sigma: \mathbb R^{d_j}\rightarrow \mathbb R^{d_{j}}$ is a elementwise nonlinear activation function, usually the same across components and layers. One of the popularly used activation functions is known as ReLU, defined as
 $
 \sigma(x)= \max(0, x). 
 $
The number of \textit{neurons} being used in layer $j$, denoted by $d_j$, is called the width of that layer.

	We could also use other nonlinear approximation learning spaces, which uses nonlinear combinations of inputs and neurons. One such example is the space spanned by Gaussian radial bases, which is a multilayer compositions of functions of the form:
$$
h(x)=\alpha_0+\sum_{j=1}^J\alpha_jG(\sigma_j^{-1}\|x-\gamma_j\|),\quad \alpha_0,\alpha_j, \gamma_j\in\mathbb R, \sigma_j>0,
$$
	 where $G$ is the standard normal density function. A key feature  is that here inputs and neurons (e.g., a vector of $x$) are ``nonlinearly combined" as $\|x-\gamma_j\|$, while they are linearly combined as indices $\theta_jx$ in the ordinary neural networks. 	 Additional examples of nonlinear sieves include spline   and wavelet sieves. They are very flexible and enjoy better approximation properties than
linear sieves.

One of the key motivations of using general nonlinear sieve learning
space, besides being adaptive to high dimensional covariates, is that it allows unbounded supports of input covariates. This is particularly desirable for time series models dependent data, such as
nonlinear autoregressive models.

	\subsection{Semiparametric  learning   }
	
	We shall assume a finite-order Markov property: for some known and fixed
	integer $r\geq 1$, let $X_t:=(\mathcal{X}_t,...,\mathcal{X}_{t-r})$ for all $%
	t=1,...,n$. define
	\begin{eqnarray*}
		m(X_t,\alpha)&=&\mathbb E[\rho(Y_{t+1},\alpha)|\sigma_t(\mathcal{X})],\cr %
		\Sigma(X_t)&=&\var(\rho(Y_{t+1},\alpha_0)|\sigma_t(\mathcal{X})),
	\end{eqnarray*}
	where we assume that $\mathbb{E}[\rho(Y_{t+1},\alpha)|\sigma_t(\mathcal{X})]$
	and $\var(\rho(Y_{t+1},\alpha_0)|\sigma_t(\mathcal{X}))$ only depend on $(%
	\mathcal{X}_t,...,\mathcal{X}_{t-r})$ for all $\alpha$. The model is then
	equivalent to $Q(\alpha_0)=0$ where
	\begin{equation*}
	Q(\alpha)=\mathbb{E }m(X_t,\alpha) ^2 \Sigma(X_t)^{-1} .
	\end{equation*}
	Here we use the optimal weighting function $\Sigma(X_t)$. Suppose there are
	nonparametric estimators $\widehat m(X,\alpha)$ and $\widehat\Sigma(X_t)$
	for $m(X_t.,\alpha)$ and $\Sigma(X_t)$, we then define the sample criterion
	function
	\begin{equation*}
	Q_n(\alpha)=\frac{1}{n}\sum_{t=1}^n\widehat m(X_t,\alpha)^2
	\widehat\Sigma(X_t)^{-1} .
	\end{equation*}
	The estimated optimal weighting matrix is needed for the quasi-likelihood
	inference. In practice, one can start with the identity weighting function
	to obtain an initial estimator for $\alpha_0$, use it to estimate $\Sigma(X_t)$%
	, then update the estimator using the estimated optimal weighting matrix.
	
	We focus on the general nonlinear sieve learning approximation to the true nonparametric function,
	and restrict to the following estimation space:
	\begin{equation*}
	\mathcal{A}_n:=\Theta\times \mathcal{H}_n.
	\end{equation*}
 Here $\Theta$ is a compact set as the
	parameter space for $\theta_0$ but  not necessarily for $\mathcal{H}_n$. In addition, let $P_{en}(h)$ denote some
	functional penalty for the infinite dimensional parameter. We then
	define the estimator $\widehat\alpha=(\widehat\theta,\widehat h)\in \mathcal{%
		A}_n$ as an approximate minimizer of the penalized loss function restricted
	to the general nonlinear sieve learning space:
	\begin{equation*}
	Q_n( \widehat\alpha)+\lambda_n P_{en}(\widehat h) \leq \inf_{\alpha\in%
		\mathcal{A}_n} Q_n(\alpha)+\lambda_n P_{en}(h)+ o_P(n^{-1}).
	\end{equation*}
	
	The tuning parameter $\lambda_n $ is chosen to decay relatively fast, so
	that the penalization $P_{en}(\cdot)$ does not have a first-order impact on
	the asymptotic theory. Nevertheless, the functional penalization is imposed
	to overcome undesirable properties associated with estimates based on a
	large parameter space. Essentially, it plays a role of forcing the
	optimization to be carried out within a weakly compact set %
	\citep{shen1997methods}.
	
	The functions $(x,\alpha )\mapsto \widehat{m}(x,\alpha )$ and $x\mapsto
	\widehat{\Sigma }(x)$ are nonparametric estimators of $(x,\alpha )\mapsto
	m(x,\alpha )$ and $x\mapsto \Sigma (x)$ (a positive definite weighting
	matrix) respectively. The projection $m(X_t,\alpha)$ can be also estimated
	using linear sieves:
	\begin{equation*}
	\widehat m(\cdot,\alpha)= \min_{m\in\mathcal{D}_{n}}
	\sum_{t=1}^n[\rho(Y_{t+1},\alpha)-m(X_t)]^2
	\end{equation*}
	where we consider linear sieve space: let $\{\Psi_j: j=1,\dots,k_n\}$ denote a set of sieve bases, 
	\begin{equation*}
	\mathcal{D}_n:=\left\{ g(x)= \sum_{j=1}^{k_n} \pi_j\Psi_j(x):
	\|g\|_{\infty,\omega}<\infty, \pi_j\in\mathbb R\right\}.
	\end{equation*}
	So we use the general nonlinear sieve learning space $\mathcal{H}_n$ to approximate the function
	space for $h_0$, and a linear sieve space $\mathcal{D}_n$ to approximate the
	instrumental space, which is easier to implement computationally than using
	nonlinear sieve approximations to the instrumental space. A more important
	motivation of using linear sieve space to estimate the conditional mean
	function $\mathbb{E}[\rho(Y_{t+1},\alpha)|\sigma_t(\mathcal{X})]$ is that
	the sample loss function $Q_n(\alpha)$ can be shown to have a local
	quadratic approximation (LQA): for some $B_n=O_P(1)$ and $Z_n\to^dN(0,1)$,
	\begin{equation}
	Q_n(\alpha+xu_n)-Q_n(\alpha)= B_nx^2+2x[n^{-1/2}Z_n+ \langle u_n,
	\alpha-\alpha_0\rangle] +o_P(n^{-1})
	\end{equation}
	uniformly for all $\alpha$ in a shrinking neighborhood of $\alpha_0$ and $%
	|x|\leq Cn^{-1/2}$; here $\langle u_n, \alpha-\alpha_0\rangle$ is some inner
	product between $\alpha-\alpha_0$ and some function $u_n$, to be defined
	explicitly later. This LQA plays a fundamental role for the inferential
	theory of semiparametric inference using general nonlinear sieve learning methods.
	

	\subsection{Semiparametric efficient estimations}
	
	\label{disc:effcient}
	Let the parameter space of the true function be $\mathcal H_0$ and let $\mathcal A_0=\Theta\times \mathcal H_0$. 
	We are interested in the inference of $\phi(\alpha_0)$, where $\phi:\mathcal{%
		A}_0\to \mathbb{R}$ can be a known functional of $\alpha_0$.
	We also study the inference problem of unknown functionals, taking the form
	\begin{equation*}
	\phi(\alpha_0)= \mathbb{E }l(h_0(W_t))\,,
	\end{equation*}
	where $l(\cdot)$ is a known function. While the naive plug-in estimator $%
	\frac{1}{n}\sum_{t=1}^nl( \widehat h(W_t)) $ is also asymptotically normal,
	when the model contains endogenous variables, it is \textit{not}
	semiparametrically efficient. An important example of $\phi(\alpha_0)$ is the weighted average
	derivative of nonparametric instrumental variable regression (NPIV),
	defined as
	\begin{equation*}
	\phi(\alpha_0)= \mathbb{E}[\Omega(W_t)^{\prime }\nabla h_0(W_t)]\,,
	\end{equation*}
	where $\Omega(\cdot)$ is a known positive weight function and $\nabla h_0$
	denotes the gradient of the nonparametric regression function $h_0$. As
	documented by \cite{ai2012semiparametric}, the simple plug-in estimator is not an efficient estimator. To obtain a more efficient estimator, on the population level consider
	conditional (given $X_t$) projection of $l(h_0(W_t))$ onto $\rho(Y_{t+1},
	\alpha_0)$, and the corresponding  functional of interest also can be represented as $\phi(\alpha_0)$ with the functional:
	\begin{eqnarray}\label{eq2.3adfne}
\phi(\alpha)&=&\mathbb E\left[l(h(W_t)) - \Gamma_0(X_t) \rho(Y_{t+1}, \alpha)\right]\,,
	\end{eqnarray}
	where $\Gamma_0(X_t) = \mathbb{E }[l(h_0(W_t)) \rho(Y_{t+1}, \alpha_0)
	|\sigma_t(\mathcal{X}) ] \Sigma(X_t)^{-1} $ is the projection coefficient. We shall obtain efficient estimator of $\phi(\alpha_0)$ based on this expectation expression. 
	It is worthy to know that the added term $\Gamma_0(X_t)\rho(Y_{t+1,\alpha_0})$ is in effect only for endogenous regressors. In pure exogeneous models where $W_t=X_t$,  we have $\Gamma_0(X_t)=0$. In this case the moment condition (\ref{eq2.3adfne}) reduces to the original one $\phi(\alpha)= \mathbb El(h(W_t))$. 
	
	
	Let 
		\begin{equation}  
	\widehat\phi(\alpha)= \frac{1}{n}\sum_{t=1}^n [l(h(W_t)) - \widehat\Gamma_t
	\rho(Y_{t+1}, \alpha)]
	\end{equation}
	for some   estimator  $\widehat\Gamma_t$ to be defined later. Then we estimate the functional by $\widehat\phi(\widehat\alpha)$. 
	Asymptotically, we shall show that 
	\begin{equation}
\widehat\phi(\widehat\alpha)-\phi(\alpha_0)=[\phi(\widehat \alpha)-
	\phi(\alpha_0) ] + \frac{1}{n}\sum_{t=1}^n [\mathcal{W}_t-\mathbb E\mathcal W_t]+o_P(\sigma n^{-1/2}),
	\end{equation}
	where
	$\mathcal{W}_t= l(h_0(W_t)) - \Gamma_0(X_t) \rho(Y_{t+1}, \alpha_0)$
	and $\sigma^2$
	is the asymptotic variance. It is clear that the asymptotic distribution
	arises from two sources of uncertainties, and importantly, the nonparametric learning error $\phi(\widehat \alpha)- \phi(\alpha_0) $ plays a
	first-order role. 
	
	We shall show that in both known and unknown functional case, estimated $%
	\phi(\alpha_0)$ is asymptotically normal. We then provide  quasi-likelihood inference to
	construct confidence intervals for $\phi(\alpha_0)$.

	\section{Rates of Convergence}
	\label{sec:rate}
	
	\subsection{Weighted function space and sieve  learning space}
Since the supports of the endogenous variable  $W_t$ could be unbounded, we use a weighted sup-norm metric defined as 
\begin{equation}\label{eq3.1}
\|h\|_{\infty,\omega}=\sup_s|h(s)|  (1+|s|^2)^{-\omega/2},\quad \text{ for some } \omega>0.
\end{equation}
This is known as ``admissible weight" which is often used for $h_0(W_t)$ when $W_t$ has fat tailed distribution (Remark 2.6 of \cite{haroske2020nuclear}).    Smooth functions with unbounded support might still be well approximated under the weighted sup-norm. 
The $L^2(W)$-norm can be bounded by the weighted sup-norm as: for any function $h(w)$:
$$
\|h\|_{L^2(W)}^2=\int h(s)^2 f_W(s)ds\leq
\|h\|^2_{\infty,\omega}\int   (1+|s|^2)^{\omega}
 f_W(s)ds,
 $$
provided the distribution of the endogenous variable $W$ has as density $f_W$ such that  $ f_W(s)  (1+|s|^2)^{\omega}$ is integrable.

 We do not consider the overparametrized regime, but impose restrictions on the complexity of the general nonlinear sieve learning space $\mathcal H_n$, measured by the ``number of parameters" of the space, denoted by  $p(\mathcal{H}_n)$. More specifically, we impose the following condition.
 
 \begin{assum}[function and learning space]\label{assumcomplexity}
 
 (i) The function space: The unknown function $h_0\in\mathcal H_0$, which is a weighted H\"{o}lder ball: for some $\gamma>0,g\geq0$,
 $$
 \mathcal H_0=\{h: \|h(\cdot)(1+|\cdot|^2)^{-g/2}\|_{\Lambda^\gamma}\leq c \}
 $$
 where
 $$
 \|f\|_{\Lambda^\gamma}=\sup_w|f(w)|+\max_{|a|=d}\sup_{w_1\neq w_2}\frac{|\nabla^af(w_1)-\nabla^af(w_2)|}{\|w_1-w_2\|^{\gamma-d}}.
 $$
 Also, we require $g<\omega$ for $\omega$ defined in (\ref{eq3.1}).

 (ii) Approximation rate under the $\|\|_{\infty,\omega}$ norm:
 $$
\inf_{h\in\mathcal H_0} \|	h_0-h\|_{\infty,\omega}\leq cp(\mathcal H_n)^{-m}
 $$
 for some $m>0$, and some sequence    $p(\mathcal H_n)\to\infty$,  $p(\mathcal{H}_n)\log n=o(n)$.
 
 (iii)    Complexity:   Let $\mathcal N(\delta,\mathcal H_n, \|.\|_{\infty,\omega})$ denote the minimal covering number, that is, the minimal number of
closed balls of radius $\delta$ with respect to  $\|.\|_{\infty,\omega}$ needed to cover $\mathcal H_n$. We assume, there is a constant $C>0$,  so that for any $\delta>0$, 
$$
\mathcal N(\delta,\mathcal H_n, \|.\|_{\infty,\omega})\leq \left(\frac{Cn}{\delta}\right)^{p(\mathcal H_n)}.
$$

 \end{assum}
	
	%
	
We need to   assume that $h(w)$ is smooth in some sense with respect to $h(w)$. Condition (i) is a
 standard weighted smoothness condition for functions with unbounded support.   Here two weighted norms are being defined, the weighted sup norm $\|.\|_{\infty,\omega}$ with a weight parameter $\omega$ in (\ref{eq3.1}). The weighted sup norm intead of the usual sup norm is being considered, as discussed above, for the purpose of allowing the nonparametric function $h(\cdot)$ to have possibly unbounded support, which is the typical case for autoregressive models.  The other norm is  $\|.\|_{\Lambda^\gamma}$ for the H\"{o}lder ball with a weight parameter $g$. Here we require $g<\omega$ so that the closure of the function space $\mathcal H_0$ with respec to the norm $\|.\|_{\infty,\omega}$ is   compact, following from \cite{gallant1987semi}.
 
 In Condition (ii),  $p(\mathcal H_n)\to\infty$ measures  the dimension of of the learning space.  	For multilayer neural networks with ReLU activation functions, \cite{anthony2009neural} showed that the bound holds with $p(\mathcal H_n)$ being the pseudo-dimension of the space and is bounded by $ C J^2 K^2\log(JK^2) $, where $J$ and $K$ respectively denote the width and depth of the network. For finite-dimensional linear sieve, the inequality also holds with $p(\mathcal H_n)$ being bounded by the number of sieve bases.
 
When the function $h$ has bounded support,   Condition (ii) has been verified for numerous learning spaces. For instance,  for feed forward multilayer neural networks, \cite{bauer2019deep} showed that the approximation rate is 
$n^{-c},$ for $ c= \frac{p}{2p+d^*}$ and $ p=a+\gamma$, with 
 properly chosen depth and width of layers. Importantly, $d^*\leq \dim(W_t)$ is the  ``intrinsic dimension" of the true function. For instance if $h_0$ has a hierarchical interaction structure or multi-index structure, $d^*$ is the number of index. When the function $h$ has unbounded support, it is known that for linear sieves such as B-splines and wavelets  the approximation rate is  $m=p(\mathcal H_n)^{-\gamma/\dim(W_t)}$ where $p(\mathcal H_n)$ is the number of basis. The approximation rate is  however still an open question for feed forward neural networks in this case.  
	

	\subsection{Ill-posedness}
	
	In this section we present the rate of convergence. 	For simplicity throughout the rest of the paper, we focus on the case $%
	\dim(\rho(Y_{t+1},\alpha))=1$. By the identification
	condition, $Q(\alpha)=0$ if and only if $\alpha=\alpha_0.$ So the usual
	\textit{risk consistency} refers to $Q(\widehat\alpha)=o_P(1)$. In the
	presence of endogenous variables, the risk consistency however, is not
	sufficient to guarantee the estimation consistency. The latter is often
	defined under a strong norm:
	\begin{equation*}
	\|\alpha_1-\alpha_2\|_{\infty,\omega}:= \|\theta_1-\theta_2\|+\|h_1-h_2\|_{\infty,\omega}.
	\end{equation*}
 
	We first introduce a pseudometric on $\mathcal{A}_n$ that is weaker than $%
	\|.\|_{\infty,\omega}$. To do so, recall the general Gateaux derivative. Given generic $%
	\alpha=(\theta, h) $ and $v=(v_\theta, v_h)$, let $F(x,\alpha)=F(x,\theta,
	h) $ be a function that is assumed to be differentiable with respect to $%
	\theta$. Define
	\begin{eqnarray*}
		\frac{dF(x, \alpha)}{d\alpha}[v]&=&\frac{\partial F(x,\alpha)}{\partial\theta%
		}^{\prime }v_{\theta} + \frac{dF(x, \theta, h+\tau v_h)}{d\tau}\bigg{|}%
		_{\tau=0},
	\end{eqnarray*}
	where we implicitly assume $\frac{dF(x, \theta, h+\tau v_h)}{d\tau}$ exists
	at $\tau=0.$ Then the \textit{weak} norm is defined to be
	\begin{equation*}
	\|v\|^2:= \mathbb{E}\left( \frac{dm(X_t, \alpha_0)}{d\alpha}[v] \right)^2
	\Sigma(X_t)^{-1} .
	\end{equation*}
	Define $\pi_n\alpha_0\in\mathcal{A}_n$ be such that
	\begin{equation*}
	\|\pi_n\alpha_0-\alpha_0\|_{\infty,\omega}=\min_{\alpha\in\mathcal{A}%
		_n}\|\alpha-\alpha_0\|_{\infty,\omega}.
	\end{equation*}
	The following assumption imposes conditions on the local curvature of the
	criterion function.
	
	\begin{assum}[criterion function]
		\label{ass2.1} There are $c_1, c_2>0$ so that
		
		(i) $\|\alpha-\alpha_0\|^2\leq c_1\mathbb{E }m(X_t,\alpha) ^2
		\Sigma(X_t)^{-1}   $ for all $\alpha\in\mathcal A_n$. 
		
		(ii) $\mathbb{E }m(X_t,\pi_n\alpha_0)^2 \Sigma(X_t)^{-1} \leq c_2
		\|\alpha_0-\pi_n\alpha_0\|^2$.
	\end{assum}
	
	We now discuss the ill-posedness which reflects the relation between the
	risk consistency and estimation consistency. Let the \textit{sieve modulus
		of continuity} be
	\begin{equation*}
	\omega_n(\delta):= \sup_{\alpha\in\mathcal{A}_n:
		\|\alpha-\pi_n\alpha_0\|\leq\delta} \|\alpha-\pi_n\alpha_0\|_{\infty,\omega}.
	\end{equation*}
	
	We say that the problem is \textit{ill-posed} if $\delta=o(\omega_n(\delta))$
	as $\delta\to0.$ The growth of $\omega_n(\delta)\delta^{-1}$ reflects the
	difficulty of recovering $\alpha_0$ through minimizing the criterion
	function.
	
	\subsection{Rates of convergence}
	
	Below we present regularity conditions to achieve the rates of convergence. We allow weakly dependent time series data satisfying $\beta$-mixing
	conditions. Define the mixing coefficient
	\begin{equation*}
	\beta(j):=\sup_{t}\mathbb{E}\sup\{|P(B|\mathcal{F}_{-\infty}^t)- P(B)|: B\in%
	\mathcal{F}_{t+j}^{\infty}\}
	\end{equation*}
	where $\mathcal{F}_s^t$ denotes the $\sigma$-field generated by $(Y_{s+1},
	X_s),...,(Y_{t+1}, X_t)$.
	
	\begin{assum}[Dependences]
		\label{ass3.10} (i) $\{(Y_{t+1}, X_t)\}_{t=1}^n$ is a strictly stationary
		and $\beta$-mixing sequence with $\beta(j)\leq \beta_0\exp(-cj)$ for some $%
		\beta_0, c>0$.
		
		(ii) There is a known and finite integer $r\geq 1$ so that for each $%
		\alpha\in\mathcal{A}_n$ and $t=1,...,n$, The conditional expectation $%
		\mathbb{E}[ f(S_t, \alpha)|\sigma_t(\mathcal{X})]$ depend on $\sigma_t(%
		\mathcal{X})$ only through $X_t:=(\mathcal{X}_t,...,\mathcal{X}_{t-r})$, for
		$S_t=(Y_{t+1},W_t)$ and
		\begin{equation*}
		f(S_t,\alpha)\in \{ \rho(Y_{t+1},\alpha),\quad \rho(Y_{t+1},\alpha_0)^2 ,
		\quad l(h_0(W_t)) \rho(Y_{t+1}, \alpha_0) \}.
		\end{equation*}
	\end{assum}
	
	\begin{assum}
		\label{asfda.4} $Q(\alpha)=0$ if and only if $\alpha=\alpha_0.$ In
		addition, $Q(\alpha)$ is lower semicontinuous. 
	\end{assum}
	
	The lower semicontinuity of the criteria function is satisfied by the risk
	function of many interesting models. This condition ensures that it has a
	minimum on any compact set.
	
	\begin{assum}[Penalty]
		\label{ass3.2} (i) There is $M_0>0$, $P_{en}(h) \leq M_0$ for all $h\in%
		\mathcal{H}_n\cup\{h_0\}$.\newline
		(ii)  $P_{en}$ is lower semicompact on $(\mathcal{A}_n, \|.\|_{\infty,\omega})$, i.e. $\{h: P_{en}(h)\leq M\}$   is compact for any $M>0$. \newline
		(iii) $\frac{k_n}{n} +Q(\pi_n\alpha_0)=O(\lambda_n) $ where recall $k_n$ is the number of linear sieve bases in $\mathcal{D}_n$.
	\end{assum}
	
	 Define $$\epsilon(S_t,\alpha):=
	\rho(Y_{t+1},\alpha)-m(X_t,\alpha).$$ 
		One of the major technical steps is to establish the stochastic
	equicontinuity for the function class $\Psi_j(X_t)\epsilon(S_t,\alpha)$ for $%
	\beta$-mixing observations, where $\alpha$ belongs to the class of deep
	neural networks. More specifically, we shall derive the  bound for, with $\Psi(X_t):=(\Psi_j(X_t): j\leq k_n)$:
	\begin{equation*}
	\sup_{\alpha\in\mathcal{A}_n: \mathbb{E}m(X_t,\alpha)\leq r_n^2}\left\|\frac{1%
	}{\sqrt{n}}\sum_{t=1}^n\Psi(X_t)[\epsilon(S_t,\alpha)- \epsilon(S_t,\alpha_0) ]
	\right\|
	\end{equation*}
	for a given convergence sequence $r_n\to 0$. This is achieved under the following Assumption.
 	
	\begin{assum}
		\label{ass3.1} There is $C>0$,
		
		
		(i) There are $\kappa>0$ and $C>0$ so that for all $\delta>0$ and all $%
		\alpha_1, \alpha_2\in\mathcal{A}_n$,
		\begin{equation*}
		\max_{j\leq k_n} \mathbb{E}[\Psi_j(X_t)^2+1]\sup_{ \|\alpha_1-\alpha_2
			\|_{\infty,\omega}< \delta} | \epsilon( S_t, \alpha_1)- \epsilon( S_t, \alpha_2 )|
		^2\leq C\delta^{2\kappa}.
		\end{equation*}
		
		(ii) $\mathbb{E}\max_{j\leq k_n}\Psi_j(X_t)^2 \sup_{\alpha\in\mathcal{A}_n}
		\rho(Y_{t+1}, \alpha )^2\leq C.$
		
		(iii) There is a $\|.\|_{\infty,\omega}$- neighborhood of $\alpha_0$ on which $m(\cdot,
		\alpha)$ is continuously pathwise differentiable with respect to $\alpha$,
		and there is a constant $C>0$ such that $\|\alpha-\alpha_0\|\leq
		C\|\alpha-\alpha_0\|_{\infty,\omega}.$

	\end{assum}

	Next we present regularity conditions on the linear sieve space $\mathcal{D}%
	_n$ used to approximate the conditional mean function $m(X,\alpha)$.
	
	\begin{assum}[Linear sieve space]
		\label{fa.45sass}
		
		(i) There is $\varphi_n\to0$ so that uniformly for $\alpha\in\mathcal{A}_n$,
		there is $k_n\times 1$ vector $b_{\alpha}$,
		\begin{equation*}
		\mathbb{E }[g(X_t, \alpha)- \Psi(X_t)^{\prime }b_\alpha ]^2=O(\varphi_n^2),
		\end{equation*}
		for all $g(X_t,\alpha)\in \{m(X_t,\alpha),\mathbb{E }[l(h_0(W_t))
		\rho(Y_{t+1}, \alpha_0)|X_t], \frac{dm(X_t,\alpha)}{d\alpha}[u_n], \frac{%
			dm(X_t,\alpha)}{d\alpha}[u_n]\Sigma(X_t)^{-1}\}$.
		
		
		(ii) Let $\Psi_n $ be the $n\times k_n$ matrix of the linear sieve bases: $\Psi_n=(\Psi(X_t): t=1...n) _{n\times k_n}$:
		and let $A:=\frac{1}{n}\mathbb{E }\Psi_n^{\prime }\Psi_n$. The linear sieve
		satisfies: $\lambda_{\min}(A)>c$ and $\| \frac{1}{n}\Psi_n^{\prime }\Psi_n
		-A\|=o_P(1)$.
	\end{assum}
	
	Finally, we apply the \textit{pseudo dimension} to quantify the complexity
	of the neural network class. 
	
	\begin{assum}
		\label{a34}
		
		(i) $\sup_x[ \Sigma(x)^{-1}+ \Sigma(x)]<C .$ Also, $\sup_x|\widehat
		\Sigma(x)-\Sigma(x)|=o_P(1)$.
		
		(ii)   The distribution of the endogenous variable $W_t$ has a density function $f_W$, which satisfies $\int  w(x)^{-2} f_W(x)dx<\infty.$

	\end{assum}

	Recall that $k_n$ denotes the number of sieve bases being used to estimate the expectation function $m(X,\alpha)$; $\varphi_n$ is the approximation rate in Assumption \ref{fa.45sass}. Let 
 	\begin{eqnarray} \label{def_dn_deltan}
 	d_n&:=&\sqrt{\frac{ p(\mathcal{H}_n) \log^2 n}{n}},\cr
	\bar\delta_n&:=&\|\pi_n\alpha_0-\alpha_0\|+ \sqrt{\lambda_n}+\sqrt{k_n}%
	d_n+\varphi_n ,\cr 
	\delta_n&:=&\|
	\pi_n\alpha_0-\alpha_0\|_{\infty,\omega}+\omega_n(\bar\delta_n).
	\end{eqnarray}
	
	\begin{thm}[Rate of convergence]
		\label{th3.1} Under Assumptions \ref{ass2.1}-\ref{a34}, for any $\epsilon>0$%
		,
		\begin{equation*}
		\|\widehat\alpha-\alpha_0\|_{\infty,\omega}=
	O_P(\delta_n),\quad Q(\widehat\alpha)=O_P(\bar\delta_n^2).
		\end{equation*}

	\end{thm}

	The derived rate of convergence is comparable with that of \cite%
	{chen2012estimation}. In $\bar \delta_n$, the term $\|\pi_n\alpha_0-\alpha_0\|$
	is the approximation error on the general nonlinear sieve learning space; $\sqrt{\lambda_n}$ is the
	effect of penalization. In addition, $\varphi_n$ and $\sqrt{k_n}d_n$
	respectively arise from the bias and variance of estimating $m(X,\alpha)$.
	In particular, the variance term $\sqrt{k_n}d_n$ depends on the complexity
	of the general nonlinear sieve learning space, which arises from  the stochastic equicontinuity. 
	In addition, $\omega_n(\bar\delta_n)$ connects the convergence under the
	weak norm $O_P(\bar\delta_n)$ to the convergence under the strong norm via
	the sieve modulus of continuity. When there are no endogeneity, $%
	\bar\delta_n $ and $\omega_n(\bar\delta_n)$ are of the same order. General nonlinear sieve spaces  with more complicated structures (with larger ``dimension" $p(\mathcal H_n)$) have
	increased covering numbers on the learning  space, and thus lead to slower decays
	of these two terms. 
	
	\section{Asymptotic Distributions for  Functionals}
	
	\label{sec:normality}
	
	We now study estimating linear functionals of $\alpha_0$.   We establish the
	asymptotically normality of the estimated functionals formed via pluging-in the
	general learning estimators.
	
	
	\subsection{Riesz representation}
	
	A key ingredient of our analysis, as in \cite{chen2015sieve}, relies on
	representing the estimation error $\phi(\widehat\alpha)-\phi(\alpha_0)$
	using a linear inner product induced from the loss function via the
	Riesz representation theorem. We define an inner product space as follows.
	
	
	
	
	For any space $\mathcal{H}$, let span$\{\mathcal{H}\}$ denote the closed
	linear span of $\mathcal{H}$. For any $v_1, v_2$ in span$(\mathcal{A}_n
	\cup\{\alpha_0\})$, the linear span of $\mathcal{A}_n \cup\{\alpha_0\}$,
	define the inner product:
	\begin{equation*}
	\langle v_1, v_2\rangle=\mathbb{E}\Sigma(X_t)^{-1}\left( \frac{dm(X_t,
		\alpha_0)}{d\alpha}[v_1] \right)\left( \frac{dm(X_t, \alpha_0)}{d\alpha}%
	[v_2] \right).
	\end{equation*}
	Let $\alpha_{0,n}\in $ span$(\mathcal{A}_n )$ be such that
	\begin{equation*}
	\|\alpha_{0,n}-\alpha_0\|=\min_{\alpha\in \text{span}(\mathcal{A}_n
		)}\|\alpha-\alpha_0\| .
	\end{equation*}
	We note that it is likely $\alpha_{0,n}\neq\pi_n\alpha_0$ because $%
	\pi_n\alpha_0\in\mathcal{A}_n$, which is not the same as $\text{span}(%
	\mathcal{A}_n )$, when $\mathcal{A}_n$ is a nonlinear sieve space.

		Given Theorem \ref{th3.1}, we can focus on  shrinking neighborhoods
	\begin{eqnarray}
	\mathcal{A}_{osn}&:=&\{  \alpha\in\mathcal{A}_n:
	\|\alpha-\alpha_0\|_{\infty,\omega}\leq C\delta_n, Q(\alpha)\leq C\bar\delta_n^2\}\cr
		\mathcal{C}_n&:=&\{\alpha+ xu_n: \alpha\in\mathcal{A}_{osn}, |x|\leq
	Cn^{-1/2}\},\quad u_n:=v_n^*/\|v_n^*\|,\cr
		\bar V_n&:=&\text{span}( \mathcal{A}_{osn} - \{ \alpha_{0,n} \}) \subset\text{%
		span}(\mathcal{A}_n).
	\end{eqnarray}
	for a generic constant $C>0$, where $v_n^*$ is the \textit{Riesz representer} to be defined below.

	Because both $\mathcal{A}_{osn} $ and $\alpha_{0,n}$ are functions inside
	the general nonlinear sieve learning space, $(\bar V_n, \langle.\rangle)$ is a finite dimensional
	Hilbert space under the weak-norm $\|v\|=\sqrt{\langle v, v\rangle}$.
	Suppose $\frac{d\phi(\alpha_0)}{d\alpha}[v]$ is a linear functional. As any
	linear functional on a finite dimensional Hilbert space is bounded, by the
	Riesz representation Theorem, there is $v_n^*\in \bar V_n$ so that
	\begin{equation*}
	\frac{d\phi(\alpha_0)}{d\alpha}[v]=\langle v_n^*,v\rangle, \quad \forall
	v\in \bar V_n.
	\end{equation*}
	
	To appreciate the role of Riesz representation in the semiparametric
	inference, note that $\widehat\alpha- \alpha_{0,n}\in \bar V_n$, and we have, 
	\begin{eqnarray*}
		\phi(\widehat\alpha)-\phi(\alpha_0)&=& \frac{d\phi(\alpha_0)}{d\alpha}%
		[\widehat\alpha-\alpha_0] \cr &=& \frac{d\phi(\alpha_0)}{d\alpha}%
		[\widehat\alpha- \alpha_{0,n}] + \frac{d\phi(\alpha_0)}{d\alpha}[
		\alpha_{0,n}-\alpha_0] \cr &=& \langle v_n^*,\widehat\alpha-
		\alpha_{0,n}\rangle +\underbrace{\frac{d\phi(\alpha_0)}{d\alpha}[
			\alpha_{0,n}-\alpha_0] }_{\text{negligible}}.
	\end{eqnarray*}
	where the first equality follows from the smoothness condition (Assumption %
	\ref{asssmoothphi} below) of the functional; the second equality is to the
	linearity of the functional pathwise derivative. In addition, suppose $\frac{%
		d\phi(\alpha_0)}{d\alpha}[ \alpha_{0,n}-\alpha_0]$ is negligible, a claim we
	shall discuss in Remark \ref{remar4.1} later, we can then apply the Riesz
	representation theorem to reach the last line of the expansion.
	
	In addition,  one of the key technical steps in the proof,   by locally expanding  the risk function, is to prove:
	\begin{equation*}
	\sqrt{n} \langle v_n^*,\widehat\alpha-\alpha_{0,n}\rangle = \sqrt{n} \langle
	v_n^*,\widehat\alpha-\alpha_0\rangle = -\frac{1}{\sqrt{n}} \sum_t \mathcal{Z}%
	_t +o_P(\|v_n^*\| )
	\end{equation*}
	where $\mathcal{Z}_t=\rho(Y_{t+1}, \alpha_0)\Sigma(X_t)^{-1}\frac{d
		m(X_t,\alpha_0)}{d\alpha}[v^*_n] ,$ and $\|v_n^*\| ^2=\var(\frac{1}{\sqrt{n}}%
	\sum_t\mathcal{Z}_t).$ Then together we have 
	\begin{equation*}
	\frac{ \sqrt{n} (\phi(\widehat\alpha)-\phi(\alpha_0))}{\|v_n^*\| } \to^d%
	\mathcal{N}(0,1).
	\end{equation*}

	Importantly, our inference procedure \textit{does not} require estimating  the Riesz representer $v_n^*$ or $\|v_n^*\|$. Instead, we propose a quasi-likelihood ratio   (QLR) inference. 
	We shall provide regularity conditions in the next section to formalize the
	above derivations, and subsequently address estimating the known and unknown
	functionals.
	
	\subsection{Asymptotic distributions for known functionals}

	We have the following assumptions.
	
	\begin{assum}[smoothness]
		\label{asssmoothphi} (i) The functional $\phi$ is linear in the sense that 
		the functional $\phi$ is linear in the sense that  $
	\phi(\alpha)-\phi(\alpha_0)= \frac{d\phi(\alpha_0)}{d\alpha}%
		[\alpha-\alpha_0]  $.

		
		(ii) $\sqrt{n}\frac{d\phi(\alpha_0)}{d\alpha}[ \alpha_{0,n}-\alpha_0] =
		o_P(\|v_n^*\|).$
	\end{assum}
	
	\begin{remark}
		\label{remar4.1} Assumption \ref{asssmoothphi} (iii) requires that the
		neural network bias term $\frac{d\phi(\alpha_0)}{d\alpha}[
		\alpha_{0,n}-\alpha_0]$ should be negligible. Here we present a sufficient
		condition following the discussion of \cite{chen2015sieve}. First, since $%
		\alpha_{0,n}$ is the projection of $\alpha_0$ on to span$(\mathcal{A}_n)$
		and $v_n^*\in \bar V_n\subset$ span$(\mathcal{A}_n)$, we have $\langle
		v_n^*, \alpha_{0,n}-\alpha_0\rangle=0$. In addition, define an infinite
		dimensional Hilbert space $\bar V$ as the closure of the linear span of $%
		\mathcal{A}- \{\alpha_0\}. $ Suppose $\frac{d\phi(\alpha_0)}{d\alpha}[ \cdot
		]$ is bounded, then there is a unique Riesz representer $v^*\in\bar V$ so
		that
		\begin{equation*}
		\frac{d\phi(\alpha_0)}{d\alpha}[v]=\langle v^*,v\rangle, \quad \forall v\in
		\bar V.
		\end{equation*}
		As $\alpha_{0,n}-\alpha_0\in\bar V $, we have
		\begin{equation*}
		\left|\sqrt{n}\frac{d\phi(\alpha_0)}{d\alpha}[ \alpha_{0,n}-\alpha_0]
		\right| = \left|\sqrt{n} \langle v^*-v_n^*, \alpha_{0,n}-\alpha_0\rangle
		\right| \leq \sqrt{n} \| v^*-v_n^*\| \| \alpha_{0,n}-\alpha_0\|.
		\end{equation*}
		So condition (iii) holds as long as $\sqrt{n} \| v^*-v_n^*\| \|
		\alpha_{0,n}-\alpha_0\|=o_P(\|v_n^*\|)$.
	\end{remark}
	
	
	To allow quantile applications that involve nonsmooth loss functions, we need to show that the sample criterion function $Q_n(\alpha)$ can be replaced with a smoothed
	criterion $\widetilde Q_n(\alpha):= \frac{1}{n}\sum_t\ell(X_t,
	\alpha)^2\widehat\Sigma(X_t)^{-1}$, where $\Psi_n=(\Psi(X_t): t=1...n) _{n\times k_n}$:
	\begin{equation*}
	\ell(x, \alpha):=\widetilde m(x,\alpha) +\widehat m(x,\alpha_0),\quad
	\widetilde m(x,\alpha):=\Psi(x)^{\prime }(\Psi_n^{\prime
	}\Psi_n)^{-1}\Psi_n^{\prime }m_n(\alpha),
	\end{equation*}
	and $m_n(\alpha)$ denotes the $n\times 1$ vector of $m(X_t,\alpha)$. The replacement error is negligible: 
		\begin{equation*}
	\sup_{ \alpha\in\mathcal{A}_{osn}} \sup_{|x|\leq Cn^{-1/2}}|Q_n(\alpha+xu_n)-
	\widetilde Q_n(\alpha+xu_n) |=o_P(n^{-1}).
	\end{equation*}
	Therefore, theoretical analysis of $Q_n(\alpha)$ is asymptotically
	equivalent to that of $\widetilde Q_n(\alpha)$, while the latter is
	second-order pathwise differentiable, and admits a local quadratic
	approximation. Formalizing this argument  would require the following conditions.

	\begin{assum}
		\label{ass4.2} $m(x, t)$ is twice differentiable with respect to $t$, and
		there is $C>0$, so that, recall that $u_n= v_n^*/\|v_n^*\|$ being the ``normalized Riesz representer":
		
		(i) $\mathbb{E }| \rho(Y_{t+1}, \alpha_0)|^{2+\zeta}\left| \frac{d
			m(X_t,\alpha_0)}{d\alpha}[u_n] \right |^{2+\zeta}+\mathbb{E }| \rho(Y_{t+1},
		\alpha_0)|^{2+\zeta}<C$ for some $\zeta > 0$;
		
		(ii) $\mathbb{E }\sup_{\alpha\in\mathcal{C}_{n}} \sup_{|\tau|\le Cn^{-1/2}} 
		\frac{1}{n}\sum_t\left[\frac{d^2}{d\tau^2} m(X_t, \alpha+\tau u_n)| \right]%
		^2 <C$;
		
		(iii) $\sup_{\tau\in(0,1)}\sup_{ \alpha\in\mathcal{C}_n} \mathbb{E }\left[
		\frac{d ^2}{d\tau^2} m(X_t,\alpha_0+\tau(\alpha-\alpha_0))\right]^2
		=o(n^{-1})$;
		
		
		(iv) $k_n \sup_{\alpha\in\mathcal{C}_n}\frac{1}{n}\sum_t[ \frac{%
			dm(X_t,\alpha )}{d\alpha}[u_n]- \frac{dm(X_t,\alpha_0)}{d\alpha}%
		[u_n]]^2=o_P(1)$;
		
		
		(v) $\mathbb{E} \big\{[\max_{j\leq k_n}\Psi_j(X_t)^2+1]\sup_{\alpha\in\mathcal{C}%
			_n} (\rho(Y_{t+1}, \alpha)-\rho(Y_{t+1}, \alpha_0) )^2\big\} < C\delta_n^{2\eta} $
		for some $\kappa, \eta >0$.
	\end{assum}

	Finally, we need to strengthen conditions on the penalty and some rates of convergence as follows. 
	
	\begin{assum}	\label{ass4.3new}
	(i)	Let $\mathcal{C}_h:=\{h: (\theta, h)\in\mathcal{C}_n \text{ for some $%
			\theta\in \Theta$}\}$, which is the local neighborhood for the estimated $%
		h(\cdot)$. We assume
		\begin{equation*}
		\lambda_n \sup_{h\in\mathcal{C}_h }|P_{en}( h) - P_{en}(h_0 ) | + \lambda_n
		\sup_{h\in\mathcal{C}_h }|P_{en}( \pi_nh) - P_{en}(h_0 ) | =o(n^{-1}).
		\end{equation*}
		(ii)  
		$\sqrt{n}\bar \delta_n\|\widehat\Sigma_n-\Sigma_n\| =o(1)$, where $\widehat\Sigma_n$ and $\Sigma_n$ be the diagonal matrix of $\widehat\Sigma(X_t)$ and $\Sigma(X_t)$ for all $t$, and furthermore $%
		\varphi_n^2\bar\delta_n^2 + k_nd_n^2\delta_n^{2\eta} + \sqrt{k_n}
		d_n\delta_n^{\eta}\bar\delta_n =o(n^{-1})$.
		
	\end{assum}
	


 The following condition is similar to Condition C in \cite{shen1997methods}, which is
used to control the approximation error of the  learning space for locally
perturbed elements.
	
	\begin{assum}
		\label{ass4.4} There is $\mu_n\to 0$ so that $\mu_n\bar \delta_n = o(n^{-1})
		$, we have
		\begin{equation*}
		\sup_{\alpha\in\mathcal{C}_n} \frac{1}{n}\sum_{t=1}^n [m( X_t, \pi_n
		\alpha)- m( X_t, \alpha)]^2 =O_P(\mu_n^2).
		\end{equation*}
	\end{assum}

	\begin{thm}[Limiting distribution]
		\label{th4.1} Under Assumptions \ref{ass3.10}-\ref{ass4.4},
		\begin{eqnarray*}
			\sqrt{n} \frac{\phi(\widehat\alpha)-\phi(\alpha_0)}{ \|v_n^*\| } \to^d%
			\mathcal{N}(0,1).
		\end{eqnarray*}
	\end{thm}
	
	An important insight from this theorem is that the asymptotic distribution  does not depend on the actual choice of the learning space. The asymptotic variance  
	$$\|v_n^*\| ^2=\mathbb{E}
	\Sigma(X_t)^{-1}\left( \frac{dm(X_t, \alpha_0)}{d\alpha}[v_n^*] \right)^2$$ is only determined by the functional forms $\phi$ and $m(X,\alpha)$, and more generally, the loss function. So whether the multilayer neural network, B-spline, Gaussian radial basis, etc, are being used to estimate $\alpha_0$, the asymptotic distribution is the same.  What really matters is the loss function.

	\subsection{Estimation for unknown functionals}
	
	\label{sec:effi}
	
	We now consider estimating unknown (probably not $\sqrt{n}$-estimable)
	functionals, taking the form
	\begin{equation*}
	\gamma_0:= \mathbb{E }l(h_0(W_t)),
	\end{equation*}
	where $l(\cdot)$ is a known function.  
	\cite{ai2012semiparametric} used
	the  following moment condition (\ref{eq4.1adfae}) to construct the optimal criterion
	function: \\
	\begin{equation}  \label{eq4.1adfae}
	\gamma_0= \mathbb{E}\mathcal{W}_t,\quad \mathcal{W}_t= l(h_0(W_t)) -
	\Gamma(X_t) \rho(Y_{t+1}, \alpha_0),
	\end{equation}
	where $\Gamma(X_t) = \mathbb{E }[l(h_0(W_t)) \rho(Y_{t+1}, \alpha_0)
	|\sigma_t(\mathcal{X}) ] \Sigma(X_t)^{-1} . $ They showed that estimating $\gamma_0 $
	based on this moment condition leads to more efficient estimator than based on the naive plug-in method $\frac{1}{n}\sum_i l(\widehat h(W_t))$, whenever $W_t$ is endogenous. Because the naive plug-in estimator  does not take into account the
	potential correlations between the moment functions $m(X_t, \alpha)$ and $%
	l(h(W_t))$.
	
	Using the more efficient moment condition of $\gamma_0$, and letting $$\phi(\alpha):= \mathbb{E}l(h(W_t))-\mathbb{E}\Gamma(X_t)\rho(Y_{t+1},\alpha),$$ 
	we note that $%
	\phi(\alpha_0)=\gamma_0.$ Suppose the functional $\phi(\cdot)$ were known, and Assumption \ref{asssmoothphi}
	continues to hold for $\phi(\alpha)$, then we can show
	\begin{equation*}
	\sqrt{n}(\phi(\widehat \alpha)-\phi(\alpha_0))\approx- \frac{1}{\sqrt{n}}%
	\sum_{t=1}^n\mathcal{Z}_t,\quad \mathcal{Z}_t:=\rho(Y_{t+1},
	\alpha_0)\Sigma(X_t)^{-1}\frac{d m(X_t,\alpha_0)}{d\alpha}[v_n^*] ,
	\end{equation*}
	where $v^*_n$ is the Riesz representer. But  we in fact are facing a problem of estimating an \textit{unknown functional} $\phi(\cdot)$. To do so,
	we first estimate $%
	\Gamma (X_t)$ by
	\begin{equation*}
	\widehat\Gamma_t:= \sum_{s=1}^n l(\widehat
	h(W_s))\rho(Y_{s+1},\widehat\alpha) \phi(X_s)^{\prime }(\Psi_n^{\prime
	}\Psi_n)^{-1} \Psi(X_t)\widehat \Sigma(X_t)^{-1}.
	\end{equation*}
	Then define the final estimator: 
	\begin{equation}  \label{eqa4.1}
	\widehat\gamma:= \widehat\phi(\widehat\alpha), \text{ where }
	\widehat\phi(\alpha)= \frac{1}{n}\sum_{t=1}^n [l(h(W_t)) - \widehat\Gamma_t
	\rho(Y_{t+1}, \alpha)].
	\end{equation}
	

	The following asymptotic expansion holds for the estimated functional:
	\begin{eqnarray*}
		\widehat\gamma-\gamma_0&=&[\phi(\widehat \alpha)- \phi(\alpha_0) ] +\frac{1}{n}%
		\sum_{t=1}^n[ \mathcal{W}_t-\mathbb{E }\mathcal{W}_t ]+o_P(\sigma n^{-1/2})\cr
		&=& \frac{1}{n}\sum_{t=1}^n[-\mathcal Z_t+ \mathcal{W}_t-\mathbb{E }\mathcal{W}_t ]+o_P(\sigma n^{-1/2})
	\end{eqnarray*}
 	where $\mathcal{Z}_t=\rho(Y_{t+1}, \alpha_0)\Sigma(X_t)^{-1}\frac{d
		m(X_t,\alpha_0)}{d\alpha}[v^*_n] .$
This explicitly presents two leading sources for the asymptotic
	distribution, where the asymptotic variance is given by 
	\begin{equation}\label{eq4.3ardtga}
	\sigma^2:=\frac{1}{n}\var\left( \sum_{t=1}^n(\mathcal{W}_t-\mathcal{Z}_t)\right) = \frac{1}{n}\var\left( \sum_{t=1}^n \mathcal{W}_t\right) +  \|v_n^*\| ^2.
	\end{equation}
	where $\mathcal W_t$ and $\mathcal Z_t$ are uncorrelated. 
	
	We impose the following conditions
	
	\begin{assum}
		\label{a5.1} (i) $\sup_x|\Gamma(x)|^2 +\sup_w\sup_{h\in\mathcal{H}%
			_n}l(h(w))^2<C.$
		
		(ii) $l(h)$ is linear in $h$. 
		
		(iii) $\mathbb{E}\sup_{\alpha\in\mathcal{C}_n} |
		l(h(W_t))-l(h_0(W_t))|^2\leq C\delta_n^{2\eta}$, where  for simplicity we assume the same $\eta$ as in Assumption \ref{ass4.2} (v).
		
	\end{assum}
	
	Assumption \ref{a5.1} regulates the approximation quality of the
	instrumental space using linear sieves, which is not stringent since $%
	\mathbb{E}(l(h(W_t))\rho(Y_{t+1}, \alpha)|\sigma_t(\mathcal{X}))$ is a
	function of the instrumental variable.
	
	The next assumption imposes a condition on the accuracy of estimating the
	optimal weighting function $\Sigma(X_t)$. For the NPQIV model this
	assumption is trivially satisfied since $\widehat\Sigma(X_t)=\Sigma(X_t)=%
	\varpi(1-\varpi)$ is known (see Section \ref{sec6.3} for the definition of $\varpi$). We shall verify it for the NPIV model in Section \ref{sec6.2}.
	
	\begin{assum}
		\label{a5.555} There is a sequence $p_n $ so that $p_n
		\bar\delta_n\sigma=o(n^{-1})$ and
		\begin{equation*}
		\frac{1}{n}\sum_t \Gamma(X_t)\Sigma(X_t)
		(\widehat\Sigma(X_t)^{-1}-\Sigma(X_t)^{-1})\rho(Y_{t+1}, \alpha_0) =O_P(
		p_n).
		\end{equation*}
	\end{assum}
	
	
	The asymptotic normality requires some rate restrictions, which we impose
	below.
	
	\begin{assum}
		\label{a4.55} (i) There is $c_0>0$ so that $\sigma^2>c_0.$
		
		(ii) Let $\nu_n:=\delta_n^{\eta} \sup_x|\widehat\Sigma(x)-\Sigma(x)| +\sqrt{%
			k_n} d_n\delta_n^{\eta} +\varphi_n^2. $ Then $\nu_n
		\bar\delta_n\sigma=o(n^{-1})$.
	\end{assum}
	
	\begin{thm}
		\label{th4.2} Suppose Assumptions \ref{ass3.10}-\ref{ass4.4} hold for $%
		\phi(\alpha)=\mathbb{E}l(h(W_t))-\mathbb{E}\Gamma(X_t)\rho(Y_{t+1},\alpha)$.
		In addition, Assumptions \ref{a5.1}-\ref{a4.55} hold. Then
		\begin{equation*}
		\sqrt{n}\sigma^{-1}( \widehat\gamma-\gamma_0 ) \to^d\mathcal{N}(0,1).
		\end{equation*}
	\end{thm}
	
	
	\section{ Quasi-Likelihood Ratio Inference for Functionals}
	
	\label{sec:qlr}
	
	
	As shown by Theorems \ref{th4.1} and \ref{th4.2}, computing the asymptotic variance requires 
	estimating Riesz representer. While \cite{chen2015sieve} and \cite{chernozhukov2018automatic} proposed   framework of estimating the Riesz representer, the task is in general quite challenging when   its does not
	have closed-form approximations.  In this section we propose to make inference directly
	using the optimally weighted quas-likelihood ratio statistic (QLR).
	
	\subsection{QLR Inference for known functionals}
	
	Consider testing
	\begin{equation*}
	H_0: \phi(\alpha_0) =\phi_0
	\end{equation*}
	for some known $\phi_0\in\mathbb{R}.$ Consider the restricted null space $%
	\mathcal{A}_n^R:=\{\alpha\in\mathcal{A}_n: \phi(\alpha) =\phi_0\}$. The 
	GN-QLR statistic is defined as
	\begin{equation*}
	S_n(\phi_0)= n\left( Q_n(\widehat\alpha^R) -Q_n(\widehat\alpha) \right)
	\end{equation*}
	where $\widehat\alpha^R\in\mathcal{A}_n^R$ approximately minimizes the
	penalized loss function over the general nonlinear sieve learning restricted on the null space:
	\begin{equation*}
	Q_n( \widehat\alpha^R)+\lambda_n P_{en}(\widehat h^R) \leq \inf_{\alpha\in%
		\mathcal{A}_n^R} Q_n(\alpha)+\lambda_n P_{en}(\alpha)+o_P(n^{-1}).
	\end{equation*}
	
	Define
	\begin{eqnarray*}
		\pi_n^R\alpha= \arg\min_{b\in\mathcal{A}_n, \phi(b)=\phi_0}
		\|b-\alpha\|_{\infty,\omega}.
	\end{eqnarray*}
	
	\begin{assum}
		\label{ass5.1} (i) Recall $\mu_n$ as defined in Assumption \ref{ass4.4}%
		. It also satisfies:
		\begin{equation*}
		\sup_{\alpha\in\mathcal{A}_{osn}, \phi(\alpha)=\phi_0} \frac{1}{n}\sum_{t=1}^n
		[m( X_t, \pi_n^R (\alpha+xu_n))- m( X_t, \alpha+xu_n)]^2 =O_P(\mu_n^2)
		\end{equation*}
		
		(ii) $(1+\|v_n^*\|) \sup_{\alpha\in\mathcal{C}_n}|\phi(\pi_n\alpha)-\phi(%
		\alpha)| =o (n^{-1/2})$.
	\end{assum}
	
	The following theorem shows the asymptotic null distribution of $S_n(\phi_0)$%
	.
	
	\begin{thm}
		\label{thqlr} Suppose conditions of Theorem \ref{th4.1} and Assumption \ref{ass5.1} hold. Then under $H_0: \phi(\alpha_0)=\phi_0$,
		\begin{equation*}
		S_n(\phi_0)\to^d\chi^2_1.
		\end{equation*}
	\end{thm}
	
	\subsection{QLR inference for unknown functionals}
	
	We now move on to the inference for the unknown functional $\gamma_0:=
	\mathbb{E}l(h_0(W_t))$, which is estimated by $\widehat\gamma$ as defined in
	(\ref{eqa4.1}). Consider testing
	\begin{equation*}
	H_0: \mathbb{E}l(h_0(W_t))=\phi_0
	\end{equation*}
	for some known $\phi_0$. 
	Define
	\begin{equation*}
	L_n(\alpha,\gamma):= Q_n(\alpha)+ (\widehat\phi(\alpha)-\gamma) ^2
	\widehat\Sigma_2^{-1},
	\end{equation*}
	where $\widehat\Sigma_2$ consistently estimates the long-run variance (e.g. \cite{NW87}):  $$\Sigma_2:=\var%
	\left( \frac{1}{\sqrt{n}}\sum_{t=1}^{n} \mathcal W_t\right)
	=\frac{1}{n}\sum_{t=1}^n\var(\mathcal W_t) +\frac{1}{n}\sum_{t\neq s} \text{cov}(\mathcal W_t, \mathcal W_s)
	.$$
	We recall that 
	$ \mathcal{W}_t= l(h_0(W_t)) -
	\Gamma(X_t) \rho(Y_{t+1}, \alpha_0)$. 
	
	Note that $(\widehat\alpha,\widehat\gamma)$ is
	numerically equivalent to the solution to the following problem:
	\begin{equation*}
	L_n(\widehat \alpha, \widehat\gamma)+\lambda_n P_{en}(\widehat h)\leq
	\inf_{\alpha\in\mathcal{A}_n}\min_\gamma L_n(\alpha,\gamma)+\lambda_n
	P_{en}(h)+o_P(n^{-1}).
	\end{equation*}
	We define the   GN-QLR statistic as
	\begin{equation*}
	\widetilde S_n(\phi_0)= n\left( L_n(\widehat\alpha^R,\phi_0)
	-L_n(\widehat\alpha,\widehat\gamma) \right),
	\end{equation*}
	where $\widehat\alpha^R\in\mathcal{A}_n^R$ approximately minimizes the
	penalized loss function in the learning space $\mathcal H_n$, but fixing $\gamma=\phi_0$:
	\begin{equation*}
	L_n(\widehat \alpha^R, \phi_0)+ \lambda_n P_{en}(\widehat h^R) \leq
	\inf_{\alpha\in\mathcal{A}_n} L_n( \alpha , \phi_0) +\lambda_n
	P_{en}(\alpha)+o_P(n^{-1}).
	\end{equation*}
	
	
	The asymptotic analysis of $\widetilde S_n(\phi_0)$ is rather sophisticated,
	which requires additional rate constraints stated as follows.
	
	\begin{thm}
		\label{thqlr:eff}   Suppose  $
		\widehat\Sigma_2-\Sigma_2=o_P(1)\Sigma_2$ and conditions of Theorem \ref%
		{th4.2} hold. Then under $H_0: \gamma_0=\phi_0$
		\begin{equation*}
		\widetilde S_n(\phi_0)\to^d\chi^2_1.
		\end{equation*}
	\end{thm}
	

	\section{Examples}
	
	\label{sec:examples}

	In this section, we illustrate our main results using three important models:
Reinforcement learning,	NPIV and NPQIV. We impose premitive conditions to verify the high level   Assumptions \ref{ass2.1}, \ref{ass3.1} and \ref{ass4.2} respectively in the two models.  
	
	\subsection{Reinforcement learning}

	Reinforcement learning (RL) has been an important learning device behind  many successes in applications of  artificial intelligence. 
	Theories of RL have been developed in the literature of statistical learning and computer science. Most of the existing theoretical works formulate the problem as a least-square regression and approximate the value function by a linear function, such as \cite{bradtke1996linear}, etc.  
Nonlinear approximations using kernel methods or deep learning  appeared in the more recent literature, for example \cite{farahmand2016regularized, geist2019theory, fan2020theoretical, duan2021optimal, long20212,chen2022well}.  \cite{shi2020statistical} also conducted inference for the optimal policy  using  linear sieve representations.  


	
	We proceed learning using neural networks, and study the inference for a \textit{given} policy.	We follow the recent literature on the \textit{off-policy evaluation} problem, and formulate the  reinforcement learning problem as a conditional moment restriction model.  Assume the observed data trajectory $\{(S_t, A_t, R_t)\}_{t \ge 0}$ is obtained from an unknown behavior policy probability  $\pi^b(a|s)$, where $(S_t, A_t, R_t)$ denote the state, action and observed reward at time $t$ respectively and $\pi^b(a|s)$ is the distribution to take action $a$ at state $s$.  We denote the space of states and actions as $\mathcal S$ and $\mathcal A$. 	It is assumed that the  reward $R_t$  is jointly  determined by $(S_t, A_t, S_{t+1})$. Standing at state $S_t$ at period $t$, one takes action $A_t$ and receives reward $R_t$. The state then transits to $S_{t+1}$ at the next period.

	The value of a given policy $\pi$ is measured by the so-called $Q$-function. Specifically, for any given $\pi$ and any state-action pair $(s,a)$, $Q$-function is defined as the expected discounted reward: 
	\begin{equation*}
	Q^\pi(s,a) = \sum_{t=0}^{\infty} \gamma^t \mathbb E^\pi (R_t| S_0 = s, A_0 = a)\,,
	\end{equation*}
	where $\mathbb E^\pi$ or in short $\mathbb E$ is the expectation when we take actions according to $\pi$, $0\le \gamma < 1$ is the discount factor and we consider the discounted infinite-horizon sum of expected rewards. To estimate $Q^\pi$, a classical approach is to solve the Bellman equation below:
	\begin{equation*}
	Q^\pi(s,a) = \mathbb E \bigg[R_t + \gamma \int_{x \in \mathcal A} \pi(x|S_{t+1}) Q^\pi(S_{t+1}, x) \mathrm{d} x \bigg | S_t = s, A_t = a \bigg]\,.
	\end{equation*}
	The goal is to recover $Q^\pi$ of a given target policy $\pi$. 
 In practice,   multiple trajectories $\{(S_{i,t}, A_{i,t}, R_{i,t}, S_{i,t+1})\}_{0\le t \le T,1\le i\le N}$ may be observed to help estimate the $Q$-function. But for simplicity we assume $N=1$ and $T=n$. The more general case can be cast by merging the $N$ time series into a single series of size $n=TN. $

The Bellman equation can be  formulated as a conditional moment restriction with respect to $Q^{\pi}$ for weakly dependent time series:
	$$
	\mathbb E[\rho(Y_{t+1}, Q^{\pi})| S_t, A_t]=0,\quad Y_{t+1}= (R_t, S_t, A_t, S_{t+1}), \quad  X_t=(S_t, A_t),
	$$
	where
	$$
	\rho(Y_{t+1}, h)=R_t- h(S_t,A_t) + \gamma \int_{x \in \mathcal A} \pi(x|S_{t+1}) h(S_{t+1}, x) \mathrm{d} x.
	$$

	In this  framework, the  estimation of the function $Q^{\pi}(s,a)$ can be conducted on the neural network space, and we assume that computationally the integration in the $\rho$-function can be well approximately by the Monte Carlo method.  For off-policy evaluations, the following \textit{value function} is of  the major  interest in this section: given state $s\in\mathcal S$, 
	\begin{equation}\label{eq5.1functphi}
	\phi_s(Q^\pi) =    \int_{a \in \mathcal A} \pi(a|s) Q^\pi(s, a) \mathrm{d} a ,
	\end{equation}	 
	which is a known functional $\phi_s(\cdot)$  for a single state $s$. 
	
	The Bellman equation also admits a Fredholm integral equation of the second kind \citep{kress1989linear}, which is a well-posed problem.  Therefore, estimating the $Q$-function may achieve fast-rate of convergence. That is, the  
sieve modulus
		of continuity satisfies: 
	\begin{equation*}
	\omega_n(\delta):= \sup_{\alpha\in\mathcal{A}_n:
		\|\alpha-\pi_n\alpha_0\|\leq\delta} \|\alpha-\pi_n\alpha_0\|_s\asymp \delta
	\end{equation*}
Recently 	\cite{chen2022well}  showed this result for $\|.\|_s$ to be either the sup-norm or the $\ell_2$-norm. 	The inner product is defined, in this case, as
	$
	\langle v_1, v_2\rangle=\mathbb{E}\Sigma(X_t)^{-1}\left( \frac{dm}{d h}[v_1] \right)\left( \frac{dm}{dh}%
	[v_2] \right), $
	where 
	\begin{equation}\label{eq6.2mdq}
	  	\frac{dm}{dh}[v] = \gamma \int_{x \in \mathcal A} \mathbb E\left[\pi(x|S_{t+1}) v(S_{t+1}, x) | S_t, A_t\right]\mathrm{d} x
	- v(S_t, A_t),
	\end{equation}
	and induced a Riesz representer $v^*$ whose closed form is unavailable. Meanwhile, it follows from the Bellman equation that
	$
	m(X_t, h) =	\frac{dm}{dh}[h-Q^{\pi}] 
	$  for all $h\in\mathcal H_n$. Therefore,  
	the weak norm $\|.\|$ can be expressed as:
	$$
	\|h-Q^{\pi}\|^2= \mathbb E m(X_t, h)^2\Sigma(X_t)^{-1},
	$$
	which shows that the employed minimum distance criterion function is directly estimating the squared weak norm.

	Let $\widehat Q^{\pi}$ be the estimated  $Q^{\pi}$ using the general nonlinear learning space, and the functional is naturally estimated using 
	$$
	\phi_s(\widehat Q^{\pi})=  \int_{a \in \mathcal A} \pi(a|s) \widehat Q^{\pi}(s, a) \mathrm{d} a
	$$
	As the moment restriction function
	$ \mathbb E [	\rho(Y_{t+1}, h)| S_t, A_t] $ is linear in $h$ in this case, it is straightforward to verify  the high-level conditions as follows. 
	
\begin{assum}\label{ass6.1rl}

(i) For some $\zeta>4$, the Riesz representer satisfies

$	\mathbb E\int \pi(x|S_{t+1})|v^*_n(S_{t+1}, x)|^{\zeta}dx+ \mathbb E |v_n^*(S_t, A_t)|^{\zeta}\leq \|v_n^*\|^{\zeta}$.

(ii)  	$\mathbb ER_t^4<\infty $, 	$\mathbb E\max_{j\leq k_n} \Psi_j(X_t)^4<\infty$, 
	 $\mathbb E    (1+|S_t|^2+|A_t|^2)^{2\omega}<\infty 
	 $
	  and $\mathbb E M(S_{t+1})^4<\infty$,
	where 
	$M(S_{t+1}):=\int\pi(x|S_{t+1}) (1+x^2+S_{t+1}^2)^{\omega/2}dx$, and $\omega$ is the degree of the weighted-sup metric $\|.\|_{\infty,\omega}$.

	\end{assum}

	\begin{prop}
		\label{prop5.0RL} For the Reinforcement Learning model considered here, 
	  Assumption \ref{ass6.1rl} implies Assumptions \ref{ass2.1}, \ref{ass3.1} and  \ref{ass4.2}.

	\end{prop}

	It then follows from Theorem \ref{th4.1} that 
	$$\|v_n^*\|^{-1}\sqrt{n}\left(\phi_s(\widehat Q^{\pi}) -\phi_s(Q^{\pi})\right)\to^d\mathcal N(0,1)
	$$ Inference about $\phi_s(Q^{\pi})$ based on pivotal statistics can be conducted using the GN-QLR test.  


	\subsection{The NPIV model} \label{sec6.2}
	
In the nonparametric instrumental variable model (NPIV), 	consider 
	\begin{equation*}
	y_{t+1}= h_0(W_t)+U_{t+1},\quad \mathbb{E}(U_{t+1}|\sigma_t(\mathcal{X}))=0.
	\end{equation*}
	where $\sigma_t(\mathcal X)$ is the filtration generated from   instrumental variables $X_t$.
	Then $m(X_t, \alpha)= \mathbb{E}[(y_{t+1} - h(W_t))|\sigma_t(\mathcal{X})] $
	and the Gateaux derivative is defined as $\frac{dm(X_t, \alpha)}{dh}[v] =
	\mathbb{E}(v(W_t)|\sigma_t(\mathcal{X})), $ implying
	\begin{equation*}
	\langle u_n, h-h_0\rangle= \mathbb{E}\left[\mathbb{E}(u_n(W_t)|\sigma_t(%
	\mathcal{X})) \mathbb{E}(h-h_0|\sigma_t(\mathcal{X}))\Sigma(X_t)^{-1} \right]%
	.
	\end{equation*}
	We estimate the conditional variance $%
	\Sigma(X_t)$ by $
	\widehat\Sigma_t = \widehat A_n^{\prime }\Psi_n(\Psi_n^{\prime
	}\Psi_n)^{-1}\Psi(X_t) $
	where $\widehat A_n$ is a $n\times 1$ vector of $\rho(Y_{t+1}, \widehat
	\alpha)^2$.	Recall that for $\delta_n$ and $\bar\delta_n$ defined in (\ref{def_dn_deltan}), 
	\begin{equation*}
	\|\widehat h-h\|_{\infty,\omega}=O_P(\delta_n),\quad \|\widehat h-h\|=O_P(\bar\delta_n).
	\end{equation*}
	We impose the following low-level conditions to verify Assumptions \ref%
	{ass3.1} and \ref{ass4.2}.
	
	\begin{assum}
		\label{as5.1}  (i) $\delta_n^2\bar\delta_n\sigma= o(n^{-1})$,  $\mathbb{E }\max_{j\leq k_n}|\Psi_j(X_t) |^2 (U_t^2+1)<C $,  and $\mathbb E(U_t^2|\sigma_t(\mathcal X))<C$ almost surely. Also, $\mathbb E(1+|W_t|^2)^{\omega}<C$ and $\mathbb E\max_{j\leq k_n}\Psi_j(X_t)^2(1+|W_t|^2)^{\omega}<C.$

 (ii) The Riesz representer $v_n^*$  satisfies: there are $C, \zeta>0$, 
 
			$\mathbb E(\max_{j\leq k_n}\Psi_j(X_t)^2+1)v_n^*(W_t)^2< C \mathbb EK_t^2$
		and 
$	 \mathbb E |  U_t|^{2+\zeta}| K_t|^{2+\zeta} \leq C (\mathbb EK_t^2)^{1+\zeta/2}$, 
	 where $K_t:=\mathbb E (v_n^*(W_t)|\sigma_t(\mathcal X)).$
	\end{assum}

	\begin{prop}
		\label{prop5.1} For the NPIV model,
		
		(i) Assumption \ref{as5.1} implies Assumptions  \ref{ass2.1}, \ref{ass3.1}, \ref{ass4.2}
		and \ref{a5.555}.
		
		(ii) For the known functional $\phi(\cdot)$, in addition Assumptions %
		\ref{ass3.10}, \ref{asfda.4}, \ref{ass3.2}, \ref{fa.45sass}, \ref{a34}, \ref{asssmoothphi}, \ref{ass4.3new}, \ref{ass4.4} hold. Then
		\begin{eqnarray*}
			\frac{\sqrt{n}(\phi(\widehat\alpha)-\phi(\alpha_0))}{\sigma_n } \to^d%
			\mathcal{N}(0,1),
		\end{eqnarray*}
		where $\sigma_n^2:= \var\left ( \mathbb{E}(v_n^*(W_t)|\sigma_t(\mathcal{X}))
		\Sigma(X_t)^{-1} U_t \right) .$
		
		(iii) For the unknown functional $\gamma_0= \mathbb{E }l(h_0(W_t)),$ if
		additionally Assumptions \ref{a5.1},\ref{a4.55} hold, then
		\begin{equation*}
		\sqrt{n}v^{-1}( \widehat\gamma-\gamma_0 ) \to^d\mathcal{N}(0,1),
		\end{equation*}
		where $v^2:=\frac{1}{n}\var(\sum_t\mathcal W_t- \mathcal Z_t) $ with $\mathcal W_t=l(h_0(W_t))-\Gamma_0(X_t)U_{t+1}$ and $\mathcal Z_t= U_{t+1}\Sigma(X_t)^{-1}\mathbb E [v_n^*(W_t)|\sigma_t(\mathcal X)]$.
		
	\end{prop}

	\subsection{The NPQIV model} \label{sec6.3}
	
	Consider the nonparametric quantile instrumental variable  (NPQIV) model
	\begin{equation*}
	\mathbb{E}[1\{y_{t+1}\leq h_0(W_t)\}|\sigma_t(\mathcal{X})]=\varpi\in(0,1).
	\end{equation*}
	Then $m(X_t, \alpha)= P(U_{t+1}<h-h_0|\sigma_t(\mathcal{X}))-\varpi$ where $%
	U_{t+1}=y_{t+1}- h_0(W_t)$ and $\alpha=h$. Within this framework, we now verify the high-level assumptions presented in the previous sections.
	
	Suppose the conditional distribution of $U_t$ given $%
	(X_t, W_t)$ is absolutely continuous with density function $f_{U_t|\sigma_t(%
		\mathcal{X}), W_t}(u)$. In this context, $\Sigma(X_t)$ is known, given by
	\begin{equation*}
	\Sigma(X_t)= \var(1\{y_{t+1}\leq h_0(W_t)\}|\sigma_t(\mathcal{X}))=
	\varpi-\varpi^2.
	\end{equation*}
	Then the Gateaux derivative is defined as
	\begin{equation*}
	\frac{dm(X_t, \alpha)}{dh}[v] = \mathbb{E}(f_{U_t|\sigma_t(\mathcal{X}),
		W_t}(h(W_t)-h_0(W_t))v(W_t)|\sigma_t(\mathcal{X})),
	\end{equation*}
	implying, for $g_1= f_{U_t|\sigma_t(\mathcal{X}),W_t}(0) u_n(W_t) $ and $%
	g_2=f_{U_t|\sigma_t(\mathcal{X}),W_t}(0) (h(W_t)-h_0(W_t)) $,
	\begin{equation*}
	\langle u_n, h-h_0\rangle= \mathbb{E}\left[\mathbb{E}(g_1|\sigma_t(\mathcal{X%
	})) \mathbb{E}(g_2|\sigma_t(\mathcal{X})) \right] (\varpi-\varpi^2)^{-1} .
	\end{equation*}
	Also, $\|v_n^*\| ^2=(\varpi-\varpi^2)^{-1}\mathbb{E }g(X_t)^2 $ where $%
	g(X_t)= \mathbb{E }[ f_{U_t|\sigma_t(\mathcal{X}),W_t}(0)
	v^*_n(W_t)|\sigma_t(\mathcal{X})]. $
	
	We impose the following low-level conditions to verify Assumptions \ref%
	{ass3.1} and \ref{ass4.2}. Let
	\begin{eqnarray*}
		A_t( v)&:=& \int_0^1 f_{U_t|\sigma_t(\mathcal{X}), W_t}\left( x(v(W_t)-
		h_0(W_t)) \right)dx\cr B_t(v,h)&:=&\mathbb{E }\left\{ A_t( v) [ h(W_t)-
		h_0(W_t)] |X_t\right\}.
	\end{eqnarray*}
	
	\begin{assum}
		\label{ass6.2}
		
		(i) There are $c_1, c_2, \epsilon_0>0$ so that for all $\|h-h_0\|_{\infty,\omega}<%
		\epsilon_0$,
		\begin{equation*}
		c_2\mathbb{E }B_t(h ,h)^2 \Sigma(X_t)^{-1}\leq \mathbb{E }B_t(h_0,h)^2
		\Sigma(X_t)^{-1}\leq c_1 \mathbb{E }B_t(h ,h)^2 \Sigma(X_t)^{-1} .
		\end{equation*}
		
		(ii) Almost surely, $\sup_{u} f^{^{\prime }}_{U_t|\sigma_t(\mathcal{X}),W_t}
		(u) <C$ and $\sup_{u, x,w} f_{U_t|\sigma_t(\mathcal{X}),W_t} (u) <C.$ Also
		and there is $L>0$, for all $u$, almost surely, $\sup_{x, w}|f_{U_t|\sigma_t(%
			\mathcal{X}),W_t}(u)-f_{U_t|\sigma_t(\mathcal{X}),W_t}( 0)|\leq L |u|. $
		
		(iii) $\mathbb{E}[\max_{j\leq k_n}\Psi_j(X_t)^2+A_t(h_0)^2](1+|W_t|^2)^{\omega}<C $ and $ \mathbb{E }%
		[u_n(W_t)^4|\sigma_t(\mathcal{X}) ]<C$.
		
		(iv) $\delta_n^2k_n=o(1)$ and $\delta_n^4=o(n^{-1})$. 
	\end{assum}

	The following proposition, proved in the appendix, is the main result in this subsection, which verifies the high-level conditions in the NPQIV context. 
	
	\begin{prop}
		\label{prop5.2} For the NPQIV model,
		
		(i) Assumption \ref{ass6.2} implies Assumptions \ref{ass2.1}, \ref{ass3.1}, \ref{ass4.2}
		and \ref{a5.555}.
		
		(ii) For the known functional $\phi(\cdot)$, in addition Assumptions %
		\ref{ass3.10}, \ref{asfda.4}, \ref{ass3.2}, \ref{fa.45sass}, \ref{a34}, \ref{asssmoothphi}, \ref{ass4.3new}, \ref{ass4.4} hold. Then
		\begin{eqnarray*}
			\frac{\sqrt{n}(\phi(\widehat\alpha)-\phi(\alpha_0))}{\sigma_n } \to^d%
			\mathcal{N}(0,1),
		\end{eqnarray*}
		where $\sigma_n^2:=(\varpi-\varpi^2)^{-1}\mathbb{E}\left([\mathbb{E }%
		f_{U_t|\sigma_t(\mathcal{X}),W_t}(0) v^*_n(W_t)|\sigma_t(\mathcal{X}%
		)]^2\right) .$
		
		(iii) For the unknown functional $\gamma_0= \mathbb{E }l(h_0(W_t)),$ if
		additionally Assumptions \ref{a5.1},\ref{a4.55} hold, then
		\begin{equation*}
		\sqrt{n}v^{-1}( \widehat\gamma-\gamma_0 ) \to^d\mathcal{N}(0,1),
		\end{equation*}
		where  $v^2:=\frac{1}{n}\var(\sum_t\mathcal W_t - \mathcal Z_t) $ with $\mathcal W_t=l(h_0(W_t))-\Gamma_0(X_t)U_{t+1}$ and \\$\mathcal Z_t= (\varpi-\varpi^2)^{-1}1\{U_{t+1}\leq0\} \mathbb{E }%
		f_{U_t|\sigma_t(\mathcal{X}),W_t}(0) v^*_n(W_t)|\sigma_t(\mathcal{X}%
		)$.
	\end{prop}
	
	\section{Simulation Studies}
	
	\label{sec:MC}
	
	In this section, we set up nonparametric endogenous models to illustrate the
	performance of our proposed estimators and testing statistics using some
	synthetic data.
	Consider the following data generating process
	\begin{equation*}
	Y_t = h(Z_t, Y_{t-1}, \dots, Y_{t-L}) + e_t \,,
	\end{equation*}
	where
	\begin{equation*}
	h(Z_t, Y_{t-1}, \dots, Y_{t-L}) = Z_t \vartheta_0 + f(\sum_{l=1}^{L} b_l
	Y_{t-l})\,,
	\end{equation*}
	and $\phi(\alpha) = \mathbb E[\partial h/\partial Z_t]=\vartheta_0 =1$ is the quantity
	to be estimated. We choose $L=3, b_l = 0.4^l$ and consider the nonlinear
	mapping $f(x) = \frac{1-\exp(-x)}{1+\exp(-x)}$. The endogenous  	$Z_t$ is generated using the following auto-regressive model:
	\begin{equation*}
	Z_t = 0.3 Z_{t-1} + u_t, \quad (u_t, \varepsilon_t) \sim_{iid} N(0, \Sigma),
	\quad \Sigma =
	\begin{pmatrix}
	1 & \rho \\
	\rho & 1%
	\end{pmatrix}%
	\,.
	\end{equation*}
	And $e_t$ is generated with the following ARCH model using $\varepsilon_t$
	as the innovation:
	\begin{equation*}
	e_t = \sigma_t \varepsilon_t, \quad \sigma_t^2 = 0.5 + 0.5 (1-0.3^2)
	Z_{t-1}^2\,.
	\end{equation*}
	We set $\rho = 0.5$ to make $Z_t$ endogenous. We also make $e_t$
	heterogeneous. Note that $\mathbb E[e_t^2] =
\mathbb 	E[\sigma_t^2] = 1$.  The endogenous variable is $W_t= Z_t$. The instruments are $X_t= (Z_{t-1}, Y_{t-1},...,Y_{t-L})$.
	We chose to generate $n = 5000$ samples (some burning period has been thrown
	away to make sure data are stationary). Note that the model can be used for
	both NPIV and NPQIV with $\varpi = 0.5$.
	
	We applied a fully-connected $J$-layer ReLU-activated NN with hidden layer
	width of $K$. The optimization of the unconstrained NPIV or NPQIV objective
	used vanilla gradient descent. We did not apply mini-batch in gradient
	descent training as using mini-batches may hurt performance due to
	insufficient smoothing. The training epoch was as large as $10000$ with
	learning rate $0.01$ for NPIV and $0.1$ for NPQIV. Furthermore we did not apply any penalty term for this
	example since the problem is relatively easy and the NN under consideration
	is of a small scale. The linear sieve bases $(\Psi_1,...,\Psi_{k_n})$ for the instrumental variable
	space were $\tilde k_n$ cubic B-splines for $X$ and each of the three $Y$ lags concatenated together. For simplicity, no interaction terms between X and Y lags were included. Thus in total, we have $k_n = 4 \tilde k_n - 3$ bases (since all B-spline bases sum up to 1, we remove the last basis for each dimension and finally add the intercept term as another basis).
	In our simulations, we find that NPQIV requires more number of sieve basis $k_n$ for estimating the instrumental space.
	
	For the  NPIV problem, we first optimize the equal weighted quadratic loss to
	obtain $\widehat h$, which is used to estimate $\Sigma(X_t)$ and $%
	\Gamma(X_t) $ consistently. In the second step, we optimize the optimally
	weighted quadratic loss with the weighting matrix $\widehat\Sigma(X_t)^{-1}$
	and apply the forward filter to estimate our expectation functional, which
	in this example is the constant $\vartheta_0=1$. Finally, we carry out the hypothesis
	testing for $H_0: \phi(h) =\mathbb E[\partial h/\partial Z_t] = \phi_0 = 1$ to
	check the size of the testing statistic. Specifically, we estimated the
	forward filtered residuals as $\widehat{\mathcal{W}}_t= \partial\widehat
	h(W_t)/\partial W_t - \widehat\Gamma_t (Y_t - \widehat h(W_t))$ and estimated $%
	\Sigma_2 = \var(\mathcal{W}_t)$ by the Newey-West estimator given $\widehat{%
		\mathcal{W}}_t$, then solved the constrained optimization of $L_n(h,
	\phi_0)$ and finally constructed the testing statistic. For NPQIV problem,
	since the optimal weighting is proportional to equal weighting, we do not
	need the initial step to estimate $\Sigma(X_t)$. So we directly optimized the
	optimally weighted quadratic loss and estimated $\Gamma(X_t)$ using the
	results and then used the forward filter to correct the estimation of the average
	partial derivative. Finally, similar to NPIV, we conduct the hypothesis
	testing for $H_0: \phi(\alpha) = 1$ under NPQIV.
	
	\begin{table}[]
		\caption{Estimation and hypothesis testing under NPIV and NPQIV with
			synthetic data. Here $(J,K, k_n)$ respectively denote the number of layers, width of the neural nets and the number of sieve bases for estimating the instrumental space. The true value for $\vartheta_0=1$. 95\% qtl refers to the  empirical 95\% quantile, where the theoretical quantile for the chi square distribution is 3.84. \\}
		\label{tab1}\centering
		\begin{tabular}{cccccccccc}
			\hline
			\hline
			& Layer & Width &Basis& \multicolumn{2}{c}{Estimator of $\vartheta_0$} & \multicolumn{4}{c}{Testing Statistic for $\vartheta_0$} \\ \cline{5-10}
			Problem & $J$ & $K$ & $k_n$ & mean & std & mean & std & 95\% qtl &
			size \\ \hline
			NPIV & 3 & 10 & 17 & 0.968 & 0.116 & 0.999 & 1.432 &  3.814 &
			5.0\% \\
				NPIV & 3 & 10 & 13 & 0.957 & 0.115 & 0.874 & 1.236 &  3.727 &
			4.8\% \\
				NPIV & 1 & 40 & 13 & 0.984 & 0.108 & 1.032 & 1.418 & 4.215 &
			6.0\% \\
					&  &  & & &  &  & &   &
			 \\
			NPQIV & 3 & 10 & 49 & 0.997 & 0.129 & 1.086 &  1.565 &  4.280  & 6.4\% \\
			NPQIV & 3 & 10 & 45 & 0.994 & 0.130 & 1.002 &  1.409 &  3.955 & 5.6\% \\
			NPQIV & 1 & 40 & 29 & 0.977 & 0.126 & 1.050 & 1.421 & 3.678 & 4.9\% \\
			\hline
		\end{tabular}%
	\end{table}

	As for the computational practice, we find that for NPQIV models, it is helpful to apply   truncations to the  learned gradients in each step  of training the network. Specifically, we smooth the loss function of the NPQIV model and  truncate the  updated gradient:
	$$
	\theta_{k+1}= \theta_k - \text{lr} *\min\{|\nabla L_{n,k}|, 0.001\}*\text{sgn}(\nabla L_{n,k})
	$$
	where lr is the learning rate, fixed to be 0.1 for NPQIV;  $\nabla L_{n,k} $ is the gradient of the NN at the current step; $\theta_{k+1}$ is the updated neural network coefficients at the current step. The truncation prevents the network from having very large gradients during iterations,  helping stabilize the training process empirically.

	We repeat each setting for $1000$ times. For the efficient estimation, we
	report the mean and standard deviation of 
	the forward
	filtered average gradient 
	for the optimal weighting
	optimizaiton in Table \ref{tab1}. For hypothesis testing, we also report in
	Table \ref{tab1} the mean, standard deviation and 95\% quantile of the
	empirical testing statistic. In addition, if we use the theoretical critical
	value corresponding to 5\% significance level, which is 3.84 for $\chi^2_1$,
	the p-value is also reported.

	As we can see from Table \ref{tab1}, for NPIV, optimal weighting estimates $%
	\phi(h)$ accurately in the sense that the mean insignificantly differs from the true value $\vartheta_0 = 1$. NPQIV is less efficient with a larger standard deviation, and thus requires more samples to be estimated to the same accuracy. Note that the instrumental
	space with a step function can be harder to approximate with the cubic B-spline linear
	sieve bases. In terms of the performance of QLR testing
	statistic, the p-values are all close to the nominal  5\% level for the NPIV and NPQIV models. Admittedly through our experiments the results can be sensitive to some tuning parameters, which is typically the case when applying deep learning for statistical inference: at the moment we still heavily rely on ad-hoc tuning in many problems. In comparison, the
	estimation of $\phi(h)$ is more stable with respect to different $J$
	and $K$ values.
	Here we only mean to present some results without heavily tuning the
	parameters. Methods using NN for real applications require more extensive tuning
	in practice and some rough sense on the model
	complexity would be useful to determine the balance between the dimensions of the NN sieve
	and the linear IV sieve.
	
	\section{Conclusion}
	
	\label{sec:conclusion} In this paper we establish neural network estimation
	and inference on functionals of unknown function satisfies a general time
	series conditional moment restrictions containing endogenous variables. 
	We consider quasi-likelihood		ratio (GN-QLR) based inference, where nonparametric functions are learned using multilayer neural networks. While the asymptotic normality of the estimated functionals depends on some unknown Riesz representer of the functional space, we show that the GN-QLR statistic is  asymptotically 		Chi-square distributed, 	regardless whether the expectation functional is regular (root-$n$ estimable) or not. This holds when the data  are  weakly dependent and satisfy the beta-mixing condition.

	In addition to estimating partial derivatives in nonparametric endogenous problems as examples, 	our study is well motivated by the setting of reinforcement learning where data are time series in nature. We apply our method to the off-policy evaluation, by formulating the Bellman equation into the conditional moment restriction framework, so that we can make inference about the state-specific value functional using the proposed GN-QLR method  with time series data. 
	
	\newpage
	
	\appendix

	\section{Stochastic equicontinuity on the learning space for $\protect\beta$-mixing observations}
		

	{\small A key technical result is the stochastic equicontinuity of the
		residual function on the  general nonlinear sieve learning space, which is established in the following
		proposition in this section.  Let   $S_t=(Y_{t+1}, X_t)$ and $$
		\epsilon_t(\alpha) \equiv \epsilon(S_t, \alpha):=\rho(Y_{t+1}, \alpha)-m(X_t, \alpha).$$ 
		We derive bounds that require the \textit{pseudo dimension} of the deep
		neural network class.  Recall
		\begin{equation*}
		\delta_n:=\|\pi_n\alpha_0-\alpha_0\|_{\infty,\omega}+\omega_n(\bar\delta_n),\quad
		\bar\delta_n^2:= \|\pi_n\alpha_0-\alpha_0\|^2+ \lambda_n+k_nd_n^2+\varphi_n^2
		\end{equation*}
		where $d_n:=\sqrt{\frac{ p(\mathcal{H}_n) \log^2 n}{n}}$.
	}
	
	\begin{prop}
		{\small \label{lem5.2fd} Let $\mathcal{C}_n=\{\alpha+xu_n: \alpha\in\mathcal{%
				A}_n, \|\alpha-\alpha_0\|_{\infty,\omega}\leq C\delta_n,  Q(\alpha)\leq
			C\bar\delta_n^2 , |x|\leq Cn^{-1/2}\}$. Suppose : }
		
		{\small (a) $\mathbb{E}\max_{j\leq k_n}\Psi_j(X_t)^2\sup_{\alpha\in\mathcal{C%
				}_n} (\rho(Y_{t+1}, \alpha)-\rho(Y_{t+1}, \alpha_0) )^2=C\delta_n^{2\eta} $
			for some $\eta, C>0.$ }
		
		{\small (b) For some $\kappa, C>0$ , $\mathbb{E }\Psi_j(X_t)^2\sup_{
				\|\alpha_1-\alpha\|_{\infty,\omega}<\delta}
			|\epsilon_t(\alpha_1)-\epsilon_t(\alpha)|^2\leq C\delta^{2\kappa} $ for all $%
			\delta>0$ and $\alpha,\alpha_1\in cl\{a+xb: a, b\in\mathcal{A}_n, x\in%
			\mathbb{R}\}$. }
		
		{\small Then
			\begin{equation*}
			\max_{j\leq k_n}\sup_{|x|\leq Cn^{-1/2}}\sup_{\alpha\in\mathcal{C}_n} |\frac{%
				1}{n}\sum_t\Psi_j(X_t)
			(\epsilon(S_t,\alpha+xu_n)-\epsilon(S_t,\alpha_0))|\leq O_P(d_n
			\delta_n^{\eta}).
			\end{equation*}
		}
	\end{prop}
	
	{\small
		\begin{proof}
			

			Let $\mathcal E:=\{ (\epsilon(, \alpha+xu_n)-\epsilon(, \alpha_0)   )\Psi_j : \alpha\in\mathcal C_n, j\leq k_n, |x|\leq Cn^{-1/2}\}$ and let $S_t=(Y_{t+1}, X_t)$. We divide the proof into several steps.
			
			\textbf{Step 1: construct blocks.} Consider the following independent blocks: for any integer pair $(a_{n}, b_{n})$, with $b_{n}=[n/(2a_{n})]$, divide $\{S_t:t\leq n\}$ into $2b_n$ blocks with length $a_n$ and the remaining block of length $n-2a_nb_n$:
			\begin{eqnarray*}
				H_{1,l}&=&\{i: 2(l-1)a_n+1\leq i\leq (2l-1)a_n\}  \cr
				H_{2,l}&=&\{i: (2l-1)a_n+1\leq i\leq 2la_n\}
			\end{eqnarray*}
			where $l=1,...,b_n$. Let $\Upsilon=\{i: 2a_nb_n+1\leq i\leq n\}$. Now let $\{\widetilde S_1,...,\widetilde S_n\}$ be a random sequence that is independent of $\{S_1,..,S_n\}$ and has independent blocks such that each block has the same joint distribution as the corresponding block of the $S_t$-sequence. Because the $S_t$-sequence is $\beta$-mixing, by Lemma 2 of \cite{eberlein1984weak}, for any measurable set $A$, with the mixing coefficient $\beta()$,
			\begin{equation}\label{eqb.1ddaa}
			|P\left(\{\widetilde S_t: t\in H_{1,l}, l=1,...,b_n\}\in A\right) - P\left(\{  S_t: t\in H_{1,l}, l=1,...,b_n\}\in A\right) |\leq  (b_n-1) \beta(a_n).
			\end{equation}
			The same inequality holds when $H_{1,l}$ is replaced with $H_{2,l}$. In addition, for any function $f$,  define
			$$
			U_{1,f}(\widetilde S^l) =\frac{1}{a_n}\sum_{t\in H_{1,l}} f(\widetilde S_t),\quad U_{2,f}(\widetilde S^l) =\frac{1}{a_n}\sum_{t\in H_{2,l}} f(\widetilde S_t),
			$$
			where $\widetilde S^l=\{\widetilde S_t: t\in H_{1,l}\}$.
			By construction, $U_{1,f}(\widetilde S^l) $ and $U_{2,f}(\widetilde S^l) $ are independent across $l$.    Similarly, let   $  S^l=\{ S_t: t\in H_{1,l}\}$.   Then
			\begin{eqnarray}\label{eqb.2adfad}
			\frac{1}{n}\sum_t f(S_t)-\mathbb E f(S_t) &=&\frac{1}{n}\sum_{t\in \Upsilon} f(S_t)-\mathbb E f(S_t)+ \frac{1}{b_n}\sum_{l\leq b_n} a_nb_nn^{-1} [U_{1,f}(  S^l) - \mathbb E  U_{1,f}(  S^l)  ] \cr
			&& + \frac{1}{b_n}\sum_{l\leq b_n} a_nb_nn^{-1} [ U_{2,f}(  S^l)  -\mathbb EU_{2,f}(  S^l) ] .
			\end{eqnarray}
			
			Next, we shall bound each term on the right hand side uniformly for $f\in\mathcal E.$ We replace $U_{1,f}(  S^l) $ with $U_{1,f}( \widetilde S^l) $; the latter is easier to bound because  blocks $\widetilde S^l$  are independent. We  then show that the effect of such   replacements is negligible due to (\ref{eqb.1ddaa}) by properly chosen $(a_n, b_n)$.

			\textbf{Step 2: the envelop function for $U_{1,f}$.}
			Note that $\mathbb Ef=0$ for $f\in\mathcal E$  and that $\widetilde S_t$ and $S_t$ are identically distributed within each  block $H_{1,l}$. 	By Cauchy-Schwarz,
			\begin{eqnarray*}
				\mathbb E\sup_{f\in\mathcal E} U_{1,f}(\widetilde S^l)^2&\leq&  \mathbb E\sup_{f\in\mathcal E} \left(\frac{1}{a_n}\sum_{t\in H_{1,l}} f(\widetilde S_t) \right)^2\leq  \frac{1}{a_n} \sum_{t\in H_{1,l}}\mathbb E \sup_{f\in\mathcal E}  f( S_t)   ^2\cr
				&\leq&
				2\mathbb E\max_{j\leq k_n}\Psi_j(X_t)^2\sup_{|x|\leq C n^{-1/2}}\sup_{\alpha\in\mathcal C_n} (\rho(Y_{t+1}, \alpha+xu_n)-\rho(Y_{t+1}, \alpha_0)   )^2\leq C \delta_n^{2\eta}.
			\end{eqnarray*}
			Now take some $p>\eta.$
			Let $\mathcal F=\{U_{1,f}: f\in\mathcal E\}$ and
			let  $ F:=\max\{ n^{-p}, \sup_{f\in\mathcal E}|U_{1,f}| \}$.  Then both $\sup_{f\in\mathcal E}|U_{1, f}| $ and $F$ are envelope functions of $\mathcal F$, and
			$$
			n^{-p}\leq G:=\|F\|_{L_2(S_t)}\leq Cn^{-p} + C\delta_n^\eta\leq2C \delta_n^{\eta}.
			$$

			\textbf{Step 3: the bracketing number.}
			We aim to apply Theorem 2.14.2 of \cite{VW} to bound $ \frac{1}{b_n}\sum_{l\leq b_n} a_nb_nn^{-1} U_{1,f}( \widetilde S^l) $, which requires bounding the bracketing number of $\mathcal F.$
			To do so,
			suppose $h_1,..., h_N$ is a $\delta$-cover of $\mathcal H_n$ under the norm $\|h\|_{\infty,\omega} $ and $N:= \mathcal N(\delta, \mathcal H_n, \|.\|_{\infty,\omega})$; $\theta_1,...,\theta_R$ is a $\delta$-cover of $\Theta$ and $R:=\mathcal N(\delta, \Theta, \|.\|)$ (the Euclidean norm in $\Theta$). Here $\mathcal N(\delta, \mathcal A,.)$ denotes the covering number for space $\mathcal A.$   Also let $x_1...x_{M_n}$ be a $\delta$-cover of $[-Cn^{-1/2}, Cn^{-1/2}]$, with $M_n\leq 4Cn^{-1/2}/\delta$.
			
			Then    for any $f=(\epsilon(,\alpha+xu_n)-\epsilon(, \alpha_0)   )\Psi_j \in\mathcal E$,  there are $\Psi_j$, $x_q$ and $\alpha_{ik}=(\theta_k, h_i)$ so that $\|\alpha-\alpha_{ik}\|_{\infty,\omega}\leq \|h-h_i\|_{\infty,\omega}+\|\theta-\theta_k\|\leq2\delta$ and $|x-x_q|<\delta$.    Let $f_{ijkq}=(     \epsilon(, \alpha_{ik}+x_qu_n)-  \epsilon(, \alpha_0)   )\Psi_j$.
			\begin{eqnarray*}
				&&  \sup_{f=(\epsilon(,\alpha+xu_n)-\epsilon(,\alpha_0)   )\Psi_j : \|\alpha-\alpha_{ik}\|_{\infty,\omega}<2\delta, |x-x_q|<\delta}  | U_{1,f}(\widetilde S^l)
				-  U_{1,f_{ijkq}}(\widetilde S^l)  |
				\cr
				& \leq&  \sup_{f=(\epsilon(,\alpha+xu_n)-\epsilon(,\alpha_0)   )\Psi_j : \|\alpha-\alpha_{ik}\|_{\infty,\omega}<2\delta, |x-x_q|<\delta}  | \frac{1}{a_n}  \sum_{t\in H_{1,l}} f(\widetilde S_t)-     f_{ijkq}(\widetilde S_t) |
				\cr
				&\leq&  \frac{1}{a_n}  \sum_{t\in H_{1,l}} |\Psi_j(\widetilde X_t)|\sup_{ \|\alpha-\alpha_{ik}\|_{\infty,\omega}<2\delta}\sup_{|x-x_q|<\delta}  | \epsilon(\widetilde S_t,\alpha+xu_n)- \epsilon(\widetilde S_t,\alpha_{ik}+x_qu_n)|  :=b_{ijkq}(\widetilde S^l,\delta).
			\end{eqnarray*}
			Then $U_{1,f}\in[l_{ijkq}, u_{ijkq}]$, where $l_{ijkq}:= U_{1, f_{ijkq}}- b_{ijkq}(,\delta)$ and $u_{ijkq}=U_{1, f_{ijkq}}+b_{ijkq}(,\delta)$.    In addition,
			\begin{eqnarray*}
				\mathbb E[u_{ijkq}-l_{ijkq}]^2&\leq&4\mathbb Eb_{ijkq}(\widetilde S^l,\delta)^2\cr
				&\leq&  C \mathbb E\left(  \frac{1}{a_n}  \sum_{t\in H_{1,l}} |\Psi_j(\widetilde X_t)|\sup_{|x-x_q|<\delta}\sup_{ \|\alpha-\alpha_{ik}\|_{\infty,\omega}<2\delta}  |   \epsilon(\widetilde S_t,\alpha+xu_n)-\epsilon(\widetilde S_t, \alpha_{ik}+x_qu_n)| \right)^2\cr
				&\leq& C \mathbb E\Psi_j(\widetilde X_t)^2\sup_{ \|\alpha-\alpha_{ik}\|_{\infty,\omega}<2\delta}\sup_{|x-x_q|<\delta}  |  \epsilon(\widetilde S_t  ,\alpha+xu_n)- \epsilon(\widetilde S_t, \alpha_{ik}+x_qu_n)| ^2\leq C\delta^{2\kappa}.
			\end{eqnarray*}
			Hence $\{[l_{ijkq}, u_{ijkq}]:i\leq N, j\leq k_n, k\leq R \}$ is a $C \delta^{\kappa}$ bracket of $\mathcal F$, whose bracketing number satisfies
			$$\mathcal N_{[]}( C\delta^{\kappa},\mathcal F, \|.\|_{L^2(\widetilde S_t)})\leq \underbrace{\mathcal N(\delta, \mathcal H_n, \|.\|_{\infty,\omega})}_{N}  \underbrace{(C/\delta)^d}_R \underbrace{(n^{-1/2}/\delta)}_{M_n}k_n, $$
			where we used $R\leq (C/\delta)^d$ for $d=\dim(\theta_0)$ since $\theta_0\in\Theta$ is compact.
			Then for a generic constant $C>0$,
			$$\mathcal N_{[]}( Gx,\mathcal F, \|.\|_{L^2(\widetilde S_t)})\leq C\mathcal N(x^{1/\kappa}(G/C)^{1/\kappa}, \mathcal H_n, \|.\|_{\infty,\omega})G^{-(d+1)/\kappa} x^{-(d+1)/\kappa}k_n,\quad \forall x>0.$$

			\textbf{Step 4: bound independent blocks.} Note that $U_{1,f}(\widetilde S^l)$ are independent across $l$ and is mean-zero. For the envelop $G$ defined in step 2 and some constant $\bar M>0$,
			\begin{eqnarray*}
				&& \mathbb E\sup_{f\in\mathcal E}\left|  \frac{1}{b_n}\sum_{l\leq b_n} a_nb_nn^{-1} U_{1,f}(  \widetilde S^l)\right| \leq  \frac{1}{2} \mathbb E\sup_{g\in\mathcal F}\left|  \frac{1}{b_n}\sum_{l\leq b_n} g(  \widetilde S^l)\right| \cr
				&\leq^{(i)}& b_n^{-1/2}G
				\int_0^1\sqrt{1+\log\mathcal N_{[]}(Gx, \mathcal F, \|.\|_{L_2(\widetilde S_t)})} dx    \cr
				&\leq &
				\frac{C }{\sqrt{b_n}} \delta_n^\eta    \int_0^1\sqrt{1+\log\mathcal N(x^{1/\kappa} (G/C)^{1/\kappa} , \mathcal  H_n, \|.\|_{\infty})+ \log \frac{C}{n^{1/2}G^{(d+1)/\kappa} x^{(d+1)/\kappa}}+\log k_n} dx    \cr
				&\leq^{(ii)}&\frac{C   \delta_n^\eta  }{\sqrt{b_n}}   \int_0^1\sqrt{1+p(\mathcal H_n)\log
					\frac{C n}{x^{1/\kappa} G^{1/\kappa} }+ (d+1)\log \frac{C}{G^{1/\kappa} x^{1/\kappa}}+\log k_n}dx     \cr
				&\leq^{(iii)}&\frac{C \delta_n^\eta  }{\sqrt{b_n}}   \int_0^1\sqrt{2\log k_n+2p(\mathcal H_n)\log
					\frac{C  n}{x^{1/\kappa} G^{1/\kappa} } }dx  \cr
				&\leq^{(iv)}  & \delta_n^\eta  \sqrt{\frac{Cp(\mathcal H_n)\log n}{b_n}}   ,
			\end{eqnarray*}
			where (i)   follows   from Theorem  2.14.2 of  \cite{VW};  (ii) follows  from Assumption \ref{assumcomplexity}; (iii) is due to $p(\mathcal H_n)\to\infty $ .
			
			We now prove the inequality (iv), which is to show $\int_0^1\sqrt{2\log k_n+g(x)}dx\leq C\sqrt{p(\mathcal H_n)\log n}$ where $g(x)= 2p(\mathcal H_n)\log
			\frac{Cn} { x^{1/\kappa} G^{1/\kappa} }$.
			Let $A:= 2\log k_n+2p(\mathcal H_n)\log
			\frac{Cn} {  G^{1/\kappa} } -2\kappa^{-1}  p(\mathcal H_n) $. We have $\log
			\frac{Cn} { G  }\to \infty$, hence   $2\kappa^{-1} p(\mathcal H_n)\leq A$.
			Note that $\log(y)\leq y-1$ for all $y>0$. Hence
			\begin{eqnarray*}
				2\log k_n+g(x)&=& 2\log k_n+ 2p(\mathcal H_n)\log
				\frac{C n} {  G^{1/\kappa} }+2\kappa^{-1}p(\mathcal H_n)\log
				\frac{1} { x  }
				\cr
				&\leq& 2\log k_n+ 2p(\mathcal H_n)\log
				\frac{C n} {  G^{1/\kappa} }+2\kappa^{-1}p(\mathcal H_n) (
				\frac{1} { x  }-1)\cr
				&=& A+2\kappa^{-1} p(\mathcal H_n) x^{-1}\leq  A+Ax^{-1}\leq 2Ax^{-1}.
			\end{eqnarray*}
			The last inequality holds for $x<1.$ Thus  with $n^{-10}\leq G$, and $k_n=O(b_n)$,
			\begin{eqnarray*}
				\int_0^1\sqrt{2\log k_n+g(x)}dx&\leq& \sqrt{ 2A } \int_0^1 x^{-1/2}dx \leq 4\sqrt{ 2\log k_n+2p(\mathcal H_n)[\log (
					C n )+\log G^{-1/\kappa} ]}\cr
				&\leq&4\sqrt{ 2\log k_n+2p(\mathcal H_n)[\log (
					C n )+\log n^{10/\kappa} ]} \leq C\sqrt{ p(\mathcal H_n)\log n   }. \end{eqnarray*}

			Therefore by the  Markov inequality,  for any $\varepsilon>0$,  with probability at least $1-\varepsilon/4,$
			$$
			\sup_{f\in\mathcal E} \left|\frac{1}{b_n}\sum_{l\leq b_n} a_nb_nn^{-1} U_{1,f}( \widetilde S^l)\right|
			\leq \frac{c_n}{\varepsilon},\quad c_n=  \delta_n^{\eta}  \sqrt{\frac{C p(\mathcal H_n)\log n}{b_n}}.
			$$

			\textbf{Step 5: completion.}  By (\ref{eqb.1ddaa}) and step 4,
			$$
			P\left( \sup_{f\in\mathcal E} \left|\frac{1}{b_n}\sum_{l\leq b_n} a_nb_nn^{-1} U_{1,f}(  S^l)\right|  > \frac{c_n}{\varepsilon}\right)
			\leq  P\left( \sup_{f\in\mathcal E} \left|\frac{1}{b_n}\sum_{l\leq b_n} a_nb_nn^{-1} U_{1,f}(  \widetilde S^l)\right|  > \frac{c_n}{\varepsilon}\right)+(b_n-1) \beta(a_n).
			$$
			We now take $a_n= M\log n/2 $ with  $M>0$ and $b_n=[n/(M\log n)]$. Then
			$(b_n-1)\beta(a_n)\to 0$ for sufficiently large $M.$ Also, the requirement in step 4 that $p(\mathcal H_n)=o(b_n)$ holds as long as $ p(\mathcal H_n)\log n=o(n)$.
			Hence with this choice of $b_n$,
			$$\sup_{f\in\mathcal E} \left|\frac{1}{b_n}\sum_{l\leq b_n} a_nb_nn^{-1} U_{1,f}(  S^l)\right|= O_P\left( \delta_n^\eta  \sqrt{\frac{ p(\mathcal{H}_n) \log^2 n}{n}}\right).$$
			The same rate applies when $U_{1,f}$ is replaced with $U_{2,f}$ following from the same proof of steps 2,3,4.

			In addition, $|\Upsilon|_0\leq 2a_n$. Hence
			\begin{eqnarray*}
				\mathbb E  \sup_{f\in\mathcal E}  \left| \frac{1}{n}\sum_{t\in \Upsilon} f(S_t)-\mathbb E f(S_t)\right|
				&\leq&\mathbb E  \frac{1}{n}\sum_{t\in \Upsilon}  \sup_{f\in\mathcal E}|f(S_t)|
				\leq\frac{Ca_n}{n} \mathbb E\max_{j\leq k_n}|\Psi_j(X_t) |  \sup_{\alpha\in\mathcal C_n }|\epsilon_t(\alpha)-\epsilon_t(\alpha_0)   |\cr
				&\leq &\frac{C\delta_n^{\eta}\log n}{n} .
			\end{eqnarray*}
			Together, by (\ref{eqb.2adfad})
			$
			\max_{j\leq k_n}\sup_{\alpha\in\mathcal C_n} |\frac{1}{n}\sum_t\Psi_j(X_t) (\epsilon_t( S_t, \alpha)-\epsilon(S_t,\alpha_0))|
			=O_P\left( \delta_n^\eta  d_n \right).
			$

		\end{proof}
	}

	\section{Proof of Theorem \protect\ref{th3.1}}
	
	\subsection{\protect\small Consistency}
	
	\begin{lem}[Consistency]
		{\small \label{lemconsist} Suppose $\frac{k_n}{n} +Q(\pi_n\alpha_0)=O(%
			\lambda)$. Also suppose $P_{en}(h)$ is lower semicompact on $(\mathcal{H}_n,
			\|.\|_{\infty,\omega})$ and $Q(\alpha)$ is lower semicontinuous. Then $\|\widehat\alpha-%
			\alpha_0\|_{\infty,\omega}=o_P(1)$. }
	\end{lem}
	
	{\small
		\begin{proof}  The proof of this lemma \textit{does not} depend on Assumption \ref{ass2.1}.
			First we show $P_{en}(\widehat h)=O_P(1)$.  Let $\rho_n(\alpha)$, $m_n(\alpha)$ be the $n\times 1$ vectors of $\rho(Y_{t+1},\alpha)$ and $m(X_t,\alpha)$. Let $\widehat\Sigma_n^{-1}$ be the diagonal matrix of $\widehat\Sigma(X_t)^{-1}$ for all $t$. By steps 1, 3  of the proof of Theorem \ref{th3.1} below,
			\begin{eqnarray*}
				\lambda P_{en}(\widehat h)& \leq& Q_n(\pi_n\alpha_0) + \lambda P_{en}(\pi_n h_0) +o_P(n^{-1})\cr
				&\leq &\frac{2}{n}\sum_t[\widetilde m (X_t,\pi_n\alpha_0) - \widehat  m (X_t,\pi_n\alpha_0)  ]^2 \widehat\Sigma(X_t)^{-1}
				+C\mathbb E\widetilde m(X_t,\pi_n\alpha_0)^2
				+ \lambda P_{en}(\pi_n h_0) +o_P(n^{-1})\cr
				&\leq& 2 [\rho_n(\pi_n\alpha_0) - m_n(\pi_n\alpha_0)]' P_n \widehat\Sigma_n^{-1}P_n[\rho_n(\pi_n\alpha_0) - m_n(\pi_n\alpha_0)]\cr
				&&+ C\mathbb E  m(X_t,\pi_n\alpha_0)^2  + \lambda P_{en}(\pi_n h_0) +o_P(n^{-1})\cr
				&\leq& O_P(\frac{k_n}{n} +Q(\pi_n\alpha_0) +\lambda )=O_P(\lambda)
			\end{eqnarray*}
			with the condition that $\frac{k_n}{n} +Q(\pi_n\alpha_0)=O(\lambda)$.
			So  let $M_0>0$ be a large constant so that $P_{en}(\widehat h)\leq M_0$ with probability arbitrarily close to one.
			
			Now take an arbitrary $\epsilon>0$, let
			$
			\mathcal B_\epsilon=\{\alpha=(\theta, h)\in \mathcal A_n: \| \alpha-\alpha_0\|_{\infty,\omega}\geq \epsilon, P_{en}(h)\leq M_0 \}.
			$
			Because $P_{en}(h)$ is lower semicompact on $(\mathcal H_0, \|.\|_{\infty,\omega})$ and $Q(\alpha)$ is lower semicontinuous, $\min_{\alpha\in \mathcal B_\epsilon }Q (\alpha)$ exists, that is, there is $\alpha^*\in \mathcal B_\epsilon$ so that $\inf_{\alpha\in \mathcal B_\epsilon }Q (\alpha)= Q (\alpha^*)>c_0.$
			If $\|\widehat\alpha-\alpha_0\|_{\infty,\omega}>\epsilon$,  then $Q(\widehat\alpha)\geq \inf_{\alpha\in \mathcal B_\epsilon }Q (\alpha) >c_0.$ Meanwhile,  by (\ref{eqb.1drad}) (to be proved below),
			$$
			c_0\leq Q(\widehat\alpha)  \leq  Q(\pi_n\alpha_0) + \lambda_n| P_{en}(\pi_n h_0)- P_{en}(\widehat h) |  +O_P(k_nd_n^2+\varphi_n^2).
			$$
			But the right hand side is $o_P(1)$. Hence we must have $\|\widehat\alpha-\alpha_0\|_{\infty,\omega}=o_P(1)$.
			
		\end{proof}
	}
	
	\subsection{{\protect\small Proof of Theorem \protect\ref{th3.1}}}
	
	{\small The proof depends on some important technical lemmas, one of which
		is the stochastic equicontinuity of $\epsilon(S_t,\alpha)=\rho(Y_{t+1},
		\alpha)-m(X_t, \alpha)$, given by Proposition \ref{lem5.2fd}. }
	
	{\small
		\begin{proof}

			We divide the proof in the following steps.  Let $\mathcal D_n$ be the sieve space used to estimate $m(X,\alpha)$, and
			$$
			\widetilde m(X,\alpha) = \argmin_{\widetilde m\in \mathcal D_n}\sum_{t=1}^n(m(X_t,\alpha)- \widetilde m(X_t))^2.
			$$
			We show the  following steps:
			
			\textbf{step 1. } Show that for $c, C>0$, uniformly in $\alpha\in\mathcal A_n$,
			$$
			c \mathbb E \widetilde m(X_t,\alpha)^2 \leq    \frac{1}{n}\sum_{t=1}^n\widetilde m(X_t,\alpha)^2\leq  C\mathbb E \widetilde m(X_t,\alpha)^2 .
			$$

			To prove it, we shall apply an \textit{empirical identifiability} result that first proved by \cite{huang1998projection} for the i.i.d. case and then extended by \cite{chen2015optimal} to more general setting with a much simpler proof.  We note that $\widetilde m(\cdot, \alpha)\in \mathcal D_n:=\{ g(x)= \sum_{j=1}^{k_n} \pi_j\Psi_j(x): \|g\|_{\infty,\omega}<\infty\}$. Let    $ \Psi_n  $ be the  $n\times k_n$ matrix of the   linear sieve bases, and let $A:=\frac{1}{n}\mathbb E \Psi_n' \Psi_n$.
			Suppose the   linear sieve   satisfies: $\lambda_{\min}(A)>c$ and  $\|  \frac{1}{n}\Psi_n' \Psi_n -A\|=o_P(1)$.
			Then  $\|A^{-1/2} \frac{1}{n}\Psi_n' \Psi_nA^{-1/2}-I\|=o_P(1) $, so the conditions of Lemma 4.1 of \cite{chen2015optimal}  are satisfied. We then apply this lemma to reach that
			$$
			\sup_{\alpha\in\mathcal A_n} \frac{|\frac{1}{n}\sum_t\widetilde m(X_t,\alpha)^2-\mathbb E\widetilde m(X_t,\alpha)^2|}{\mathbb E\widetilde m(X_t,\alpha)^2}
			\leq\sup_{g\in \mathcal D_n}\frac{|\frac{1}{n}\sum_tg(X_t)^2-\mathbb Eg(X_t)^2|}{\mathbb Eg(X_t)^2} =o_P(1).
			$$
			This then leads to the desired result.

			\textbf{step 2. } Show that
			$$
			\sup_{\alpha\in\mathcal A_n}  \frac{1}{n}\sum_{t=1}^n[\widetilde m(X_t,\alpha)-\widehat m(X_t,\alpha)]^2=O_P(k_nd_n^2),\quad d_n^2:=\frac{  p(\mathcal H_n)\log^2 n }{ n}.
			$$

			Let   $\epsilon(S_t, \alpha)=\rho(Y_{t+1}, \alpha)-m(X_t, \alpha).$  Also
			let $P_n=\Psi_n(\Psi_n'\Psi_n)^{-1}\Psi_n'$ and $\bar\epsilon_n(\alpha)$ be the $n\times 1$ vector of  $\epsilon(S_t, \alpha)$.  We then have
			\begin{eqnarray*}
				&&  \sup_{\alpha\in\mathcal A_n}  \frac{1}{n}\sum_{t=1}^n[\widetilde m(X_t,\alpha)-\widehat m(X_t,\alpha)]^2=
				\sup_{\alpha\in\mathcal A_n} \frac{1}{n} \bar\epsilon_n(\alpha)' P_n \bar\epsilon_n(\alpha)
				=O_P(1)\sup_{\alpha}\|\frac{1}{n}\Psi_n'\bar\epsilon_n(\alpha)\|^2\cr
				&\leq& O_P(k_n)\sup_{\alpha}\max_{j\leq k_n}|\frac{1}{n}\sum_{t=1}^n\Psi_j(X_t) \epsilon(S_t,\alpha)|^2
				=O_P(k_nd_n^2).
			\end{eqnarray*}
			The last bound is given by   Lemma \ref{sharp_iter}.

			\textbf{step 3. }   Show that  $  \sup_{\alpha\in\mathcal A_n}\mathbb E[\widetilde m(X_t,\alpha)-m(X_t,\alpha)]^2=O(\varphi_n^2).$
			
			Let $\widetilde m_n(\alpha)$ and $m_n(\alpha)$ respectively be the $n\times 1$ vectors of $\widetilde m(X_t,\alpha)$ and $m(X_t,\alpha)$. Also let $m_n(\alpha)=\Psi_nb_\alpha+r_\alpha$ where $r_\alpha$ is the sieve approximation error and $b_\alpha$ is the sieve coefficient to approximate  $m_n(X,\alpha)$. Then $\widetilde m_n(\alpha)= P_nm_n(\alpha)$ and
			\begin{eqnarray*}
				&&     \sup_{\alpha\in\mathcal A_n}\mathbb E[\widetilde m(X_t,\alpha)-m(X_t,\alpha)]^2=
				\frac{1}{n}       \sup_{\alpha\in\mathcal A_n}\mathbb E  m_n(\alpha)'(I-P_n)m_n(\alpha)
				= \frac{1}{n}       \sup_{\alpha\in\mathcal A_n}\mathbb E  r_\alpha'(I-P_n)r_\alpha\cr
				&\leq& \frac{1}{n}\sup_{\alpha}\mathbb E\|r_\alpha\|^2 =O_P(\varphi_n^2).
			\end{eqnarray*}
			
			After achieving the above three steps, then we have
			(since  $\widehat\Sigma(X_t)^{-1}$ and $\Sigma(X_t)^{-1}$ are  bounded away from zero)
			\begin{eqnarray*}
				Q_n(\widehat\alpha)&\geq&  \frac{c}{n}\sum_t\widehat m(X_t,\widehat\alpha)^2
				\geq  \frac{0.5c}{n}\sum_t\widetilde m(X_t,\widehat\alpha)^2
				-\frac{ c}{n}\sum_t[ \widehat m(X_t,\widehat\alpha)-\widetilde m(X_t,\widehat\alpha)]^2\cr
				&\geq^{(i)}&  c \mathbb E \widetilde m(X_t,\widehat\alpha)^2 -O_P(k_nd_n^2)
				\geq^{(ii)}  c \mathbb E   m(X_t,\widehat\alpha)^2 -O_P(k_nd_n^2+\varphi_n^2)
				\geq Q(\widehat\alpha)-O_P(k_nd_n^2) \cr
				Q_n(\pi_n\alpha_0)&\leq& \frac{C}{n}\sum_t\widehat m(X_t, \pi_n\alpha_0)^2
				\leq  \frac{2C}{n}\sum_t\widetilde m(X_t,\pi_n\alpha_0)^2
				+\frac{ 2C}{n}\sum_t[ \widehat m(X_t,\pi_n\alpha_0)-\widetilde m(X_t,\pi_n\alpha_0)]^2\cr
				&\leq^{(iii)}&  C \mathbb E \widetilde m(X_t, \pi_n\alpha_0)^2 +O_P( k_nd_n^2)
				\leq^{(iv)}  C \mathbb E   m(X_t, \pi_n\alpha_0)^2 +O_P(k_nd_n^2+\varphi_n^2)
				\cr
				&   \leq& Q( \pi_n\alpha_0)+O_P(k_nd_n^2+\varphi_n^2)
			\end{eqnarray*}
			where (i) (iii) follow from steps 1,2; (ii) (iv) follow  from step 3.
			
			Hence
			$
			Q_n(\widehat\alpha) + \lambda_nP_{en}(\widehat h) \leq  Q_n(\pi_n\alpha_0) + \lambda_nP_{en}(\pi_n h_0)+ o_P(n^{-1})
			$
			implies
			\begin{equation}\label{eqb.1drad}
			Q(\widehat\alpha)  \leq  Q(\pi_n\alpha_0) + \lambda_n| P_{en}(\pi_n h_0)- P_{en}(\widehat h) |  +O_P(k_nd_n^2+\varphi_n^2).
			\end{equation}
			Now by Assumption \ref{ass2.1},
			$$
			\|\widehat \alpha -\alpha_0\|^2\leq C\|\pi_n\alpha_0-\alpha_0\|^2+O_P(\lambda_n+k_nd_n^2+\varphi_n^2) .
			$$
			Hence $  \|\widehat \alpha -\pi_n\alpha_0\|\leq
			\|\widehat \alpha - \alpha_0\|+\|\pi_n\alpha_0-\alpha_0\|\leq
			C\|\pi_n\alpha_0-\alpha_0\|+O_P(\sqrt{\lambda_n}+\sqrt{k_n}d_n+\varphi_n) $. Thus
			\begin{eqnarray*}
				\|\widehat \alpha -\alpha_0\|_{\infty,\omega}&\leq&       \|\widehat \alpha -\pi_n\alpha_0\|_{\infty,\omega}
				+ \| \pi_n\alpha_0-\alpha_0\|_{\infty,\omega}
				\cr
				&       \leq & O_P(\| \pi_n\alpha_0-\alpha_0\|_{\infty,\omega}+\omega_n(\|\pi_n\alpha_0-\alpha_0\|+ \sqrt{\lambda_n}+\sqrt{k_n}d_n+\varphi_n)).
			\end{eqnarray*}
		\end{proof}
	}

	\begin{lem}
		{\small \label{sharp_iter} Suppose }
		
		{\small (a) $\mathbb{E}\max_{j\leq k_n}\Psi_j(X_t)^2 \sup_{\alpha\in\mathcal{%
					A}_n} \rho(Y_{t+1}, \alpha )^2\leq C^2$ }
		
		{\small (b) There are $\kappa>0$ and $C>0$ so that $\mathbb{E}\Psi_j(
			X_t)^2\sup_{ \|\alpha_1-\alpha_{2}\|_{\infty,\omega}< \delta} | \epsilon( S_t,
			\alpha_1)- \epsilon( S_t, \alpha_{2} )| ^2\leq C\delta^{2\kappa}$ holds for
			any $\delta>0.$ }
		
		{\small (c) $p(\mathcal{H}_n)\to\infty$ and $p(\mathcal{H}_n)\log n=o(n)$. }
		
		{\small Then $\sup_{\alpha}\max_{j\leq k_n}|\frac{1}{n}\sum_{t=1}^n%
			\Psi_j(X_t) \epsilon(S_t,\alpha)|=O_P(\sqrt{\frac{ p(\mathcal{H}_n)\log^2 n
				}{ n}}). $ }
	\end{lem}
	
	{\small
		\begin{proof}


			Let $\mathcal E:=\{  \epsilon( \cdot, \alpha)  \Psi_j : \alpha\in \mathcal A_n, j\leq k_n\}$. We divide the proof into several steps.
			
			\textbf{Step 1: construct blocks.}  This step is the same as that of the proof of  Proposition \ref{lem5.2fd}.
			
			\textbf{Step 2: the envelop function for $U_{1,f}$.}
			Note that $\mathbb Ef=0$ for $f\in\mathcal E$  and that $\widetilde S_t$ and $S_t$ are identically distributed within each  block $H_{1,l}$. By Cauchy-Schwarz,
			\begin{eqnarray*}
				\mathbb E\sup_{f\in\mathcal E} U_{1,f}(\widetilde S^l)^2&\leq&  \mathbb E\sup_{f\in\mathcal E} \left(\frac{1}{a_n}\sum_{t\in H_{1,l}} f(\widetilde S_t) \right)^2\leq  \frac{1}{a_n} \sum_{t\in H_{1,l}}\mathbb E \sup_{f\in\mathcal E}  f( S_t)   ^2\cr
				&\leq&
				2\mathbb E\max_{j\leq k_n}\Psi_j(X_t)^2 \sup_{\alpha\in\mathcal A_n}  \rho(Y_{t+1}, \alpha   )^2\leq C^2 .
			\end{eqnarray*}
			Let $\mathcal F=\{U_{1,f}: f\in\mathcal E\}$ and
			let  $ F:=\max\{ n^{-10}, \sup_{f\in\mathcal E}|U_{1,f}| \}$.  Then both $\sup_{f\in\mathcal E}|U_{1, f}| $ and $F$ are envelope functions of $\mathcal F$, and
			$$
			n^{-10}\leq G:=\|F\|_{L_2(S_t)}\leq C .
			$$
			
			\textbf{Step 3: the bracketing number.}
			We aim to apply Theorem 2.14.2 of \cite{VW} to bound $ \frac{1}{b_n}\sum_{l\leq b_n} a_nb_nn^{-1} U_{1,f}( \widetilde S^l) $, which requires bounding the bracketing number of $\mathcal F.$
			To do so,
			suppose $h_1,..., h_N$ is a $\delta$-cover of $\mathcal H_n$ under the norm $\|h\|_{\infty,\omega} $ and $N:= \mathcal N(\delta, \mathcal H_n, \|.\|_{\infty,\omega})$; $\theta_1,...,\theta_R$ is a $\delta$-cover of $\Theta$ and $R:=\mathcal N(\delta, \Theta, \|.\|)$ (the Euclidean norm in $\Theta$). Here $\mathcal N(\delta, \mathcal A,.)$ denotes the covering number for space $\mathcal A.$

			Then    for any $f=\epsilon(, \alpha ) \Psi_j \in\mathcal E$,  there are $\Psi_j$  and $\alpha_{ik}=(\theta_k, h_i)$ so that $\|\alpha-\alpha_{ik}\|_{\infty,\omega}\leq \|h-h_i\|_{\infty,\omega}+\|\theta-\theta_k\|\leq2\delta$.    Let $f_{ijk}= \epsilon (\cdot,  \alpha_{ik} ) \Psi_j$.
			We have
			\begin{eqnarray*}
				&&  \sup_{f=\epsilon(\cdot, \alpha)\Psi_j : \|\alpha-\alpha_{ik}\|_{\infty,\omega}<2\delta }  | U_{1,f}(\widetilde S^l)
				-  U_{1,f_{ijk}}(\widetilde S^l)  |
				\leq  \sup_{f=\epsilon(\cdot, \alpha)\Psi_j : \|\alpha-\alpha_{ik}\|_{\infty,\omega}<2\delta }  | \frac{1}{a_n}  \sum_{t\in H_{1,l}} f(\widetilde S_t)-     f_{ijk }(\widetilde S_t) |
				\cr
				&\leq&  \frac{1}{a_n}  \sum_{t\in H_{1,l}} |\Psi_j(\widetilde X_t)|\sup_{ \|\alpha-\alpha_{ik}\|_{\infty,\omega}<2\delta}  |   \epsilon_t(\widetilde S_t, \alpha )-\epsilon_t(\widetilde S_t, \alpha_{ik})|  :=b_{ijk}(\widetilde S^l,\delta).
			\end{eqnarray*}
			Then $U_{1,f}\in[l_{ijk}, u_{ijk}]$, where $l_{ijk}:= U_{1, f_{ijk}}- b_{ijk}(,\delta)$ and $u_{ijk}=U_{1, f_{ijk}}+b_{ijk}(,\delta)$.    In addition,
			\begin{eqnarray*}
				\mathbb E[u_{ijk}-l_{ijk}]^2&\leq&4\mathbb Eb_{ijk}(\widetilde S^l,\delta)^2\cr
				&\leq&  C \mathbb E\left(  \frac{1}{a_n}  \sum_{t\in H_{1,l}} |\Psi_j(\widetilde X_t)| \sup_{ \|\alpha-\alpha_{ik}\|_{\infty,\omega}<2\delta}  |  \epsilon_t(\widetilde S_t, \alpha)- \epsilon_t(\widetilde S_t, \alpha_{ik})| \right)^2\cr
				&\leq& C \mathbb E\Psi_j(\widetilde X_t)^2\sup_{ \|\alpha-\alpha_{ik}\|_{\infty,\omega}<2\delta} | \epsilon( \widetilde   S_t, \alpha)- \epsilon(  \widetilde  S_t, \alpha_{ik} )| ^2\leq C\delta^{2\kappa}.
			\end{eqnarray*}
			Hence $\{[l_{ijk}, u_{ijk}]:i\leq N, j\leq k_n, k\leq R \}$ is a $C \delta^{\kappa}$ bracket of $\mathcal F$, whose bracketing number satisfies
			$$\mathcal N_{[]}( C\delta^{\kappa},\mathcal F, \|.\|_{L^2(\widetilde S_t)})\leq \underbrace{\mathcal N(\delta, \mathcal H_n, \|.\|_{\infty,\omega})}_{N}  \underbrace{(C/\delta)^d}_R  k_n, $$
			where we used $R\leq (C/\delta)^d$ for $d=\dim(\theta_0)$ since $\theta_0\in\Theta$ is compact.
			Then for a generic constant $C>0$,
			$$\mathcal N_{[]}( Gx,\mathcal F, \|.\|_{L^2(\widetilde S_t)})\leq C\mathcal N(x^{1/\kappa}(G/C)^{1/\kappa}, \mathcal H_n, \|.\|_{\infty,\omega})G^{-d/\kappa} x^{-d/\kappa}k_n,\quad \forall x>0.$$

			\textbf{Step 4: bound independent blocks.} Note that $U_{1,f}(\widetilde S^l)$ are independent across $l$ and is mean-zero. For the envelop $G$ defined in step 2 and some constant $\bar M>0$,
			\begin{eqnarray*}
				&& \mathbb E\sup_{f\in\mathcal E}\left|  \frac{1}{b_n}\sum_{l\leq b_n} a_nb_nn^{-1} U_{1,f}(  \widetilde S^l)\right| \leq  \frac{1}{2} \mathbb E\sup_{g\in\mathcal F}\left|  \frac{1}{b_n}\sum_{l\leq b_n} g(  \widetilde S^l)\right| \cr
				&\leq^{(i)}& b_n^{-1/2}G
				\int_0^1\sqrt{1+\log\mathcal N_{[]}(Gx, \mathcal F, \|.\|_{L_2(\widetilde S_t)})} dx    \cr
				&\leq &
				\frac{C }{\sqrt{b_n}}     \int_0^1\sqrt{1+\log\mathcal N(x^{1/\kappa} (G/C)^{1/\kappa} , \mathcal  H_n, \|.\|_{\infty})+ \log \frac{C}{ G^{d/\kappa} x^{d/\kappa}}+\log k_n} dx    \cr
				&\leq^{(ii)}&\frac{C    }{\sqrt{b_n}}   \int_0^1\sqrt{1+p(\mathcal H_n)\log
					\frac{C    n}{x^{1/\kappa} G^{1/\kappa} }+ d\log \frac{C}{G^{1/\kappa} x^{1/\kappa}}+\log k_n}dx     \cr
				&\leq^{(iii)}&\frac{C   }{\sqrt{b_n}}   \int_0^1\sqrt{2\log k_n+2p(\mathcal H_n)\log
					\frac{C    n}{x^{1/\kappa} G^{1/\kappa} } }dx   \leq^{(iv)}     \sqrt{\frac{Cp(\mathcal H_n)\log n}{b_n}},
			\end{eqnarray*}
			where (i)   follows   from Theorem  2.14.2 of  \cite{VW};  (ii) follows  from Assumption     \ref{assumcomplexity}; (iii) is due to $p(\mathcal H_n)\to\infty $ . (iv) follows from the same proof as that of Proposition   \ref{lem5.2fd}.

			\textbf{Step 5: completion.}  By an inequality similar to (\ref{eqb.1ddaa}) and step 4,
			$$
			P\left( \sup_{f\in\mathcal E} \left|\frac{1}{b_n}\sum_{l\leq b_n} a_nb_nn^{-1} U_{1,f}(  S^l)\right|  > \frac{c_n}{\varepsilon}\right)
			\leq  P\left( \sup_{f\in\mathcal E} \left|\frac{1}{b_n}\sum_{l\leq b_n} a_nb_nn^{-1} U_{1,f}(  \widetilde S^l)\right|  > \frac{c_n}{\varepsilon}\right)+(b_n-1) \beta(a_n).
			$$
			We now take $a_n= M\log n/2 $ with  $M>0$ and $b_n=[n/(M\log n)]$. Then
			$(b_n-1)\beta(a_n)\to 0$ for sufficiently large $M.$ Also, the requirement in step 4 that $p(\mathcal H_n)=o(b_n)$ holds as long as $ p(\mathcal H_n)\log n=o(n)$.
			Hence with this choice of $b_n$,
			$$\sup_{f\in\mathcal E} \left|\frac{1}{b_n}\sum_{l\leq b_n} a_nb_nn^{-1} U_{1,f}(  S^l)\right|= O_P\left(  \sqrt{\frac{  p(\mathcal H_n)\log^2 n }{ n}}\right).$$
			The same rate applies when $U_{1,f}$ is replaced with $U_{2,f}$ following from the same proof of steps 2,3,4.

			In addition, $|\Upsilon|_0\leq 2a_n$. Hence
			\begin{eqnarray*}
				\mathbb E  \sup_{f\in\mathcal E}  \left| \frac{1}{n}\sum_{t\in \Upsilon} f(S_t)-\mathbb E f(S_t)\right|
				&\leq&2\mathbb E  \frac{1}{n}\sum_{t\in \Upsilon}  \sup_{f\in\mathcal E}|f(S_t)|
				\leq\frac{Ca_n}{n} \mathbb E\max_{j\leq k_n}|\Psi_j(X_t) |  \sup_{\alpha }|\epsilon (S_t, \alpha)   |\leq \frac{C \log n}{n} .
			\end{eqnarray*}
			Together,  
			$
			\max_{j\leq k_n}\sup_{\alpha\in\mathcal A_n} |\frac{1}{n}\sum_t\Psi_j(X_t)  \epsilon (S_t, \alpha) |
			=O_P\left( \sqrt{\frac{  p(\mathcal H_n)\log^2 n }{ n}}\right).
			$

		\end{proof}
	}
	
	\section{Proofs for Section \ref{sec:normality}}
	
	\subsection{\protect\small Local quadratic approximation}
	
	\begin{prop}[LQA]
		{\small \label{assLQA} Let $\mathcal{C}_n=\{\alpha+xu_n: |x|<Cn^{-1/2}, \alpha\in\mathcal{A}_n, 
			\|\alpha-\alpha_0\|_{\infty,\omega}\leq C\delta_n, Q(\alpha)\leq
			C\bar\delta_n^2 \}$. Suppose for $u_n=v_n^*/\|v_n^*\| $, there are $C>0$, so
			that }
		
		{\small (a) $\sqrt{n}\bar \delta_n\|\widehat\Sigma_n-\Sigma_n\| =o(1)$, $%
			\varphi_n^2\bar\delta_n^2 + k_nd_n^2\delta_n^{2\eta} + \sqrt{k_n}
			d_n\delta_n^{\eta}\bar\delta_n =o(n^{-1})$. }
		
		{\small (b) $\frac{1}{\sqrt{n}}\| (I-P_n) \Sigma_n^{-1} \frac{dm_n(\alpha)}{%
				d\alpha}[u_n] \| +\frac{1}{\sqrt{n}}\| (I-P_n) \frac{dm_n(\alpha)}{d\alpha}%
			[u_n] \|= O_P(\varphi_n) .$ }
		
		{\small (c) $k_n\sup_{\alpha\in\mathcal{C}_n}\frac{1}{n}\sum_t[ \frac{%
				dm(X_t,\alpha )}{d\alpha}[u_n]- \frac{dm(X_t,\alpha_0)}{d\alpha}%
			[u_n]]^2=o_P( 1)$. }
		
		{\small (d) conditions of Proposition \ref{lem5.2fd} hold. }
		
		
		{\small (e) $\sup_{\tau \in (0,1)}\sup_{\alpha\in\mathcal{C}_n} \mathbb{E }\left[ \frac{d ^2
				m(X_t,\alpha_0+\tau(\alpha-\alpha_0))}{d\tau^2}|\right]^2 =o(n^{-1})$ and }
		
		{\small (f) $\mathbb{E }\sup_{\alpha\in\mathcal{C}_n}\sup_{|\tau|\leq
				Cn^{-1/2}}\frac{1}{n}\sum_t\left[\frac{d^2}{d\tau^2} m(X_t, \alpha+ \tau
			u_n)| \right]^2 =O(1)$. }
		
		{\small
		}
		
		{\small
		}
		
		{\small Then
			\begin{equation*}
			\sup_{ \alpha\in\mathcal{A}_{osn}} \sup_{|x|\leq
				Cn^{-1/2}}|Q_n(\alpha+xu_n)-Q_n(\alpha)- A_n(\alpha(x))|=o_P(n^{-1}),
			\end{equation*}
			where }
		
		{\small (a1) $A_n(\alpha(x)):= 2 x[n^{-1/2} Z_n+\langle
			u_n,\alpha-\alpha_0\rangle]+ B_nx^2$ }
		
		{\small (a2) $B_n=\frac{1}{n} \frac{d m_n( \alpha_0 )}{d\alpha}[u_n]^{\prime
			}\Sigma_n^{-1} \frac{dm_n( \alpha_0 )}{d\alpha}[u_n] \to^P1$, and }
		
		{\small (a3) $Z_n\to^d\mathcal{N}(0,1)$. }
	\end{prop}
	
	{\small
		\begin{proof}  Let $\widetilde Q_n(\alpha)= \frac{1}{n}\sum_t\ell(X_t, \alpha)^2\widehat\Sigma(X_t)^{-1}$,  and
			$$
			\ell(x, \alpha):=\widetilde m(x,\alpha) +\widehat m(x,\alpha_0),\quad
			\widetilde m(x,\alpha):=\Psi(x)'(\Psi_n'\Psi_n)^{-1}\Psi_n'm_n(\alpha).
			$$

			\textbf{Step 1: expansions.}    By assumption, $\widetilde Q_n(\alpha)$ is differentiable. So we shall prove the LQA for $\widetilde Q_n(\alpha)$ via the mean value theorem, and show that $\widetilde Q_n(\alpha)-  Q_n(\alpha)$ is ``small" locally. Indeed, Lemma \ref{lc.1} shows that $\sup_{ \alpha\in\mathcal C_n}      |Q_n(\alpha)-  \widetilde Q_n(\alpha)   |=o_P(n^{-1}).$
			
			We write $f(s):=\widetilde Q_n(\alpha+sxu_n) $
			and by the second order mean value theorem, for some $s\in(0,1)$,
			\begin{eqnarray*}
				&&\widetilde Q_n(\alpha+xu_n)-\widetilde  Q_n(\alpha) =f'(0)+\frac{1}{2}f''(s)
				=2x G(\alpha) +x^2 B_x
				+x^2D_x,\cr
				G(\alpha)&:=&\frac{1}{n}\sum_t\ell(X_t,\alpha)\widehat\Sigma(X_t)^{-1}\frac{d  \widetilde  m(X_t,\alpha)}{d\alpha}[u_n]\cr
				B_x&:=&\frac{1}{n}\sum_t\left(\frac{d   \widetilde m(\alpha+sxu_n)}{d\alpha}[u_n]\right)^2\widehat\Sigma(X_t)^{-1}\cr
				D_x&:=&\frac{1}{n}\sum_t\ell(\alpha+sxu_n)\widehat\Sigma(X_t)^{-1}\frac{d^2}{d\tau^2}  \widetilde m(\alpha+\tau xu_n)|_{\tau=s} .
			\end{eqnarray*}
			Lemma \ref{lc.2} shows that uniformly $D_x=o_P(1) $. Hence $\sup_{|x|\leq Cn^{-1/2}}x^2 |D_x|=o_P(n^{-1})$.

			\textbf{Step 2: convergence of $B_x$.}    Let $\frac{d m_n( \alpha )}{d\alpha}[u_n]$ and $\rho_n$ be the $n\times 1$ vectors  of   $\frac{d m(X_t, \alpha )}{d\alpha}[u_n]$ and $\rho(Y_{t+1},\alpha_0)$. Also let $\|v\|^2_{\Sigma}:=v'\Sigma_n^{-1}v$. Write $ B_n:=\frac{1}{n}\| \frac{d  m_n( \alpha_0  )}{d\alpha}[u_n]\|_{\Sigma}^2=O_P(1)$. Uniformly for $\alpha(x),s$,
			\begin{eqnarray*}
				B_x-B_n&\leq&
				\frac{1}{n}\| \frac{d \widetilde m_n(\alpha+sxu_n ) }{d\alpha}[u_n]\|^2_{\widehat\Sigma_n}
				-  \frac{1}{n}\| \frac{d \widetilde m_n(\alpha_0 ) }{d\alpha}[u_n]\|^2_{\widehat\Sigma_n}
				\cr
				&&+
				\frac{1}{n}\| \frac{d \widetilde m_n(\alpha_0 ) }{d\alpha}[u_n]\|^2_{\widehat\Sigma_n}
				-\frac{1}{n}\| \frac{d   m_n(\alpha_0 ) }{d\alpha}[u_n]\|^2_{\widehat\Sigma_n}
				\cr
				&&+ \frac{1}{n}\| \frac{d  m_n(\alpha_0 ) }{d\alpha}[u_n]\|^2_{\widehat\Sigma_n}-\frac{1}{n}\| \frac{d  m_n(\alpha_0 ) }{d\alpha}[u_n]\|^2_{ \Sigma_n} =o_P(1).
			\end{eqnarray*}
			Hence $\sup_{|x|\leq Cn^{-1/2} }| B_x-B_n|x^2=o_P(n^{-1})$. To show that $B_n\to^P1$, we have
			$$
			B_n= \langle u_n, u_n\rangle+\left[\frac{1}{n} \frac{d  m_n( \alpha_0  )}{d\alpha}[u_n]'   \Sigma_n^{-1}   \frac{dm_n( \alpha_0  )}{d\alpha}[u_n]-\langle u_n, u_n\rangle\right]=  \langle u_n, u_n\rangle+o_P(1).
			$$
			Let $Z_t=\rho(Y_{t+1}, \alpha_0)\Sigma(X_t)^{-1}\frac{d  m(X_t,\alpha_0)}{d\alpha}[v] $.  Then for each $v$,
			\begin{eqnarray*}
				\|v\| ^2&= &\var(\frac{1}{\sqrt{n}}\sum_tZ_t) =\var(Z_t) +\frac{2}{n}\sum_{s>t} \mathbb  E Z_t\mathbb E (Z_s|\sigma_s(\mathcal X))=\var(Z_t)=\langle v, v\rangle.
			\end{eqnarray*}
			Hence   $\langle u_n, u_n\rangle=\langle v^*_n, v^*_n\rangle \|v_n^*\| ^{-2}=1.$ Hence $B_n=1+o_P(1).$

			\textbf{Step 3: expansion of $G(\alpha)$.}
			We have   $\sup_{\alpha\in\mathcal C_n}\frac{1}{\sqrt{n}}\|m_n(\alpha)'P_n\|+\sup_{\alpha\in\mathcal C_n}\frac{1}{\sqrt{n}}\|m_n(\alpha) \|  =O_P(\bar \delta_n)$ and $ (\sqrt{n}\bar \delta_n+\sqrt{k_n})\|\widehat\Sigma_n-\Sigma_n\| =o(1)$.
			\begin{eqnarray*}
				G(\alpha)&=& \frac{1}{n}    m_n(\alpha) 'P_n\widehat\Sigma_n^{-1} \frac{d\widetilde  m_n(\alpha)}{d\alpha}[u_n] + \frac{1}{n}   \rho_n'P_n \widehat\Sigma_n^{-1} \frac{d  \widetilde  m_n(\alpha)}{d\alpha}[u_n]\cr
				&=& \frac{1}{n}    m_n(\alpha) 'P_n \Sigma_n^{-1} \frac{d\widetilde m_n(\alpha)}{d\alpha}[u_n] + \frac{1}{n}   \rho_n'P_n  \Sigma_n^{-1} \frac{d\widetilde m_n(\alpha)}{d\alpha}[u_n]+o_P(n^{-1/2})\cr
				&=&  \frac{1}{n}    m_n(\alpha) 'P_n \Sigma_n^{-1} \frac{d m_n(\alpha)}{d\alpha}[u_n] + \frac{1}{n}   \rho_n'P_n  \Sigma_n^{-1} \frac{dm_n(\alpha)}{d\alpha}[u_n]\cr
				&&+ \frac{1}{n}    m_n(\alpha) 'P_n \Sigma_n^{-1} (P_n-I)\frac{d m_n(\alpha)}{d\alpha}[u_n] + \frac{1}{n}   \rho_n'P_n  \Sigma_n^{-1} (P_n-I)\frac{d m_n(\alpha)}{d\alpha}[u_n]+o_P(n^{-1/2})\cr
				&=&  \frac{1}{n}    m_n(\alpha) 'P_n \Sigma_n^{-1} \frac{d m_n(\alpha)}{d\alpha}[u_n] + \frac{1}{n}   \rho_n'P_n  \Sigma_n^{-1} \frac{dm_n(\alpha)}{d\alpha}[u_n]
				+ O_P(\varphi_n\bar\delta_n) +o_P(n^{-1/2}) \cr
				&=&  \frac{1}{n}    m_n(\alpha) ' \Sigma_n^{-1} \frac{d m_n(\alpha)}{d\alpha}[u_n] + \frac{1}{n}   \rho_n'P_n  \Sigma_n^{-1} \frac{dm_n(\alpha)}{d\alpha}[u_n]
				+o_P(n^{-1/2})
				\cr
				&=^{(a)}& \langle u_n,\alpha-\alpha_0\rangle + \frac{1}{n}     \rho_n'P_n  \Sigma_n^{-1} \frac{dm_n(\alpha)}{d\alpha}[u_n] +o_P(n^{-1/2})\cr
				&=^{(b)}&  \langle u_n,\alpha-\alpha_0\rangle + \underbrace{  \frac{1}{n}   \rho_n(\alpha_0)'  \Sigma_n^{-1} \frac{dm_n(\alpha_0)}{d\alpha}[u_n] }_{\frac{1}{\sqrt{n}}Z_n} +o_P(n^{-1/2}),
			\end{eqnarray*}
			where (a) follows from Lemma \ref{lc.2}; (b) is  due to
			\begin{eqnarray*}
				&&\sqrt{\mathbb E\rho_n(\alpha_0)'P_n\rho_n(\alpha_0)}\sqrt{\sup_{\alpha\in\mathcal C_n} \frac{1}{n}\sum_t[ \frac{dm(X_t,\alpha )}{d\alpha}[u_n]- \frac{dm(X_t,\alpha_0)}{d\alpha}[u_n]]^2  }\cr
				&\leq& \sqrt{\mathbb E \text{tr} P_n\Sigma(X_t)^{-1}}\sqrt{\sup_{\alpha\in\mathcal C_n} \frac{1}{n}\sum_t[ \frac{dm(X_t,\alpha )}{d\alpha}[u_n]- \frac{dm(X_t,\alpha_0)}{d\alpha}[u_n]]^2  } =o_P(1).
			\end{eqnarray*}
			
			\textbf{Step 4: weak convergence of $Z_n$.}
			It then remains to show  $Z_n\to^d\mathcal N(0,1)$.
			Note that
			$$
			Z_n      =  \frac{1}{\sqrt{n}} \sum_t  \mathcal Z_t\|v_n^*\| ^{-1},\quad \mathcal Z_t=\rho(Y_{t+1}, \alpha_0)\Sigma(X_t)^{-1}\frac{d  m(X_t,\alpha_0)}{d\alpha}[v^*_n]  ,
			$$
			where $u_n=v_n^*/\|v_n^*\| $.     When $s>t$,  we have $\mathcal Z_t\in\sigma_s(\mathcal X)$. Hence
			$
			\mathbb E(\mathcal Z_t\mathcal Z_s|\sigma_s( X))
			=\mathcal Z_t  \mathbb E(\mathcal Z_s|\sigma_s( X))=0.
			$
			Thus
			\begin{eqnarray*}
				\var(\frac{1}{\sqrt{n}}\sum_t\mathcal Z_t)
				&=& \var( \mathcal Z_t)+2\frac{1}{n}\sum_{s>t} \mathbb E \mathbb E(\mathcal Z_t\mathcal Z_s|\sigma_s( X))  = \var( \mathcal Z_t)\cr
				&=& \mathbb E\var\left[  \rho(Y_{t+1}, \alpha_0)\Sigma(X_t)^{-1}\frac{d  m(X_t,\alpha_0)}{d\alpha}[v^*_n]\bigg{ |} \sigma_t(\mathcal X)\right]\cr
				&=&\mathbb E\Sigma(X_t)^{-2}\left(\frac{d  m(X_t,\alpha_0)}{d\alpha}[v^*_n]\right)^2\var( \rho(Y_{t+1}, \alpha_0)|\sigma_s(\mathcal X))\cr
				&=&\langle v_n^*,v_n^*\rangle =\|v_n^*\| ^2
			\end{eqnarray*}
			where we used $\var( \rho(Y_{t+1}, \alpha_0)|\sigma_s(\mathcal X))=\Sigma(X_t)$.
			
			Next, it is assumed that there is some $\zeta>0$,
			$$
			\mathbb E |\mathcal Z_t\|v_n^*\| ^{-1}|^{2+\zeta} \leq C\mathbb E | \rho(Y_{t+1}, \alpha_0)|^{2+\zeta}\left| \frac{d  m(X_t,\alpha_0)}{d\alpha}[u_n]   \right |^{2+\zeta}<\infty.
			$$
			In addition,  $\mathcal Z_t$ is strictly stationary, satisfying the $\beta$-mixing condition (Assumption \ref{ass3.10}). Let $\alpha(n)$ denote the $\alpha$-mixing coefficient (the strong mixing coefficient). We have that, by Assumption \ref{ass3.10},
			$
			\alpha(n)\leq \frac{1}{2}\beta(n)\leq C\exp(-cn)
			$
			for some $c, C>0$.
			Hence
			$$\sum_{n=1}^{\infty}\alpha(n)^{\zeta/(2+\zeta)}\leq C \sum_{n=1}^{\infty}\exp(-c\zeta n/(2+\zeta))<\infty. $$
			Then by Theorem 1.7 of \cite{ibragimov1962some},  $ Z_n\to^d\mathcal N(0,1)$.
		\end{proof}
	}
	
	\begin{lem}
		{\small \label{lc.1} Let $\widetilde Q_n(\alpha)= \frac{1}{n}\sum_t\ell(X_t,
			\alpha)^2\widehat\Sigma(X_t)^{-1}$, and
			\begin{equation*}
			\ell(x, \alpha):=\widetilde m(x,\alpha) +\widehat m(x,\alpha_0),\quad
			\widetilde m(x,\alpha):=\Psi(X_t)^{\prime }(\Psi_n^{\prime
			}\Psi_n)^{-1}\Psi_n^{\prime }m_n(\alpha).
			\end{equation*}
			Suppose   $k_nd_n^2\delta_n^{2\eta} + \sqrt{k_n} d_n\delta_n^{\eta}\bar%
			\delta_n =o(n^{-1})$ and $\frac{1}{\sqrt{n}}\| m_n(\pi_n \alpha) -
			m_n(\alpha)\| \bar\delta_n \leq o(n^{-1}). $
		}
		
		{\small Then for $\mathcal{C}_n=\{\alpha+ xu_n: \alpha\in\mathcal{A}_n:
			\|\alpha-\alpha_0\|_{\infty,\omega}\leq C\delta_n,  Q(\alpha)\leq
			C\bar\delta_n^2, |x|\leq Cn^{-1/2} \}$, }
		
		{\small (i)
			\begin{equation*}
			\sup_{ \alpha\in\mathcal{C}_n} |Q_n(\alpha)- \widetilde Q_n(\alpha)
			|=o_P(n^{-1}).
			\end{equation*}
		}
		
		{\small (ii)
			\begin{equation*}
			\sup_{ \alpha\in\mathcal{C}_n}| Q_n(\alpha)-Q_n(\pi_n\alpha)| =o_P(n^{-1}).
			\end{equation*}
		}
	\end{lem}
	
	{\small
		\begin{proof}  (i)  Recall that   $\epsilon_t(\alpha)=\rho(Y_{t+1}, \alpha)-m(X_t, \alpha)$  and $m_n(\alpha)$,   $\bar\epsilon_n(\alpha) $  and $\rho_n(\alpha)$ are $n\times 1$ vectors of   $m(X_t, \alpha)$, $\epsilon_t(\alpha)$ and $\rho(Y_{t+1},\alpha)$.  Also write $\alpha(x):=\alpha+xu_n$.
			\begin{eqnarray*}
				&&   Q_n(\alpha+xu_n)-  \widetilde Q_n(\alpha+xu_n)=\frac{1}{n}\sum_t[\widehat m(X_t,\alpha(x))^2 -\ell(X_t, \alpha(x))^2]\widehat\Sigma(X_t)^{-1}\cr
				&=&\frac{1}{n}[\bar\epsilon_n(\alpha+xu_n) -\bar\epsilon_n(\alpha_0)]'P_n\widehat\Sigma_n^{-1} P_n    [\bar\epsilon_n(\alpha+xu_n) -\bar\epsilon_n(\alpha_0)  +2 m_n(\alpha+xu_n) +2\rho_n(\alpha_0)]\cr
				&\leq& O_P(1)  \frac{1}{n}\|P_n    [\bar\epsilon_n(\alpha+xu_n) -\bar\epsilon_n(\alpha_0)\|^2
				+O_P(1)\frac{1}{n}\|P_n    [\bar\epsilon_n(\alpha+xu_n) -\bar\epsilon_n(\alpha_0)\| \|P_nm_n(\alpha+xu_n) \|\cr
				&&+O_P(1)\frac{1}{n}\|P_n    [\bar\epsilon_n(\alpha+xu_n) -\bar\epsilon_n(\alpha_0)\| \| P_n\rho_n(\alpha_0)\|\cr
				&\leq& O_P(d_1^2+ d_1\times d_2+ d_1\times d_3)\cr
				d_1&:=& \frac{1}{\sqrt{n}}\|P_n    [\bar\epsilon_n(\alpha+xu_n) -\bar\epsilon_n(\alpha_0)\|\cr
				d_2&:=& \frac{1}{\sqrt{n}}\|P_n    m_n(\alpha+xu_n) \|,\quad d_3:= \frac{1}{\sqrt{n}}\|P_n    \rho_n(\alpha_0) \|.
			\end{eqnarray*}
			We shall respectively calculate $d_1\sim d_3.$
			By Proposition \ref{lem5.2fd}, $d_1= O_P(\sqrt{k_n} d_n\delta_n^{\eta})$ uniformly in $\alpha(x)$.
			As for $d_2$, by steps 1 and 3 in the proof of Theorem  \ref{th3.1}, uniformly in $\alpha(x)$,
			$$
			d_2^2\leq\frac{1}{n}\sum_t\widetilde m(X_t,\alpha(x))^2\leq C\mathbb E\widetilde m(X_t,\alpha(x))^2
			\leq C(\varphi_n^2+ \mathbb E m(X_t,\alpha(x))^2)
			\leq  C\bar\delta_n^2.
			$$
			Finally, $d_3^2=O_P(\frac{k_n}{n})$. Together,
			$
			Q_n(\alpha+xu_n)-  \widetilde Q_n(\alpha+xu_n)
			\leq O_P(k_nd_n^2\delta_n^{2\eta} + \sqrt{k_n} d_n\delta_n^{\eta}\bar\delta_n )=o_P(n^{-1}).
			$

			(ii) Let  $m_n(\alpha)$ and $\widetilde m_n(\alpha)$ respectively be the $n\times 1$ vectors of $m(X_t,\alpha)$ and $\widetilde m(X_t, \alpha)$. First, $
			\frac{1}{\sqrt{n}}\| \widetilde   m_n(\pi_n \alpha) -  \widetilde  m_n(\alpha)\|\leq
			\frac{1}{\sqrt{n}}\|     m_n(\pi_n \alpha) -    m_n(\alpha)\|\leq   O_P(\mu_n).
			$
			Second, $ \frac{1}{n}\|  \widetilde m_n(\alpha)\| ^2\leq   O_P( \bar \delta_n^2 )$. Third,
			$\frac{1}{\sqrt{n}}\| \widehat m_n(  \alpha_0) \|=O_P(1)\sqrt{\frac{1}{n}\rho_n(\alpha_0)'P_n\rho_n(\alpha_0)}=O_P(\sqrt{\frac{k_n}{n}}).$
			
			Hence
			for  $\widetilde Q_n(\alpha)= \frac{1}{n} [\widetilde m_n(\alpha)+\widehat m_n(\alpha_0)]'\widehat \Sigma_n^{-1}[\widetilde m_n(\alpha)+\widehat m_n(\alpha_0)]$, we have
			\begin{eqnarray*}
				&& \widetilde Q_n(\alpha)- \widetilde Q_n(\pi_n\alpha)\cr
				&\leq& O_P(1)\frac{1}{\sqrt{n}}\|  \widetilde m_n(\pi_n \alpha) - \widetilde  m_n(\alpha)\|
				\left[\frac{1}{\sqrt{n}}\|  \widetilde   m_n(\pi_n \alpha) - \widetilde   m_n(\alpha)\| +\frac{1}{\sqrt{n}}\|  \widetilde   m_n( \alpha) \| + \frac{1}{\sqrt{n}}\|  \widehat m_n(  \alpha_0)  \| \right]\cr
				&\leq& O_P(\mu_n \bar\delta_n)=  o(n^{-1}).
			\end{eqnarray*}
			Finally,  by  part (i) $|\widetilde  Q_n(\alpha) -  Q_n(\alpha)   |=o_P(n^{-1})$  uniformly in $\alpha$.

		\end{proof}
	}
	
	\begin{lem}
		{\small \label{lc.2} Suppose $\sup_{\tau \in (0,1)} \sup_{\alpha\in\mathcal{C}_n} \mathbb{E }\left[
			\frac{d ^2 m(X_t,\alpha_0+\tau(\alpha-\alpha_0))}{d\tau^2}|\right]^2
			=o(n^{-1})$ and }
		
		{\small $\mathbb{E }\sup_{\alpha\in\mathcal{C}_n}\sup_{|\tau|\leq Cn^{-1/2}}%
			\frac{1}{n}\sum_t\left[\frac{d^2}{d\tau^2} m(X_t, \alpha+ \tau u_n)| \right]%
			^2 =O(1)$. Then uniformly for $\alpha\in\mathcal{C}_n$, }
		
		{\small (i) $\sup_{ \alpha\in\mathcal{C}_n} \sup_{|s|\leq 1, |x|\leq
				Cn^{-1/2}}\left|\frac{1}{n}\sum_t\ell(X_t,
			\alpha+sxu_n)\widehat\Sigma(X_t)^{-1}\frac{d^2}{d\tau^2} \widetilde m(X_t,
			\alpha+\tau xu_n)|_{\tau=s}\right|=o_P(1)$. }
		
		{\small (ii) $\sup_{\alpha\in\mathcal{C}_n} \sqrt{n} |\frac{1}{n}
			m_n(\alpha)^{\prime }\Sigma_n^{-1}\frac{d m_n(\alpha_0)}{d\alpha}%
			[u_n]-\langle u_n,\alpha-\alpha_0\rangle|=o_P(1)$. }
	\end{lem}
	
	{\small
		\begin{proof}
			(i)  We have that $\left|\frac{1}{n}\sum_t\ell(X_t,\alpha+sxu_n)\widehat\Sigma(X_t)^{-1}\frac{d^2}{d\tau^2}    \widetilde   m(X_t,\alpha+\tau xu_n)|_{\tau=s}\right|^2\leq O_P(1) AB$ where
			$$
			A:=       \frac{1}{n}\sum_t\ell(X_t, \alpha+sxu_n)^2 ,\quad B:= \frac{1}{n}\sum_t\frac{d^2}{d\tau^2}  \widetilde    m(X_t, \alpha+\tau xu_n)|_{\tau=s}^2.
			$$
			Let $m_n$ and $\rho_n$ denote the $n\times 1$ vectors of $m(X_t,\cdot)$ and $\rho(Y_{t+1},\alpha_0)$.   Uniformly for $\alpha\in\mathcal C_n$,
			$$
			A\leq     \frac{2}{n} \| P_n m_n(\alpha+sxu_n)\|^2+   \frac{2}{n} \| P_n\rho_n\|^2
			=o_P(1).
			$$
			We have $B \leq O_P(1)\mathbb E   \sup_{\alpha\in\mathcal C_n} \sup_{|\tau|\leq Cn^{-1/2} }|\frac{d^2}{d\tau^2} m(X_t, \alpha+ \tau u_n)  |^2=O_P(1)$.
			
			(ii)   By the second order mean value theorem, for some $\xi\in(0,1)$,
			\begin{eqnarray*}
				&&     \frac{1}{n}   m_n(\alpha)'\Sigma_n^{-1}\frac{d  m_n(\alpha_0)}{d\alpha}[u_n]
				=  \frac{1}{n} \sum_t[m(X_t,\alpha)-  m(X_t,\alpha_0) ]\Sigma(X_t)^{-1}\frac{d  m(X_t,\alpha_0)}{d\alpha}[u_n] \cr
				&=&\frac{1}{n} \sum_t  f(X_t)- \mathbb Ef(X_t) + \mathbb E[m(X_t,\alpha)-  m(X_t,\alpha_0) ]\Sigma(X_t)^{-1}\frac{d  m(X_t,\alpha_0)}{d\alpha}[u_n]\cr
				&=&\frac{1}{n} \sum_t  f(X_t)- \mathbb Ef(X_t) +   \mathbb E \frac{d  m(X_t,\alpha_0)}{d\alpha}  [\alpha-\alpha_0]  \Sigma(X_t)^{-1}   \frac{d  m(X_t,\alpha_0)}{d\alpha}   [u_n]      \cr
				&&+     \frac{1}{2} \mathbb E  \frac{d ^2 m(X_t,\alpha_0+\tau(\alpha-\alpha_0))}{d\tau^2}|_{\tau=\xi}    \Sigma(X_t)^{-1}\frac{d  m(X_t,\alpha_0)}{d\alpha}[u_n] \cr
				&=&\frac{1}{n} \sum_t  f(X_t)- \mathbb Ef(X_t) +   \langle u_n,\alpha-\alpha_0\rangle+o(n^{-1/2})
				= \langle u_n,\alpha-\alpha_0\rangle+o_P(n^{-1/2})
			\end{eqnarray*}
			where $f(X_t)=[m(X_t,\alpha)-  m(X_t,\alpha_0) ]\Sigma(X_t)^{-1}\frac{d  m(X_t,\alpha_0)}{d\alpha}[u_n]$ and the last equality follows from
			\begin{equation}\label{eqc.4}
			\sup_{f\in\mathcal E_n} |\frac{1}{\sqrt{n}}\sum_t (f(X_t)-\mathbb E f(X_t))|=o_P(1)
			\end{equation}
			with  $\mathcal  E_n:=\{ m(X_t,\alpha)\Sigma(X_t)^{-1}\frac{d  m(X_t,\alpha_0)}{d\alpha}[u_n]: \alpha\in\mathcal C_n\}$ and that $m(X_t,\alpha_0)=0$.

		\end{proof}
	}

	\subsection{{\protect\small Proof of Theorem \protect\ref{th4.1} }}
	
	{\small
		\begin{proof}
			By the Riesz representation Theorem, there is $v_n^*\in cl\{\mathcal A_n-\alpha_0\}$
			$$
			\frac{d\phi(\alpha_0)}{d\alpha}[\widehat\alpha-\alpha_0]=\langle v_n^*,\widehat\alpha-\alpha_0\rangle.
			$$

			Next, we show $\sqrt{n}\langle u_n,\widehat\alpha-\alpha_0\rangle\to^d\mathcal N(0,1)$, or more precisely, for $Z_n\to^d\mathcal N(0,1)$,
			\begin{equation}\label{eqc.2}
			Z_n+  \sqrt{n} \langle u_n,\widehat\alpha-\alpha_0\rangle=o_P(1).
			\end{equation}
			The   proof of Theorem \ref{th3.1}    implies that for any $\epsilon>0$,  there is $C>0$ so that with probability at least $1-\epsilon$,   $\widehat\alpha\in\mathcal A_{osn}:=\{ \alpha\in\mathcal A_n: Q(\alpha)\leq C\bar \delta_n^2, \|\alpha-\alpha_0\|_{\infty,\omega}\leq C\delta_n\}$. We now condition on this event.
			By Proposition \ref{assLQA},
			\begin{equation}\label{eqc.4}
			\sup_{ \alpha\in\mathcal A_{osn}}   \sup_{|x|\leq Cn^{-1/2}}|Q_n(\alpha+xu_n)-Q_n(\alpha)- A_n(\alpha(x))|=o_P(n^{-1}),
			\end{equation}
			where   $ A_n(\alpha(x)):= 2 x[n^{-1/2} Z_n+\langle u_n,\alpha-\alpha_0\rangle]+ B_nx^2$  with $B_n=O_P(1)$, $Z_n\to^d\mathcal N(0,1)$. Write $u_n=(u_{\gamma}, u_h)$.  Now let $\Delta_n$ be such that $\sup_{|x|\leq Cn^{-1/2}}|P_{en}(\pi_n(\widehat h+xu_h))- P_{en}(\widehat h)|=O_P(\Delta_n)$. Then
			$$E_n:=\lambda_n P_{en}(\pi_n(\widehat h+xu_h))-\lambda_n P_{en}(\widehat h)
			=O_P(\lambda_n \Delta_n).
			$$
			Now by definition, $\pi_n(\widehat\alpha+xu_n)\in\mathcal A_n$, hence
			\begin{eqnarray*}
				0&\leq& Q_n(\pi_n(\widehat\alpha+xu_n))-Q_n(\widehat\alpha)+ E_n
				\cr
				& \leq& Q_n(\widehat\alpha+xu_n)-Q_n(\widehat\alpha)+ E_n +|Q_n(\widehat\alpha+xu_n)-Q_n(\pi_n(\widehat\alpha+xu_n))|\cr
				&\leq &  Q_n(\widehat\alpha+xu_n)-Q_n(\widehat\alpha)+ E_n+o_P(n^{-1})\cr
				&\leq&2 x[ n^{-1/2} Z_n+\langle u_n,\widehat\alpha-\alpha_0\rangle]+ B_nx^2+ E_n+o_P(n^{-1}),
			\end{eqnarray*}
			where the third inequality follows from Lemma \ref{lc.1} and the last inequality follows from (\ref{eqc.4}).

			By the assumption $\lambda_n\Delta_n=o_P(n^{-1})$. Hence there is $\eta_n=o(n^{-1})$, so that
			$$
			0\leq x[ n^{-1/2} Z_n+\langle u_n,\widehat\alpha-\alpha_0\rangle]+ B_nx^2+O_P(\eta_n).
			$$
			From $n^{1/2}\eta_n=o(n^{-1/2})$, we  can find $\epsilon_n\to0^+$ so that $n ^{1/2}\eta_n\ll \epsilon_n\ll n^{-1/2} $. Set $x\in\{\epsilon_n, -\epsilon_n\}$.   Multiply by $(2\epsilon_n )^{-1} n^{1/2}$ on both sides,
			\begin{eqnarray*}
				-\frac{1}{2}  \sqrt{n} B_n\epsilon_n \leq     Z_n+  \sqrt{n}\langle u_n,\widehat\alpha-\alpha_0\rangle+O_P(\eta_n\epsilon_n^{-1}    n^{1/2})  \leq \frac{1}{2} \sqrt{n} B_n\epsilon_n  .
			\end{eqnarray*}
			We have $\eta_n\epsilon_n^{-1}n^{1/2}+\sqrt{n}B_n\epsilon_n=o_P(1)$.
			Therefore we reach
			$
			Z_n+  \sqrt{n} \langle u_n,\widehat\alpha-\alpha_0\rangle=o_P(1),
			$
			which implies $\sqrt{n} \langle u_n,\widehat\alpha-\alpha_0\rangle=  -Z_n+o_P(1)   \to^d\mathcal N(0,1)$.

			Finally, let $\zeta_n= \|v_n^*\|    n^{-1/2}$.  Apply Assumption \ref{asssmoothphi} with  $\alpha=\widehat\alpha$ and   $u_n=v_n^*/\|v_n^*\| ,$
			\begin{eqnarray*}
				\zeta_n^{-1}(\phi(\widehat\alpha)-\phi(\alpha_0)) &=&  \zeta_n^{-1}   \frac{d\phi(\alpha_0)}{d\alpha}[\widehat\alpha-\alpha_0]+o_P(1)
				\cr
				&=&     \zeta_n^{-1}   \frac{d\phi(\alpha_0)}{d\alpha}[\widehat\alpha-\alpha_{0,n}]
				+   \zeta_n^{-1}   \frac{d\phi(\alpha_0)}{d\alpha}[  \alpha_{0,n}-\alpha_0]
				+o_P(1)\cr
				&=&    \sqrt{n} \langle u_n ,\widehat\alpha-\alpha_{0,n}\rangle
				+o_P(1)\cr
				&=&   \sqrt{n} \langle u_n ,\widehat\alpha-\alpha_0\rangle+o_P(1)\to^d\mathcal N(0,1).
			\end{eqnarray*}
			where in the last equality we used $\sqrt{n} \langle u_n , \alpha_{0,n}-\alpha_0\rangle=0$ because $\alpha_{0,n}$ is the projection (under $\|.\|$) of $\alpha_0$ onto span$\{\mathcal A_n\}$ and $u_n\in$span$\{\mathcal A_n\}$.

		\end{proof}
	}
	
	\subsection{{\protect\small Proof of Theorem \protect\ref{th4.2}}}
	
	{\small \label{sec:pr4.2} }
	
	{\small
		\begin{proof} We divide the proof in the following steps.
			
			\textbf{Step 1: decompose $\widehat\gamma$.} Write $\sigma^2:=\var\left(\frac{1}{\sqrt{n}}\sum_t\mathcal W_t\right) +\|v_n^*\| ^2$, which will be shown to be the asymptotic variance. 
	 Also write    $b_n(\alpha)$ and $\bar b_n(\alpha) $ respectively as  the $n\times 1$ vectors of $b(S_t,\alpha):= l(h(W_t))\rho(Y_{t+1}, \alpha)$  and $\bar b(X_t,\alpha):=\mathbb E(l(h(W_t))\rho(Y_{t+1}, \alpha)|\sigma_t(\mathcal X))$.
			Then
			\begin{eqnarray*}
				\widehat\gamma-\gamma&=&[  \phi_n(\alpha_0) -\phi( \alpha_0)]+  [   \phi(\widehat\alpha)-\phi(\alpha_0) ]+ \frac{1}{n}\sum_{t=1}^n (  \Gamma(X_t) - \widehat  \Gamma_t) \rho(Y_{t+1},\widehat  \alpha)+a_1,\cr
			a_1&:=&\phi_n(\widehat\alpha) -\phi(\widehat\alpha)  -[ \phi_n(\alpha_0) -\phi( \alpha_0)   ] \cr
		\phi_n(\alpha)&=&\frac{1}{n}\sum_tl(h(W_t)) -
	\Gamma(X_t) \rho(Y_{t+1}, \alpha)\cr
\phi(\alpha)&=&\mathbb E \phi_n(\alpha).
			\end{eqnarray*}
			Bounding $a_1$ is based on the stochastic equicontinuity of $\phi_n-\phi$, established in Lemma  \ref{lf.1}, which yields $a_1=O_P(d_n\delta_n^{\eta})=o_P(\sigma n^{-1/2})$ by the assumption that $d_n\delta_n^{\eta}=o(\sigma n^{-1/2})$.
			
			

			\textbf{Step 2: decompose $\widehat\Gamma(X_t)$.}
			We have
			$
			\Gamma(X_t)=\bar b(X_t,\alpha_0) \Sigma(X_t)^{-1}.
			$
			Let $\widetilde\alpha\in\mathcal C_n$ denote the estimated $\alpha_0$ used in defining $\widehat\Gamma_t.$ Then 
			$$
			\widehat\Gamma_t = \Psi(X_t)'(\Psi_n'\Psi_n)^{-1}\Psi_n' b_n(\widetilde\alpha)\widehat\Sigma(X_t)^{-1}.$$
			We   then achieve the following decomposition:  \\ $ \frac{1}{n}\sum_{t=1}^n (  \widehat  \Gamma_t -\Gamma(X_t)) \rho(Y_{t+1},\widehat  \alpha)=
			\frac{	1}{n} b_n(\widetilde\alpha)' P_n\widehat\Sigma_n^{-1}\rho_n(\widehat\alpha) - \frac{	1}{n} \bar b_n( \alpha_0)'  \Sigma_n^{-1}\rho_n(\widehat\alpha) =
			a_2+... +a_8    $ where
			\begin{eqnarray}\label{eqc.3afda}
			a_2&:=& \frac{	1}{n}[ b_n(\widetilde\alpha)-\bar b_n(\widetilde\alpha)]' P_n\widehat\Sigma_n^{-1}\rho_n(\widehat\alpha)  \cr
			a_3&:=&  \frac{	1}{n}  [ \bar b_n(\widetilde\alpha)-\bar b_n(\alpha_0)]' P_n\widehat\Sigma_n^{-1}\rho_n(\widehat\alpha) \cr
			a_4&:=&  \frac{	1}{n}   \bar b_n( \alpha_0)' (P_n-I)(\widehat\Sigma_n^{-1}-\Sigma_n^{-1})\rho_n(\widehat\alpha) \cr
			a_5&:=&  \frac{	1}{n}   \bar b_n( \alpha_0)' (P_n-I)\Sigma_n^{-1}(\rho_n(\widehat\alpha) -m_n(\widehat\alpha))\cr
			a_6&:=&  \frac{	1}{n}   \bar b_n( \alpha_0)' (P_n-I)\Sigma_n^{-1}m_n(\widehat\alpha) \cr
			a_7&:=&  \frac{	1}{n}   \bar b_n( \alpha_0)'  (\widehat\Sigma_n^{-1}-\Sigma_n^{-1})(\rho_n(\widehat\alpha) -\rho_n(\alpha_0))\cr
			a_8&:=&  \frac{	1}{n}   \bar b_n( \alpha_0)'  (\widehat\Sigma_n^{-1}-\Sigma_n^{-1})\rho_n( \alpha_0)
			\end{eqnarray}
			Lemma \ref{lf.1}  shows
			$ a_2+...+a_7=O_P ( \delta_n^{\eta} \sup_x|\widehat\Sigma(x)-\Sigma(x)|  +\sqrt{k_n} d_n\delta_n^{\eta}
			+\varphi_n^2)$, which is $o_P(\sigma n^{-1/2}).$
			The bound for $a_8=o_P(\sigma n^{-1/2})$ is from Assumption \ref{a5.555}.
			Hence $$
			\frac{1}{n}\sum_{t=1}^n (  \widehat  \Gamma_t -\Gamma(X_t)) \rho(Y_{t+1},\widehat  \alpha) = o_P(\sigma n^{-1/2}).
			$$

			\textbf{Step 3: Complete proofs.}
			By the same proof of that of Theorem \ref{th4.1},   \begin{eqnarray*}
				\phi(\widehat\alpha)-\phi(\alpha_0)&=&     \|v_n^*\|  \langle u_n ,\widehat\alpha-\alpha_0\rangle+o_P(   \|v_n^*\| n^{-1/2}) \cr
				&=&-  \|v_n^*\| n^{-1/2}Z_n+o_P(  \|v_n^*\| n^{-1/2})\cr
				&=&-   \frac{1}{n}  \sum_t \mathcal Z_t
				+o_P(  \|v_n^*\| n^{-1/2})\cr
				\mathcal Z_t&:=&\rho(Y_{t+1}, \alpha_0)\Sigma(X_t)^{-1}\frac{d  m(X_t,\alpha_0)}{d\alpha}[v_n^*]  \cr
				\phi_n(\alpha_0)-\phi(\alpha_0)&=&  \frac{1}{n}\sum_{t=1}^n \mathcal W_t-\mathbb E  \mathcal W_t ,\quad \mathcal W_t:=  l(h_0(W_t))- \Gamma(X_t) \rho(Y_{t+1},\alpha_0).
			\end{eqnarray*}
			Putting together,   $\widehat\gamma-\gamma=[  \phi_n(\alpha_0) -\phi( \alpha_0)]+  [   \phi(\widehat\alpha)-\phi(\alpha_0) ]  + o_P(\sigma n^{-1/2}) $, and
			$$ [  \phi_n(\alpha_0) -\phi( \alpha_0)]+  [   \phi(\widehat\alpha)-\phi(\alpha_0) ]  =  \frac{1}{n}\sum_{t=1}^n \mathcal W_t-\mathbb E  \mathcal W_t   -    \mathcal Z_t
			+o_P(\sigma n^{-1/2}).
			$$


			Next, $\mathcal W_t-\mathbb E  \mathcal W_t   -    \mathcal Z_t $  is strictly stationary, satisfying the strong mixing condition (Assumption \ref{ass3.10}) with $\sum_{n=1}^{\infty}\alpha(n)^{\zeta/(2+\zeta)}\leq C \sum_{n=1}^{\infty}\exp(-c\zeta n/(2+\zeta))<\infty $ for any constant $\zeta>0.$ In addition,
			\begin{eqnarray*}
				&&  \mathbb E\left| (\mathcal W_t-\mathbb E  \mathcal W_t   -    \mathcal Z_t )\sigma^{-1} \right|^{2+\zeta}
				\leq C \mathbb E\left| \mathcal W_t\sigma^{-1} \right|^{2+\zeta}
				+C \mathbb E\left|     \mathcal Z_t  \|v_n^*\|^{-1} \right|^{2+\zeta}
				\cr
				&\leq& C\mathbb E\left|\rho(Y_{t+1},\alpha_0) \frac{d  m(X_t,\alpha_0)}{d\alpha}[u_n] \right|^{2+\zeta}
				+C\mathbb E\left|\rho(Y_{t+1},\alpha_0)  \right|^{2+\zeta}<C.
			\end{eqnarray*}
			Then by Theorem 1.7 of \cite{ibragimov1962some},  \begin{equation}\label{eqc.d3}
			\sqrt{n}\sigma^{-1} \left[  \phi_n(\alpha_0) -\phi( \alpha_0)  +    \phi(\widehat\alpha)-\phi(\alpha_0)   \right]\to\mathcal N(0,1).
			\end{equation}
			This implies   the asymptotic normality of $\widehat\gamma-\gamma$.
			
		\end{proof}
	}
	
	\begin{lem}[for Theorems \protect\ref{th4.2}, \protect\ref{thqlr:eff}]
		{\small \label{lf.1} Recall that $b_n(\alpha)$ and $\bar b_n(\alpha) $ are
			the $n\times 1$ vectors of $l(h(W_t))\rho(Y_{t+1}, \alpha)$ and $\mathbb{E}%
			(l(h(W_t))\rho(Y_{t+1}, \alpha)|\sigma_t(\mathcal{X}))$. Suppose }
		
		{\small (a) $\sup_x|\Gamma(x)|^2 +\sup_w\sup_{\mathcal{H}_n}l(h(w))^2<C.$ }
		
		{\small (b) $| l(h_1(w))-l(h_2(w))|\leq C  |h_1(w)-h_2(w)|$ uniformly for
			all $h_1, h_2\in \mathcal{H}_n$,and $w$. }
		
		{\small (c) $\mathbb{E}\sup_{\alpha\in\mathcal{C}_n} |
			l(h(W_t))-l(h_0(W_t))|^2\leq C\delta_n^{2\eta}$. }
		
		{\small
		}
		
		{\small (d) $\mathbb{E }\sup_{\alpha\in\mathcal{C}_n} (\rho(Y_{t+1},
			\alpha)-\rho(Y_{t+1}, \alpha_0) )^2=C\delta_n^{2\eta} $ for some $\eta, C>0.$
		}
		
		{\small (e) For some $\kappa, C>0$ , $\mathbb{E }\sup_{
				\|\alpha_1-\alpha\|_{\infty,\omega}<\delta}
			|\epsilon_t(\alpha_1)-\epsilon_t(\alpha)|^2\leq C\delta^{2\kappa} $ for all $%
			\delta>0$ and $\alpha,\alpha_1\in cl\{a+xb: a, b\in\mathcal{A}_n, x\in%
			\mathbb{R}\}$. }
		
		{\small (f) $\frac{ 1}{\sqrt{n}} \| \bar b_n( \alpha_0)^{\prime }(P_n-I) \|=
			O_P(\varphi_n ). $ }
		
		{\small Then for $\bar \epsilon_n(\alpha)$ as the $n\times 1$ vector of $%
			\rho(Y_{t+1},\alpha)- m(X_t,\alpha)$, }
		
		{\small (i) $\sup_{\alpha_1,\alpha_2\in\mathcal{C}_n\cup\{\alpha_0\}}|%
			\phi_n(\alpha_1) -\phi( \alpha_1) -[ \phi_n(\alpha_2) -\phi( \alpha_2) ]|
			=O_P(d_n\delta_n^{\eta}).$ }
		
		{\small
		}
		
		{\small (ii) $\sup_{\mathcal{C}_n}| \frac{1}{n} \bar b_n( \alpha_0)^{\prime
			}(\widehat\Sigma_n^{-1}-\Sigma_n^{-1})[\rho_n(\alpha)-\rho_n(%
			\alpha_0)]|=O_P(\delta_n^{\eta})\sup_x|\widehat\Sigma(x)-\Sigma(x)|.$ }
		
		{\small (iii) $\sup_{\mathcal{C}_n}\frac{1}{\sqrt{n}}\|
			P_n\widehat\Sigma_n^{-1}\rho_n(\alpha)
			\|=O_P(\sup_x\|\widehat\Sigma(x)-\Sigma(x)\|+d_n\delta_n^{\eta}+
			\bar\delta_n ).$ }
		
		{\small (iv) $\sup_{\mathcal{C}_n} \frac{ 1}{n} \bar b_n( \alpha_0)^{\prime
			}(P_n-I)\Sigma_n^{-1}\bar \epsilon_n(\alpha)=O_P(\sqrt{k_n}
			d_n\delta_n^{\eta}).$ }
		
		{\small (v) $\sup_{\mathcal{C}_n} \frac{ 1}{n}[ b_n(\alpha)-\bar
			b_n(\alpha)]^{\prime }P_n\widehat\Sigma_n^{-1}\rho_n(\alpha)+ \sup_{\mathcal{%
					C}_n} \frac{ 1}{n} [ \bar b_n(\alpha)-\bar b_n(\alpha_0)]^{\prime
			}P_n\widehat\Sigma_n^{-1}\rho_n(\alpha)$ }
		
		{\small $=O_P( \delta_n^{\eta} \sup_x\|\widehat\Sigma(x)-\Sigma(x)\|+ d_n
			\delta_n^{2\eta} + \sqrt{k_n}d_n\delta_n^{\eta}\bar\delta_n + \sqrt{k_n/n}
			\bar\delta_n).$ }
		
		{\small (vi) $\sup_{\mathcal{C}_n} \frac{ 1}{n} \bar b_n( \alpha_0)^{\prime
			}(P_n-I)(\widehat\Sigma_n^{-1}-\Sigma_n^{-1})\rho_n(\alpha) = O_P(\varphi_n
			\sup_x\|\widehat\Sigma(x)-\Sigma(x)\| )$ }
		
		{\small (vii) $\sup_{\mathcal{C}_n} \frac{ 1}{n} \bar b_n( \alpha_0)^{\prime
			}(P_n-I)\Sigma_n^{-1}m_n(\alpha) =O_P(\varphi_n^2)$. }
	\end{lem}
	
	{\small
		\begin{proof}
			(i)	First recall $\epsilon (S_t, \alpha)=\rho(Y_{t+1}, \alpha)-m(X_t, \alpha)$.
			\begin{eqnarray*}
				a&:=&\sup_{\alpha_1,\alpha_2\in\mathcal C_n\cup\{\alpha_0\}}| \frac{1}{n}\sum_{t=1}^n   \Gamma(X_t) [\rho(Y_{t+1},\alpha_1) - \rho(Y_{t+1},\alpha_2)]-\mathbb E     \Gamma(X_t) [\rho(Y_{t+1},\alpha_1) - \rho(Y_{t+1},\alpha_2)]|\cr
				&=&\sup_{\alpha_1,\alpha_2\in\mathcal C_n\cup\{\alpha_0\}}| \frac{1}{n}\sum_{t=1}^n   \Gamma(X_t) [\epsilon(S_t,\alpha_1)- \epsilon(S_t,\alpha_2)]| \cr
				&\leq&2\sup_{\alpha\in \mathcal C_n\cup\{\alpha_0\}}| \frac{1}{n}\sum_{t=1}^n   \Gamma(X_t) [\epsilon(S_t,\alpha)- \epsilon(S_t,\alpha_0)]|.\cr
				b&:=&\sup_{\alpha_1,\alpha_2\in\mathcal C_n\cup\{\alpha_0\}}| \frac{1}{n}\sum_{t=1}^n l(h_1(W_t))-l(h_2(W_t))-\mathbb E [l(h_1(W_t))-l(h_2(W_t))]|\cr
				&\leq&2	\sup_{\alpha \in\mathcal C_n\cup\{\alpha_0\}}| \frac{1}{n}\sum_{t=1}^n l(h(W_t))-l(h_0(W_t))-\mathbb E [l(h(W_t))-l(h_0(W_t))]|.
			\end{eqnarray*}

			Note $\mathbb E\sup_{\alpha\in\mathcal C_n} \Gamma(X_t) ^2[\epsilon(S_t,\alpha)- \epsilon(S_t,\alpha_0)]^2\leq C \mathbb E\sup_{\alpha\in\mathcal C_n}  [\epsilon(S_t,\alpha)- \epsilon(S_t,\alpha_0)]^2\leq C\delta_n^{2\eta}  $, $\eta\leq 1$.
			Then the convergence of $a$ and $b$ follow from the same argument of that of Proposition \ref{lem5.2fd} with $\Psi_j(X_t)$ replaced with $\Gamma(X_t)$. Term $b$ follows from the same proof of this Proposition.  We reach $a+b=O_P(d_n\delta_n^{\eta})$.

			Therefore $\sup_{\alpha_1,\alpha_2\in\mathcal C_n\cup\{\alpha_0\}}|\phi_n(\alpha_1) -\phi( \alpha_1)  -[ \phi_n(\alpha_2) -\phi( \alpha_2)   ]|  \leq a+b=O_P(d_n\delta_n^{\eta}).$

			(ii)	 First,
			$
			\mathbb E	\sup_{\alpha\in\mathcal C_n}[ \rho(Y_{t+1},\alpha)-\rho(Y_{t+1},\alpha_0)]^2\leq O( \delta_n^{2\eta}).
			$
			This implies $\frac{1}{\sqrt{n}}\|\rho_n(\alpha)-\rho_n(\alpha_0)\|=O_P( \delta_n^{\eta}).$ The target of interest is then bounded by
			$$
			\|\widehat\Sigma_n- \Sigma_n\| \frac{1}{\sqrt{n}}\|\rho_n(\alpha)-\rho_n(\alpha_0)\|=O_P(\delta_n^{\eta})\sup_x\|\widehat\Sigma(x)-\Sigma(x)\|.
			$$
			
			
			(iii)  First,  write $\widetilde{m}_\Sigma(X_t,\alpha):= \Psi(X_t)'(\Psi_n'\Psi_n)^{-1}\Psi_n'\Sigma_n^{-1}m_n(\alpha)$. Then step 1 of the proof of Theorem \ref{th3.1} carries over, leading to
			\begin{eqnarray*}
				&& \sup_{\mathcal C_n}\frac{1}{ n}\| P_n \Sigma_n^{-1} m_n(\alpha) \|^2
				=\sup_{\mathcal C_n}\frac{1}{ n}\sum_t\widetilde m_{\Sigma}(X_t,\alpha)^2
				\leq C\sup_{\mathcal C_n}\mathbb E\widetilde{m}_\Sigma(X_t,\alpha)^2\cr
				&\leq&C\sup_{\mathcal C_n}\mathbb E[\widetilde{m}_\Sigma(X_t,\alpha)-m(X_t,\alpha)\Sigma(X_t)^{-1}]^2
				+C\sup_{\mathcal C_n}\mathbb E m(X_t,\alpha)^2 =O_P(\bar\delta_n^2).
			\end{eqnarray*}
			Also,
			for $\bar\epsilon_n(\alpha):=\rho_n(\alpha)-m_n(\alpha)$, the first inequality below follows from the same proof of Proposition \ref{lem5.2fd},
			\begin{eqnarray*}
				&&\sup_{\mathcal C_n}\frac{1}{\sqrt{n}}\| P_n \Sigma_n^{-1}(\bar\epsilon_n(\alpha)-\bar\epsilon_n(\alpha_0))\|
				\leq O_P(d_n\delta_n^{\eta})
				\cr
				&&\sup_{\mathcal C_n}\frac{1}{\sqrt{n}}\| P_n(\widehat\Sigma_n^{-1}-\Sigma_n^{-1})(\rho_n(\alpha) -\rho_n(\alpha_0))\|
				\leq O_P(\delta_n^{\eta})\sup_x\|\widehat\Sigma(x)-\Sigma(x)\|\cr
				&&\sup_{\mathcal C_n}\frac{1}{\sqrt{n}}\| P_n \Sigma_n^{-1}(\rho_n(\alpha) -\rho_n(\alpha_0))\|
				\leq \sup_{\mathcal C_n}\frac{1}{\sqrt{n}}\| P_n \Sigma_n^{-1}(\bar\epsilon_n(\alpha)-\bar\epsilon_n(\alpha_0))\|
				+\sup_{\mathcal C_n}\frac{1}{\sqrt{n}}\| P_n \Sigma_n^{-1} m_n(\alpha) \|\cr
				&\leq& O_P(d_n\delta_n^{\eta}+ \bar\delta_n)\cr
				&&\sup_{\mathcal C_n}\frac{1}{\sqrt{n}}\| P_n \widehat \Sigma_n^{-1}(\rho_n(\alpha) -\rho_n(\alpha_0))\|
				\cr
				&\leq& \sup_{\mathcal C_n}\frac{1}{\sqrt{n}}\| P_n(\widehat\Sigma_n^{-1}-\Sigma_n^{-1})(\rho_n(\alpha) -\rho_n(\alpha_0))\|+\sup_{\mathcal C_n}\frac{1}{\sqrt{n}}\| P_n \Sigma_n^{-1}(\rho_n(\alpha) -\rho_n(\alpha_0))\|\cr
				&\leq& O_P(\delta_n^{\eta})\sup_x\|\widehat\Sigma(x)-\Sigma(x)\|+O_P(d_n\delta_n^{\eta}+ \bar\delta_n)\cr
				&& \frac{1}{\sqrt{n}}\| P_n\widehat\Sigma_n^{-1}\rho_n(\alpha_0)\|
				=O_P(1)\sup_x\|\widehat\Sigma(x)-\Sigma(x)\|+O_P(\sqrt{k_n/n}).
			\end{eqnarray*}
			Together, $
			\sup_{\mathcal C_n}\frac{1}{\sqrt{n}}\| P_n\widehat\Sigma_n^{-1}\rho_n(\alpha) \|
			\leq \sup_{\mathcal C_n}\frac{1}{\sqrt{n}}\| P_n \widehat \Sigma_n^{-1}(\rho_n(\alpha) -\rho_n(\alpha_0))\|
			+\frac{1}{\sqrt{n}}\| P_n\widehat\Sigma_n^{-1}\rho_n(\alpha_0)\|
			$
			whose final rate is $O_P(\sup_x\|\widehat\Sigma(x)-\Sigma(x)\|+d_n\delta_n^{\eta}+ \bar\delta_n )$.

			(iv) First,  $ \frac{	1}{n}   \bar b_n( \alpha_0)' (P_n-I)\Sigma_n^{-1}\bar \epsilon_n(\alpha_0) = O_P(\varphi_n n^{-1/2}). $
			Next,  from the proof of Proposition \ref{lem5.2fd},
			\begin{eqnarray*}
				&&\sup_{\mathcal C_n} \frac{	1}{n}   \bar b_n( \alpha_0)' (P_n-I)\Sigma_n^{-1}[\bar \epsilon_n(\alpha)-\bar \epsilon_n(\alpha_0)]\cr
				&\leq&  \sup_{\mathcal C_n} \frac{	1}{n}  \| \Psi_n' \Sigma_n^{-1}[\bar \epsilon_n(\alpha)-\bar \epsilon_n(\alpha_0)]\|
				+\sup_{\mathcal C_n} \frac{	1}{n}   \bar b_n( \alpha_0)'  \Sigma_n^{-1}[\bar \epsilon_n(\alpha)-\bar \epsilon_n(\alpha_0)]=O_P(\sqrt{k_n} d_n\delta_n^{\eta}).
			\end{eqnarray*}
			So $\sup_{\mathcal C_n} \frac{	1}{n}   \bar b_n( \alpha_0)' (P_n-I)\Sigma_n^{-1}\bar \epsilon_n(\alpha)=
			O_P(\sqrt{k_n} d_n\delta_n^{\eta}+ \varphi_nn^{-1/2})=O_P(\sqrt{k_n} d_n\delta_n^{\eta}).
			$

			(v)  The same proof of Proposition \ref{lem5.2fd} carries over to here. So
			$$\sup_{\alpha\in\mathcal C_n}\frac{1}{n}\|\Psi_n'(\bar b_n(\alpha)-b_n(\alpha))- \Psi_n'(\bar b_n(\alpha_0)-b_n(\alpha_0)) \| =O_P(\sqrt{k_n} d_n\delta_n^\eta).$$
			In addition, $\frac{1}{n}\|  \Psi_n'(\bar b_n(\alpha_0)-b_n(\alpha_0)) \| =O_P(\sqrt{k_n} n^{-1/2}).$
			This implies
			$$\sup_{\mathcal C_n} \frac{	1}{\sqrt{n}}[ b_n(\alpha)-\bar b_n(\alpha)]' P_n
			\leq O_P(1)\sup_{\alpha\in\mathcal C_n}\frac{1}{n}\|\Psi_n'(\bar b_n(\alpha)-b_n(\alpha))\|=O_P( d_n\delta_n^{\eta}+n^{-1/2}) \sqrt{k_n}.
			$$
			Also,
			\begin{eqnarray*}
				\frac{1}{n} \|\bar b_n(\alpha) -\bar b_n( \alpha_0)\|^2
				&\leq& O_P(1)  \mathbb E\sup_{\mathcal C_n } [\mathbb E |\rho(Y_{t+1},\alpha)-  \rho(Y_{t+1},\alpha_0)| |\sigma_t(\mathcal X)]^2
				+O_P(1)\sup_{\mathcal C_n} \mathbb E  |l(h)-l(h_0)|^2   \cr
				&\leq& O_P(\delta_n^{2\eta}).
			\end{eqnarray*}
			Hence
			$\sup_{\mathcal C_n} \frac{	1}{n}[ b_n(\alpha)-\bar b_n(\alpha)]' P_n\widehat\Sigma_n^{-1}\rho_n(\alpha)=O_P(  \sqrt{k_n}d_n\delta_n^{\eta}+ \sqrt{k_n/n} )  (\sup_x\|\widehat\Sigma(x)-\Sigma(x)\|+d_n\delta_n^{\eta}+ \bar\delta_n )$
			
			and
			$ \sup_{\mathcal C_n} \frac{	1}{n}  [ \bar b_n(\alpha)-\bar b_n(\alpha_0)]' P_n\widehat\Sigma_n^{-1}\rho_n(\alpha)
			= O_P(\sup_x\|\widehat\Sigma(x)-\Sigma(x)\|\delta_n^{\eta}+d_n\delta_n^{2\eta}+ \bar\delta_n\delta_n^{\eta} )
			$.
			So the final rate of the sum of the two is $ \delta_n^{\eta} \sup_x\|\widehat\Sigma(x)-\Sigma(x)\|+
			d_n  \delta_n^{2\eta} +  \sqrt{k_n}d_n\delta_n^{\eta}\bar\delta_n
			+     \sqrt{k_n/n}  \bar\delta_n ).
			$

			(vi) (vii) The proof is straightforward.

		\end{proof}
	}
	
	\section{Proofs for Section \protect\ref{sec:qlr}}
	
	\subsection{{\protect\small Proof of Theorem \protect\ref{thqlr}}}
	
	{\small
		\begin{proof}
			Proposition \ref{assLQA} shows the following LQA:
			\begin{equation}\label{eqa.3}
			\sup_{ \alpha\in\mathcal C_n}   \sup_{|x|\leq Cn^{-1/2}}|Q_n(\alpha+xu_n)-Q_n(\alpha)- A_n(\alpha(x))|=o_P(n^{-1})
			\end{equation}
			where   $ A_n(\alpha(x)):= 2 x[n^{-1/2} Z_n+\langle u_n,\alpha-\alpha_0\rangle]+ B_nx^2$  with $B_n=1+o_P(1)$, $Z_n\to^d\mathcal N(0,1)$.
			We respectively provide lower and upper bounds for  $Q_n(\widehat \alpha^R) -Q_n(\widehat \alpha )$.
			
			\textbf{Step 1: lower bound.} To apply the LQA,  we need to first show that $\widehat\alpha^R\in\mathcal C_n$ with a high probability.
			In fact,   there is $\pi_n^R\alpha_0\in\mathcal A_n^R$ so that   $$
			Q_n(\widehat\alpha^R)+\lambda_n P_{en}(\widehat h^R)\leq Q_n(\pi_n^R\alpha_0)+\lambda_n P_{en}(\pi_n^R h_0).
			$$
			Given the  above inequalities, the proof of Theorem \ref{th3.1} carries over, establishing that $\widehat\alpha^R\in\mathcal C_n$ with a high probability.  We now condition on this event. Hence by (\ref{eqa.3}), uniformly for all $|x|\leq C n^{-1/2} $,
			\begin{eqnarray}\label{eqe.2}
			Q_n(\widehat \alpha^R+xu_n)-Q_n(\widehat \alpha^R) &=&2 x[n^{-1/2} Z_n+\langle u_n, \widehat \alpha^R-\alpha_0\rangle]+ B_nx^2+ o_P(n^{-1})\cr
			&=& 2x n^{-1/2} Z_n+ B_nx^2+ o_P(n^{-1}),
			\end{eqnarray}
			where the second equality follows from Lemma \ref{le.1}. Next, we note one technical difficulty that  the inequality $Q_n(\widehat \alpha  )+ \lambda_n P_{en}(\widehat \alpha )\leq  Q_n(\alpha)  +
			\lambda_n P_{en}(\alpha) $   may not hold for $\alpha= \widehat \alpha^R+xu_n$, as $\mathcal A_n$ is a nonlinear space so $\widehat \alpha^R+xu_n$ is not necessarily in $\mathcal A_n$. Nevertheless, we can apply this inequality for $\alpha= \pi_n( \widehat \alpha^R+xu_n)$, and show that $|Q_n(\pi_n( \widehat \alpha^R+xu_n))- Q_n( \widehat \alpha^R+xu_n)|$ is negligible. Specifically, by Lemma \ref{le.1} and Assumption \ref{ass4.3new},
			\begin{eqnarray}\label{eqe.3}
			Q_n(\widehat \alpha  )-  Q_n(\widehat \alpha^R+xu_n) &\leq &
			\lambda_n P_{en}(\pi_n(\widehat \alpha^R+xu_n)) -  \lambda_n P_{en}(\widehat \alpha )
			+ Q_n(\pi_n(\widehat \alpha^R+xu_n))  - Q_n(\widehat \alpha^R+xu_n)\cr
			&\leq&  o_P(n^{-1}).
			\end{eqnarray}
			Thus (\ref{eqe.2})  and (\ref{eqe.3}) imply
			$Q_n(\widehat \alpha^R)  -    Q_n(\widehat \alpha )\geq - 2x n^{-1/2} Z_n-B_nx^2-o_P(n^{-1}) $. Take $x=-Z_nB_n^{-1}n^{-1/2}$ which maximizes $- 2x n^{-1/2} Z_n-B_nx^2$, then
			$$Q_n(\widehat \alpha^R)  -    Q_n(\widehat \alpha )\geq
			Z_n^2B_n^{-1}n^{-1} -o_P(n^{-1}).
			$$

			\textbf{Step 2:  upper bound.}  Fix $x^*$  determined as in  Lemma \ref{le.1}, this lemma  shows that   $x^*= n^{-1/2} Z_nB_n^{-1}+o_P(n^{-1/2})$ and that $ |Q_n(\pi_n^R(\widehat \alpha +x^*u_n))  - Q_n(\widehat \alpha +x^*u_n)| =o_P(n^{-1})$. Hence by (\ref{eqa.3}) again,
			\begin{eqnarray*}
				Q_n(\widehat\alpha^R) -Q_n(\widehat\alpha) & \leq&  Q_n(\pi_n^R(\widehat \alpha +x^*u_n)) -Q_n(\widehat\alpha)+\lambda_n(P_{en}(\pi_n^R(\widehat \alpha +x^*u_n)) - P_{en}(\widehat\alpha^R)) \cr
				&=& Q_n(\widehat \alpha +x^*u_n) -Q_n(\widehat\alpha)+o_P( n^{-1})\cr
				&=& 2 x^* n^{-1/2}  [ Z_n+n^{1/2}\langle u_n,\widehat\alpha-\alpha_0\rangle]+ B_nx^{*2}+o_P( n^{-1})\cr
				&=& B_nx^{*2}+o_P( n^{-1})= Z_n^2B_n^{-1}n^{-1} +o_P(n^{-1}),
			\end{eqnarray*}
			where the third equality is due to  the proof of Theorem \ref{th4.1} that  $
			Z_n+  \sqrt{n} \langle u_n,\widehat\alpha-\alpha_0\rangle=o_P(1).
			$
			
			\textbf{Step 3:  matching bounds.}
			
			Together, we have
			$$
			S_n(\phi_0)=n(Q_n(\widehat \alpha^R)- Q_n(\widehat \alpha ))=B_n^{-1}Z_n^2 +o_P(1)\to^d \chi^2_1
			$$
			given that $B_n\to^P1$ proved in  Proposition \ref{assLQA}.
		\end{proof}
	}
	
	\begin{lem}[for Theorem \protect\ref{thqlr}]
		{\small \label{le.1} Suppose }
		
		{\small (a) $\sup_{\alpha\in\mathcal{C}_n} \frac{1}{n}\sum_{t=1}^n [m( X_t,
			\pi_n \alpha)- m( X_t, \alpha)]^2 =O_P(\mu_n^2)$ and }
		
		{\small $\sup_{\alpha\in\mathcal{C}_n, \phi(\alpha)=\phi_0} \frac{1}{n}%
			\sum_{t=1}^n [m( X_t, \pi_n^R (\alpha+xu_n))- m( X_t, \alpha+xu_n)]^2
			=O_P(\mu_n^2). $ }
		
		{\small
		}
		
		{\small (b) $\mu_n\bar \delta_n = o(n^{-1})$. }
		
		{\small (c) $t\to \phi(\alpha+t u_n)$ is continuous. }
		
		{\small Then }
		
		{\small
			(i) $\langle u_n, \widehat \alpha^R-\alpha_0\rangle=o_P(n^{-1/2}).$ }
		
		{\small (ii) $\sup_{|x|\leq Cn^{-1/2}}|Q_n(\pi_n(\widehat \alpha^R+xu_n)) -
			Q_n(\widehat \alpha^R+xu_n)| =o_P(n^{-1})$. }
		
		{\small 
		}
		
		{\small (iii) there is $x^*$ so that $\phi(\widehat\alpha +x^* u_n)=\phi_0$
			and $|Q_n(\pi_n^R(\widehat \alpha +x^*u_n)) - Q_n(\widehat \alpha +x^*u_n)|
			=o_P(n^{-1})$ and $x^*= n^{-1/2} Z_nB_n^{-1}+o_P(n^{-1/2})$. }
	\end{lem}
	
	{\small
		\begin{proof}
			
			(i)
			Note that $\phi(\widehat\alpha^R)-\phi(\alpha_0)=0$.
			By Assumption \ref{asssmoothphi}, $ \left|\frac{d\phi(\alpha_0)}{d\alpha}[\widehat\alpha^R-\alpha_0] \right|=o(\|v_n^*\|   n^{-1/2}).$
			By the Riesz representation Theorem,
			$$
			\frac{d\phi(\alpha_0)}{d\alpha}[\widehat\alpha^R-\alpha_0]=
			\frac{d\phi(\alpha_0)}{d\alpha}[\widehat\alpha^R-\alpha_{0,n}]
			+ \frac{d\phi(\alpha_0)}{d\alpha}[ \alpha_{0,n}-\alpha_0] =
			\|v_n^*\| \langle  u_n,\widehat\alpha^R-\alpha_0\rangle+ o(\|v_n^*\|   n^{-1/2})
			$$
			with the definition $u_n=v_n^*/ \|v_n^*\| $. This finishes the proof.
			
			(ii)  The proof is the same as that of   Lemma \ref{lc.1}.

			(iii) First, we prove there is $x^*$ so that $\phi(\widehat\alpha +x^* u_n)=\phi_0.$ Define $F(x):=\langle v^*_n, \alpha -\alpha_0\rangle + x  \|v_n^*\| $.  Also define $R(x):=\phi(\alpha+xu_n)-\phi(\alpha_0)$.
			By Assumption \ref{asssmoothphi}, there is a  positive sequence $b_n=o(\|v_n^*\|   n^{-1/2})$,   uniformly for all $\alpha\in\mathcal C_n$,   for all $x\leq C n^{-1/2}$,
			$$
			|R(x)-   F(x)  |  \leq b_n
			$$
			Now fix some $r$ such that $|r- \langle v^*_n, \alpha -\alpha_0\rangle   |\leq C\|v_n^*\| n^{-1/2}$ and define  $x_1=(r-\langle v^*_n, \alpha -\alpha_0\rangle-2b_n)\|v_n^*\| ^{-1}$
			and $x_2=(r-\langle v^*_n, \alpha -\alpha_0\rangle+2b_n)\|v_n^*\| ^{-1}$.  This ensures that  $ F(x_1)  +2b_n= F(x_2)-2b_n=r$ and $|x_1|+|x_2|\leq Cn^{-1/2}$. Therefore,
			$$
			R(x_1)\leq F(x_1)+b_n<r,\quad  R(x_2)\geq F(x_2)-b_n >r.
			$$
			Hence there is $x^*$ between $x_1, x_2$ so that $R(x^*)=r$. In the above proof, suppose $r=0$ and $\alpha=\widehat\alpha$ are admitted, then $\phi(\widehat\alpha+x^*u_n)=\phi(\alpha_0)$.  To show the admissibility, we note   (\ref{eqc.2}) that  $n^{-1/2}Z_n+  \|v_n^*\| ^{-1}\langle v^*_n,\widehat\alpha-\alpha_0\rangle=o_P(n^{-1/2}).$ Hence indeed, for any $\epsilon>0$, there is $C>0$,
			$$|\langle v^*_n, \alpha -\alpha_0\rangle|= \|v_n^*\| n^{-1/2}|Z_n|+o_P(\|v_n^*\| n^{-1/2})\leq C\|v_n^*\| n^{-1/2}$$
			with probability at least $1-\epsilon.$
			
			Now
			$
			|x^*-n^{-1/2} Z_n|\leq |x_1-n^{-1/2} Z_n|+|x_2-n^{-1/2} Z_n| \leq \frac{2|b_n|}{\|v_n^*\| }+o_P(n^{-1/2})=o_P(n^{-1/2}).
			$

			Finally,  the proof of $|Q_n(\pi_n^R(\widehat \alpha +x^*u_n))  - Q_n(\widehat \alpha +x^*u_n)| =
			o_P(n^{-1})
			$
			is the same as part (ii).

		\end{proof}
	}
	
	\subsection{{\protect\small Proof of Theorem \protect\ref{thqlr:eff}}}
	
	{\small
		\begin{proof}
			As in the proof of Theorem \ref{thqlr}, we respectively provide lower and upper bounds for $\frac{1}{n}\widetilde S_n(\phi_0)=L_n(\widehat\alpha^R,\phi_0)- L_n(\widehat\alpha,\widehat\gamma)$.  Note that $L_n(\widehat\alpha,\widehat\gamma)=Q_n(\widehat\alpha)$.  Let
			\begin{eqnarray*}
				g_1&=&(\widehat\phi(\widehat \alpha^R)-\gamma_0)^2 \widehat\Sigma_2^{-1}  \cr
				g_2&=&\phi(\widehat\alpha^R)-\phi(\alpha_0)\cr
				g_3&=&[  \phi_n(\alpha_0) -\phi( \alpha_0)]+  [   \phi(\widehat\alpha)- \gamma_0 ]\cr
				g_4&=& [  \phi_n(\alpha_0)  -\gamma_0
				+ g_2]^2\widehat\Sigma_2^{-1} \cr
				g_5&=&\phi_n(\alpha_0)  -\gamma_0-  \|v_n^*\|    n^{-1/2}Z_n\cr
				g_6&=& n^{-1/2} Z_n+\|v_n^*\| ^{-1}g_2
			\end{eqnarray*}
			Also note that $\widehat\alpha^R\in\mathcal C_n$ with a high probability, by Lemma \ref{ld.2}.  We now condition on this event.

			\textbf{Step 1: lower bound.}
			Due to $ \lambda_n P_{en}( \widehat \alpha^R +xu_n )  -  \lambda_n P_{en}(\widehat \alpha ) =o_P(n^{-1})$ and  $\widehat \alpha^R\in\mathcal A_n$, so uniformly for all $|x|\leq C n^{-1/2} $,
			\begin{eqnarray*}
				&&L_n(\widehat\alpha,\widehat\gamma)-L_n(\widehat\alpha^R,\phi_0)
				= Q_n(\widehat\alpha)- Q_n(\widehat\alpha^R)- g_1+  o_P(n^{-1})\cr
				&\leq^{(a)}&Q_n(\pi_n(\widehat \alpha^R+xu_n))- Q_n(\widehat \alpha^R+xu_n) +  Q_n(\widehat \alpha^R+xu_n) -   Q_n(\widehat\alpha^R)-  g_1+   o_P(n^{-1})\cr
				&=^{(b)}& Q_n(\widehat \alpha^R+xu_n) -   Q_n(\widehat\alpha^R)-  g_1+o_P(n^{-1})\cr
				&=^{(c)}& 2 x[n^{-1/2} Z_n+\langle u_n, \widehat \alpha^R-\alpha_0\rangle]+ x^2-  g_1+ o_P(n^{-1}), \cr
				&=^{(d)}& 2 x[n^{-1/2} Z_n+\|v_n^*\| ^{-1}\frac{d\phi(\alpha_0)}{d\alpha}[\widehat\alpha^R-\alpha_0] ]+ x^2-  g_1+ o_P(n^{-1}), \cr
				&=^{(e)}&  2 x[n^{-1/2} Z_n+\|v_n^*\| ^{-1}g_2]+ x^2-  g_1  + o_P(n^{-1})\cr
				&=^{(f)}&\underbrace{ 2 x[n^{-1/2} Z_n+\|v_n^*\| ^{-1}g_2]+ x^2-  g_4}_{F(x)}+ o_P(n^{-1}),
			\end{eqnarray*}
			where in (a) we used $Q_n(\widehat\alpha)\leq Q_n(\pi_n(\widehat \alpha^R+xu_n))$; (b) follows from $|Q_n(\pi_n(\widehat \alpha^R+xu_n))- Q_n(\widehat \alpha^R+xu_n) |\leq o_P(n^{-1})$ following the same proof of that of Lemma  \ref{le.1}(ii); (c) is from  (\ref{eqa.3}); (d) is from the Riesz representation: ($\langle v_n^*, \alpha_0-\alpha_{0,n}\rangle=0$)
			$$
			\frac{d\phi(\alpha_0)}{d\alpha}[\widehat\alpha^R-\alpha_0] =
			\frac{d\phi(\alpha_0)}{d\alpha}[\widehat\alpha^R-\alpha_{0,n}]
			+\frac{d\phi(\alpha_0)}{d\alpha}[\alpha_{0,n}-\alpha_0]
			=\langle v_n^*, \widehat\alpha^R-\alpha_{0,n}\rangle + o_P(n^{-1/2} \|v_n^*\|);
			$$
			(e) is from Assumption \ref{asssmoothphi}; (f) is from Lemma \ref{led.3}.

			We choose $x=x^*$ to  minimize    $F(x)$  on the right hand side, leading to the choice $x^*=-[n^{-1/2} Z_n+\|v_n^*\| ^{-1}g_2]=-g_6$. We shall verify that $|x^*|=O_P(n^{-1/2})$ in Step 3 below. Suppose for now this is true, then we have obtained the lower bound: $\frac{1}{n}\widetilde S_n(\phi_0) \geq - F(x^*)- o_P(n^{-1})$, where
			$$
			- F(x^*)=[n^{-1/2} Z_n+\|v_n^*\| ^{-1}g_2]^2+g_4 =g_6^2+g_4.
			$$
			
			\textbf{Step 2:  upper bound.}
			Uniformly for all $|x|\leq C n^{-1/2} $,
			\begin{eqnarray*}
				&&L_n(\widehat\alpha^R,\phi_0)-L_n(\widehat\alpha,\widehat\gamma)
				\cr
				&\leq& L_n(\pi_n(\widehat\alpha +xu_n),\phi_0)-L_n(\widehat\alpha,\widehat\gamma) +  \lambda_n P_{en}( \pi_n(\widehat \alpha +xu_n) )  -  \lambda_n P_{en}(\widehat \alpha^R )   +o_P(n^{-1})\cr
				&\leq^{(g)}&L_n(\widehat\alpha +xu_n,\phi_0)-L_n(\widehat\alpha,\widehat\gamma) +o_P(n^{-1})   \cr
				&=& Q_n(\widehat\alpha +xu_n) -Q_n(\widehat\alpha) +  (\widehat\phi(\widehat\alpha+xu_n)-\phi_0)  ^2 \widehat\Sigma_2^{-1}     +o_P(n^{-1})\cr
				&=^{(h)}& x^2+2x[n^{-1/2}Z_n+\langle \widehat\alpha-\alpha_0, u_n\rangle]+  (\widehat\phi(\widehat\alpha+xu_n)-\phi_0)  ^2 \widehat\Sigma_2^{-1}     +o_P(n^{-1})\cr
				&=^{(i)}& x^2  +  (\widehat\phi(\widehat\alpha+xu_n)-\gamma_0)  ^2 \widehat\Sigma_2^{-1}     +o_P(n^{-1})\cr
				&=^{(j)}& \underbrace{x^2 +   [ x \|v_n^*\|  +g_3]^2\widehat\Sigma_2^{-1} }_{G(x)}  +o_P(n^{-1})\cr
			\end{eqnarray*}
			where (g) follows from Lemma \ref{ld.2} and that $  \lambda_n P_{en}( \pi_n(\widehat \alpha +xu_n) )  -  \lambda_n P_{en}(\widehat \alpha^R) =o_P(n^{-1})$; (h) is from  (\ref{eqa.3});   (i) is from (\ref{eqc.2}); (j) is from  Lemma \ref{led.3}.
			We choose $x=\tau^*$ to  minimize    $G(x)$, leading to the choice $\tau^*= -g_3\|v_n^*\| (\|v_n^*\| ^2+   \widehat\Sigma_2 )^{-1}    $.  It is easy to see that  $|\tau^*|=O_P(n^{-1/2})$, following this argument: from the proof of Theorem \ref{th4.2}, $g_3=O_P(\sigma n^{-1/2})$, and $\sigma^2=\Sigma_2+\|v_n^*\| ^2$. So provided that $\widehat\Sigma_2-\Sigma_2=o_P(1)\Sigma_2$,
			$$
			|\tau^*|=  O_P(n^{-1/2}) \frac{\sqrt{\Sigma_2+\|v_n^*\| ^2}   \|v_n^*\| }{ \|v_n^*\| ^2+   \widehat\Sigma_2  }  =O_P(n^{-1/2}) .
			$$
			Thus $\tau^*$ is admitted. Then we have obtained the  upper  bound: $\frac{1}{n}\widetilde S_n(\phi_0) \leq G(\tau^*)+ o_P(n^{-1})$, where
			$$
			G(\tau^*)=   \frac{g_3^2}{ \|v_n^*\| ^2+   \widehat\Sigma_2}.
			$$

			\textbf{Step 3: matching bounds.}
			
			We now show that the lower and upper bounds match, that is, $-F(x^*)=G(\tau^*)+o_P(n^{-1})$, which requires  analyzing $g_2=\phi(\widehat\alpha^R)-\phi(\alpha_0)$ and $g_6$. First, Lemma \ref{led.3} yields,   uniformly in $|x|\leq Cn^{-1/2}$,
			\begin{equation}\label{eqd.4}
			(\widehat\phi(\widehat\alpha^R+xu_n)-\gamma_0)^2\widehat\Sigma_2^{-1} - (\widehat\phi(\widehat\alpha^R)-\gamma_0)^2\widehat\Sigma_2^{-1}= H(x)+o_P(n^{-1})
			\end{equation}
			where $H(x)=\widehat\Sigma_2^{-1}x^2 \|v_n^*\| ^2 +2x \widehat\Sigma_2^{-1}\|v_n^*\|   [\widehat\phi(\alpha_0)  -\gamma_0
			+ g_2] $.
			Next, the basic inequality yields
			\begin{equation}\label{eqd.5}
			L_n(\widehat\alpha^R,\phi_0) \leq L_n(\pi_n(\widehat\alpha^R+xu_n),\gamma_0)+  o_P(n^{-1})
			\leq L_n(\widehat\alpha^R+xu_n,\gamma_0)+ o_P(n^{-1})
			\end{equation}
			where the first inequality follows from   with the assumption  that $ \lambda_n P_{en}( \widehat \alpha^R +xu_n )  -  \lambda_n P_{en}(\widehat \alpha ^R) =o_P(n^{-1})$; the  second inequality follows from Lemma \ref{ld.2}.     Uniformly for $|x|\leq Cn^{-1/2}$,
			\begin{eqnarray*}
				Q_n(\widehat\alpha^R+xu_n)-Q_n(\widehat\alpha^R)&=&o_P(n^{-1}) + x^2+2x[n^{-1/2}Z_n+\langle u_n, \widehat\alpha^R-\alpha_0\rangle]\cr
				&=&o_P(n^{-1}) + x^2+2x[n^{-1/2}Z_n+ g_2 \|v_n^*\| ^{-1}],
			\end{eqnarray*}
			where   $\langle u_n, \widehat\alpha^R-\alpha_0\rangle=\|v_n^*\| ^{-1}\frac{d\phi(\alpha_0)}{d\alpha}[\widehat\alpha^R-\alpha_0]
			+o_P(n^{-1/2})
			=\|v_n^*\| ^{-1}g_2+o_P(n^{-1/2})$.
			This along with (\ref{eqd.4}) (\ref{eqd.5}) give rise to,
			\begin{eqnarray}\label{eqdd.6}
			0&\leq& x^2+2x[n^{-1/2}Z_n+ g_2 \|v_n^*\| ^{-1} ]+H(x)+o_P(n^{-1})\cr
			&=& (1+ \|v_n^*\| ^2) x^2+2x[  n^{-1/2}Z_n+ g_2 \|v_n^*\| ^{-1}+   \|v_n^*\|   (\widehat\phi(\alpha_0)  -\gamma_0+ g_2 )
			]+o_P(n^{-1})\cr
			&=& x^2(1+\widehat\Sigma_2^{-1} \|v_n^*\| ^2 )+2x[n^{-1/2}Z_n+ g_2 \|v_n^*\| ^{-1} +\widehat\Sigma_2^{-1}\|v_n^*\|   (\widehat\phi(\alpha_0)  -\gamma_0
			+ g_2)    ]  +o_P(n^{-1})  \cr
			&=& x^2(1+\widehat\Sigma_2^{-1} \|v_n^*\| ^2 )+2x[g_6 +\widehat\Sigma_2^{-1}\|v_n^*\|   (\phi_n(\alpha_0)  -\gamma_0
			+ g_2)    ]  +o_P(n^{-1})  ,
			\end{eqnarray}
			where in the last equality, $\widehat\Sigma_2^{-1}\|v_n^*\|   (\widehat\phi(\alpha_0)  -\phi_n(\alpha_0))= o_P(n^{-1/2}) $, from Lemma \ref{ld.2}:
			\begin{eqnarray*}
				&&|  \widehat\Sigma_2^{-1}\|v_n^*\|   (\widehat\phi(\alpha_0)  -\phi_n(\alpha_0))|
				\leq |  \widehat\Sigma_2^{-1}\|v_n^*\|     \frac{1}{n}\sum_{t=1}^n  ( \Gamma(X_t)-\widehat\Gamma_t) \rho(Y_{t+1},\alpha_0)| =o_P(n^{-1/2}).
			\end{eqnarray*}
			Hence (\ref{eqdd.6}) implies there is some $\bar\eta_n=o_P(n^{-1})$ so that
			\begin{equation}\label{eqd.77}
			x^2(1+\widehat\Sigma_2^{-1} \|v_n^*\| ^2 )+2x[g_6 +\widehat\Sigma_2^{-1}\|v_n^*\|   (\phi_n(\alpha_0)  -\gamma_0
			+ g_2)    ]  + \bar\eta_n\geq 0.
			\end{equation}
			
			We now derive some important intermediate results from (\ref{eqd.77}). First, let $$C_n:=\min\{ \|v_n^*\| \Sigma_2^{-1/2}  ,\|v_n^*\|^2_*\Sigma_2^{-1},   \|v_n^*\| \sigma \Sigma_2^{-1} \} .$$
			It is known that $(C_n+C_n^2)/(1+\widehat\Sigma_2^{-1} \|v_n^*\| ^2 )\leq 2$ because $\|v_n^*\|\leq \sigma.$
			So $\bar\eta_n=o_P(n^{-1}) C_n^2/(1+\widehat\Sigma_2^{-1} \|v_n^*\| ^2 )$, implying
			$\bar\eta_n n^{1/2}C_n^{-1}\ll n^{-1/2} C_n/(1+\widehat\Sigma_2^{-1} \|v_n^*\| ^2 )$. Hence there is a positive sequence $\epsilon_n=O_P(n^{-1/2})$ so that
			$\bar\eta_n n^{1/2}C_n^{-1}\ll \epsilon_n\ll n^{-1/2} C_n/(1+\widehat\Sigma_2^{-1} \|v_n^*\| ^2 )$. Hence $\bar\eta_n/\epsilon_n+\epsilon_n(1+\widehat\Sigma_2^{-1} \|v_n^*\| ^2 )=o_P(n^{-1/2} C_n)$.
			Take $x=\pm\epsilon_n$ and divide  by $2\epsilon_n$ on (\ref{eqd.77}). We reach four intermediate results:
			\begin{eqnarray}
			&& g_6+ \widehat\Sigma_2^{-1}  \|v_n^*\|   ( \phi_n(\alpha_0)  -\gamma_0+ g_2 ) 	=  O_P(\bar\eta_n/\epsilon_n+\epsilon_n(1+\widehat\Sigma_2^{-1} \|v_n^*\| ^2 ))   =o_P(n^{-1/2} C_n)\label{eqd.a7}\\
			&& (1+\widehat\Sigma_2^{-1}  \|v_n^*\|  ^2) g_6+ \widehat\Sigma_2^{-1}  \|v_n^*\|   g_5=o_P(n^{-1/2} C_n)\label{eqd.8}\\
			&& g_4=(o_P(n^{-1/2} C_n)  - g_6)^2\widehat\Sigma_2\|v_n^*\| ^{-2}\label{eqd.9}\\
			&&g_6= O_P(n^{-1/2}) \label{eqd.10}
			\end{eqnarray}
			where (\ref{eqd.a7}) follows from (\ref{eqd.77}) with $x=\pm\epsilon_n$; the left hand sides of (\ref{eqd.a7}) and (\ref{eqd.8}) are equal; (\ref{eqd.9}) is from (\ref{eqd.a7}) and the definition of $g_4$; (\ref{eqd.10}) is  from (\ref{eqd.8}),   $g_5= O_P(\sigma n^{-1/2})$ and that $o_P(n^{-1/2} C_n)= \sigma^2\Sigma_2^{-1}O_P(n^{-1/2})$.
			Also, the proof of (\ref{eqd.10}) does not rely on the conclusion of Step 1, so it
			verifies that $|x^*|=O_P(n^{-1/2})$, a claim used in step 1.

			We are now ready to match the bounds. From   (\ref{eqd.9}) and (\ref{eqd.10}),
			\begin{eqnarray*}
				- F(x^*)&=&g_6^2+g_4
				= g_6^2
				+ ( o_P(n^{-1/2} C_n)  - g_6)^2\widehat\Sigma_2\|v_n^*\| ^{-2}
				=g_6^2 (1+\widehat\Sigma_2\|v_n^*\| ^{-2})+o_P(n^{-1})    \cr
				&=^{(k)}&\frac{[o_P(n^{-1/2} C_n) - \widehat\Sigma_2^{-1}  \|v_n^*\|  g_5	  ]^2}{   (1+\widehat\Sigma_2^{-1}  \|v_n^*\|  ^2)^2 }(1+\widehat\Sigma_2\|v_n^*\| ^{-2})+o_P(n^{-1})\cr
				&=&\frac{   g_5	  ^2}{   \widehat\Sigma_2+   \|v_n^*\|  ^2 } +o_P(n^{-1})=^{(l)}\frac{   g_3	  ^2}{   \widehat\Sigma_2+   \|v_n^*\|  ^2 } +o_P(n^{-1}) =G(\tau^*)+o_P(n^{-1}) ,
			\end{eqnarray*}
			where  (k) is from (\ref{eqd.8}); (l) is from the fact that (due to (\ref{eqc.2}))
			\begin{eqnarray*}|g_3^2-g_5^2|&\leq& |  \phi(\widehat\alpha)- \phi(\alpha_0) +\|v_n^*\|    n^{-1/2}Z_n| O_P(n^{-1/2}\sigma)
				\cr
				&\leq&  \left|  \frac{d\phi(\alpha_0)}{d\alpha}[\widehat\alpha- \alpha_0] +\|v_n^*\|    n^{-1/2}Z_n\right| O_P(n^{-1/2}\sigma)+o_P(\sigma^2n^{-1})\cr
				&=& \left| \langle\widehat\alpha- \alpha_0, u_n  \rangle   +   n^{-1/2}Z_n+ o_P  (n^{-1/2})\right| O_P(\|v_n^*\| n^{-1/2}\sigma)+o_P(\sigma^2n^{-1})=o_P(\sigma^2n^{-1}).
			\end{eqnarray*}
			
			Thus we have proved that the upper and lower bounds match up to $o_P(n^{-1}) $, implying
			$$
			\widetilde S_n(\phi_0) = nG(\tau^*)+ o_P(1) = \frac{ng_3^2}{ \widehat\sigma^2}+o_P(1)\to^d\chi^2_1
			$$
			where the convergence in distribution follows from (\ref{eqc.d3}).

		\end{proof}
	}
	
	\begin{lem}[for Theorem \protect\ref{thqlr:eff}]
		{\small \label{ld.2} Suppose $( \delta_n^{\eta}
			\sup_x|\widehat\Sigma(x)-\Sigma(x)| +\sqrt{k_n} d_n\delta_n^{\eta}
			+\varphi_n^2)= o_P(n^{-1/2}\min\{1,\sigma^{-1}\}).$ In addition, suppose $%
			(1+\|v_n^*\|) \sup_{\alpha\in\mathcal{C}_n}|\phi(\pi_n\alpha)-\phi(\alpha)|
			=o (n^{-1/2})$. }
		
		{\small
		}
		
		{\small
		}
		
		{\small
		}
		
		{\small
		}
		
		{\small
		}
		
		{\small Write $\widehat\phi(\alpha):= \frac{1}{n}\sum_{t=1}^n [l( h(W_t)) -
			\widehat\Gamma_t \rho(Y_{t+1}, \alpha)]$. Then }
		
		{\small
		}
		
		{\small (i) $\|\widehat\alpha^R-\alpha\|_{\infty,\omega}=O_P( \delta_n),
			Q(\widehat\alpha^R)\leq O_P(\bar\delta_n^2)$. }
		
		{\small (ii) $\sup_{\alpha_1,\alpha_2\in\mathcal{C}_n}[
			\widehat\phi(\alpha_1) - \widehat\phi(\alpha_2) ] -
			[\phi(\alpha_1)-\phi(\alpha_2)]= O_P ( \delta_n^{\eta}
			\sup_x|\widehat\Sigma(x)-\Sigma(x)| +\sqrt{k_n} d_n\delta_n^{\eta}
			+\varphi_n^2)$. }
		
		{\small (iii) $|L_n(\pi_n(\widehat\alpha +xu_n),\gamma_0)-L_n(\widehat\alpha
			+xu_n,\gamma_0)|=o_P(n^{-1})$ }
		
		{\small (iv) $|L_n(\pi_n(\widehat\alpha^R
			+xu_n),\gamma_0)-L_n(\widehat\alpha^R +xu_n,\gamma_0)|=o_P(n^{-1})$. }
	\end{lem}
	
	{\small
		\begin{proof}   
			

			(i)
			The inequality $ L_n(\widehat\alpha^R,\phi_0)+\lambda_n P_{en}(\widehat h^R)\leq   L_n(\pi_n  \alpha_0,\gamma_0)+\lambda_n P_{en}(\pi_n  h_0)$
			implies
			$$Q_n(\widehat\alpha^R)\leq   Q_n(\pi_n\alpha_0)+F_n(\pi_n\alpha_0) +O_P(\lambda)
			=O(\bar\delta_n^2)+ F_n(\pi_n\alpha_0)
			$$ where
			$F_n(\alpha):= (\widehat\phi(\alpha)-\gamma_0)'\widehat\Sigma_2  ^{-1}(\widehat\phi(\alpha)-\gamma_0)  $. We now bound $F_n(\pi_n\alpha_0)$.
			Note that  \begin{eqnarray*}
				\widehat\phi(\pi_n\alpha_0)-\gamma_0
				&=& \frac{1}{n}\sum_{t=1}^n  l(  \pi_nh_0(W_t))-\mathbb E l( h_0(W_t))
				- \frac{1}{n}\sum_{t=1}^n  ( \widehat\Gamma_t -\Gamma(X_t)) \rho(Y_{t+1},  \pi_n\alpha_0)\cr
				&&-\frac{1}{n}\sum_{t=1}^n  [ \Gamma(X_t) \rho(Y_{t+1},  \pi_n\alpha_0)  -\mathbb  E  \Gamma_t  \rho(Y_{t+1},   \alpha_0)]-\mathbb  E  \Gamma_t  [m(X_t,   \alpha_0)- m(X_t,  \pi_n \alpha_0) ] .
			\end{eqnarray*}
			The first term is bounded by $O_P(n^{-1/2})+\mathbb E[ l( \pi_nh_0(W_t))-l( h_0(W_t))]  $;      the second term is bounded by $O_P(\bar\delta_n)$, following from the same argument as those for (\ref{eqc.3afda});    the third and fourth terms are  bounded by $O_P(n^{-1/2})+ \mathbb E \Gamma(X_t)[ m(X_t,  \pi_n\alpha_0)- m(X_t,  \alpha_0)]\leq  O_P( \sqrt{Q(\pi_n\alpha_0)})$. Hence $   \widehat\phi(\pi_n\alpha_0)-\gamma_0=O_P(\bar \delta_n)$. This implies $F_n(\pi_n\alpha_0)=O_P(\bar\delta_n^2)$.
			This yields  $Q_n( \widehat \alpha^R)=O_P(\bar\delta_n^2)$. Then from the proof of Theorem \ref{th3.1},
			$$
			Q(\widehat\alpha^R)\leq C \mathbb E \widetilde m(X_t,\widehat\alpha^R)^2+ O_P(\bar\delta_n^2)
			\leq C  \frac{1}{n} \sum_t\widetilde  m(X_t,\widehat\alpha^R)^2+ O_P(\bar\delta_n^2)
			\leq Q_n(\widehat \alpha^R)+ O_P(\bar\delta_n^2)= O_P(\bar\delta_n^2).
			$$
			It also implies $\|\widehat\alpha^R-\pi_n\alpha_0\|\leq \|\widehat\alpha^R- \alpha_0\|+ \|  \pi_n\alpha_0-\alpha_0\|=O_P(\bar\delta_n)$, and hence
			$\|\widehat\alpha^R-\alpha_0\|_{\infty,\omega}\leq  \|  \pi_n\alpha_0-\alpha_0\|_{\infty,\omega}+\|\widehat\alpha^R-\pi_n\alpha_0\|_{\infty,\omega}
			=  \|  \pi_n\alpha_0-\alpha_0\|_{\infty,\omega}+ O_P(\omega_n(\bar\delta_n))=O_P(\delta_n).
			$

			(ii) Let $a_1=[(\phi_n(\alpha_1)-\phi(\alpha_1))  -(\phi_n(\alpha_2) -\phi(\alpha_2))]$. We have
			\begin{eqnarray*}
				&& [ \widehat\phi(\alpha_1) - \widehat\phi(\alpha_2) ]  - [\phi(\alpha_1)-\phi(\alpha_2)]=a_1
				+\frac{1}{n}\sum_{t=1}^n  (\Gamma(X_t)-\widehat\Gamma_t)  \rho(Y_{t+1},  \alpha_1) + \frac{1}{n}\sum_{t=1}^n   ( \widehat\Gamma_t -\Gamma(X_t)) \rho(Y_{t+1},  \alpha_2)\cr
				&\leq& O_P ( \delta_n^{\eta} \sup_x|\widehat\Sigma(x)-\Sigma(x)|  +\sqrt{k_n} d_n\delta_n^{\eta}
				+\varphi_n^2),
			\end{eqnarray*}
			where the first inequality follows from  bounds for (\ref{eqc.3afda}) and Lemma \ref{lf.1}.

			(iii) Let $\alpha=\widehat\alpha +xu_n$. By the same proof of Lemma \ref{le.1}(ii), $Q_n(\pi_n\alpha)-Q_n(\alpha)=o_P(n^{-1})$.
			Next,
			\begin{eqnarray*}
				\widehat\phi(\alpha)-\gamma_0&=& a_1 (\alpha)+a_3+a_4(\alpha)-\frac{1}{n}\sum_t(\widehat\Gamma_t-\Gamma(X_t))\rho(Y_{t+1},\alpha),\cr
				a_1(\alpha)&:=&\phi_n(\alpha) -\phi(\alpha)  -[ \phi_n(\alpha_0) -\phi( \alpha_0)   ] \cr
				a_3&:=&  \phi_n(\alpha_0) -\phi( \alpha_0) \cr
				a_4(\alpha)&:=&\phi(\alpha)-\phi( \alpha_0).
			\end{eqnarray*}
			By Lemma \ref{lf.1} and the same proof for bounding  (\ref{eqc.3afda}), $$a_1-\frac{1}{n}\sum_t(\widehat\Gamma_t-\Gamma(X_t))\rho(Y_{t+1},\alpha)=O_P ( \delta_n^{\eta} \sup_x|\widehat\Sigma(x)-\Sigma(x)|  +\sqrt{k_n} d_n\delta_n^{\eta}
			+\varphi_n^2)= o_P(n^{-1/2}\min\{1,\sigma^{-1}\}),$$
			where the last equality follows from the assumption $( \delta_n^{\eta} \sup_x|\widehat\Sigma(x)-\Sigma(x)|  +\sqrt{k_n} d_n\delta_n^{\eta}
			+\varphi_n^2)= o_P(n^{-1/2}\min\{1,\sigma^{-1}\}).$   The same bound holds when $\alpha$ is replaced with $\pi_n\alpha.$ Meanwhile, by the proof of Theorem \ref{th4.2}, $a_3=O_P(\sigma n^{-1/2})$.

			To bound $a_4(\alpha)$, first note that
			\begin{eqnarray*}
				&&\|\pi_n(\widehat \alpha+xu_n)-\alpha_0\|^2\leq C\mathbb E m(X_t,  \pi_n(\widehat \alpha+xu_n))^2
				\leq   C\mathbb E[ m(X_t,  \pi_n(\widehat \alpha+xu_n))-m(X_t,   \widehat \alpha+xu_n) ]^2 \cr
				&&+C \mathbb E[  m(X_t,   \widehat \alpha+xu_n) -m(X_t,   \widehat \alpha ) ]^2
				+ C Q(\widehat\alpha) \leq     O_P(\mu_n^2+ \bar\delta_n^2) \cr
				&&\|(\widehat \alpha+xu_n)-\alpha_0\|^2\leq O_P( \bar\delta_n^2).
			\end{eqnarray*}
			So    by Assumption \ref{asssmoothphi}, and that $\langle v_n^*, \alpha_0-\alpha_{0,n}\rangle=0$,
			\begin{eqnarray*}
				a_4(\widehat\alpha +xu_n)&\leq&| \phi(\widehat\alpha+xu_n ) -\phi(\alpha_0)|
				\leq   \left|\frac{d\phi(\alpha_0)}{d\alpha}[\widehat \alpha+xu_n-\alpha_0] \right|\cr
				&\leq&   \left|\frac{d\phi(\alpha_0)}{d\alpha}[\widehat \alpha+xu_n-\alpha_{0,n}] \right|
				+\left|\frac{d\phi(\alpha_0)}{d\alpha}[ \alpha_{0,n}-\alpha_0] \right|
				\cr
				&=& o_P(\|v_n^*\|) n^{-1/2}+ |\langle \widehat \alpha -\alpha_0, v_n^*\rangle
				+ x\langle   u_n , v_n^*\rangle| =  O_P(\|v_n^*\|n^{-1/2}).\cr
				a_4(\pi_n(\widehat\alpha +xu_n))&\leq&| \phi(\pi_n(\widehat\alpha+xu_n) ) -\phi(\alpha_0)|\leq   \left|\frac{d\phi(\alpha_0)}{d\alpha}[\pi_n(\widehat \alpha+xu_n)-\alpha_0] \right|\cr
				&\leq &    \left|\frac{d\phi(\alpha_0)}{d\alpha}[\pi_n(\widehat \alpha+xu_n)-\alpha_{0,n}]\right|
				+  \left|\frac{d\phi(\alpha_0)}{d\alpha}[ \alpha_{0,n}-\alpha_0]\right|
				\cr
				&=&  o_P(\|v_n^*\| ) n^{-1/2}+ |\langle \pi_n(\widehat \alpha+xu_n)-\alpha_0, v_n^*\rangle |
				\cr
				&\leq& o_P(\|v_n^*\| ) n^{-1/2}+ \|\pi_n(\widehat \alpha+xu_n)-\alpha_0\| \|v_n^*\| \cr
				&\leq&  O_P(   \bar\delta_n )\|v_n^*\| .
			\end{eqnarray*}
			Also, by the proof of Lemma \ref{le.1}(ii), $
			Q_n(\pi_n (\widehat\alpha +xu_n))-Q_n(\widehat\alpha +xu_n)  =o_P(n^{-1})  $.
			Together,  with  $\alpha=\widehat\alpha +xu_n$,  $|\widehat\phi(\alpha)-\gamma_0| \leq o_P(n^{-1/2}\min\{1,\sigma^{-1}\})+  O_P(\|v_n^*\|n^{-1/2}) $, and $c_n:= |\phi(\pi_n\alpha)-\phi(\alpha)|$,
			\begin{eqnarray*}
				&&L_n(\pi_n(\widehat\alpha +xu_n),\gamma_0)-L_n(\widehat\alpha +xu_n,\gamma_0)
				=Q_n(\pi_n\alpha)-Q_n(\alpha)
				\cr
				&&+
				(\widehat\phi(\pi_n\alpha)-\gamma_0)'\widehat\Sigma_2^{-1}(\widehat\phi(\pi_n\alpha)-\gamma_0)
				-(\widehat\phi(\alpha)-\gamma_0)'\widehat\Sigma_2^{-1}(\widehat\phi(\alpha)-\gamma_0)\cr
				&=&o_P(n^{-1}) + (\widehat\phi(\pi_n\alpha)-\widehat\phi(\alpha))'\widehat\Sigma_2^{-1}(\widehat\phi(\pi_n\alpha)-\widehat\phi(\alpha))
				+2(\widehat\phi(\pi_n\alpha)-\widehat\phi(\alpha))'\widehat\Sigma_2^{-1}(\widehat\phi(\alpha)-\gamma_0)\cr
				&=&o_P(n^{-1}) + O_P(1) |\phi(\pi_n\alpha)-\phi(\alpha)|^2 + O_P(1) |\phi(\pi_n\alpha)-\phi(\alpha)|   |\widehat\phi(\alpha)-\gamma_0| \cr
				&\leq &o_P(n^{-1}) + O_P(c_n^2)   + o_P(n^{-1/2}\min\{1,\sigma^{-1}\})c_n+  O_P(\|v_n^*\|n^{-1/2}c_n) =o_P(n^{-1}).
			\end{eqnarray*}

			(iv) The proof is the same for part (iii).
			

		\end{proof}
	}
	
	\begin{lem}
		{\small \label{led.3} Write $\nu_n:=\delta_n^{\eta}
			\sup_x|\widehat\Sigma(x)-\Sigma(x)| +\sqrt{k_n} d_n\delta_n^{\eta}
			+\varphi_n^2$. Suppose }
		
		{\small $\nu_n = o_P(n^{-1/2} \sigma^{-1} )$,  $(p_n+\nu_n)\bar\delta_n\|v_n^*%
			\|=o(n^{-1})$. }

		{\small Then Uniformly for $|x|\leq Cn^{-1/2}$, for $b(x):=\widehat\phi(%
			\alpha_0) -\gamma_0 + \phi(\widehat\alpha^R)- \phi(\alpha_0)$, }
		
		{\small (i) $(\widehat\phi(\widehat\alpha+xu_n)-\gamma_0) ^2
			\widehat\Sigma_2^{-1} = [ x \|v_n^*\| +g_3]^2\widehat\Sigma_2^{-1}
			+o_P(n^{-1})$ }
		
		{\small (ii) $(\widehat\phi(\widehat\alpha^R+xu_n)-\gamma_0)^2\widehat%
			\Sigma_2^{-1} -
			(\widehat\phi(\widehat\alpha^R)-\gamma_0)^2\widehat\Sigma_2^{-1}= x^2
			\|v_n^*\| ^2 \widehat\Sigma_2^{-1} +2x \|v_n^*\| b(x)\widehat\Sigma_2^{-1}
			+o_P(n^{-1}) $. }
		
		{\small (iii) $(\widehat\phi(\widehat \alpha^R)-\gamma_0)^2
			\widehat\Sigma_2^{-1} = [ \phi_n(\alpha_0) -\gamma_0 +
			\phi(\widehat\alpha^R)- \phi(\alpha_0)]^2\widehat\Sigma_2^{-1} +o_P(n^{-1})$%
			.
		}
		
		{\small
		}
	\end{lem}
	
	{\small
		\begin{proof}
			
			(i)  Let $g_3=[  \phi_n(\alpha_0) -\phi( \alpha_0)]+  [   \phi(\widehat\alpha)-\phi( \alpha_0) ]$.
			\begin{eqnarray*}
				\widehat\phi(\widehat\alpha+xu_n)-\gamma_0&=& \phi(\widehat\alpha+xu_n)-\phi(\widehat\alpha)+g_3+b_1+b_2\cr
				b_1&=&[\widehat\phi(\widehat\alpha+xu_n)-\widehat\phi(\widehat\alpha)]-[\phi(\widehat\alpha+xu_n)-\phi(\widehat\alpha)]  \cr
				b_2&=&\widehat\gamma-\gamma_0-g_3.
			\end{eqnarray*}
			We now work with $\phi(\widehat\alpha+xu_n)-\phi(\widehat\alpha)$. By Assumption \ref{asssmoothphi} and the Riesz representation,
			\begin{eqnarray}\label{eqd.7a}
			\phi(\widehat\alpha+xu_n)-\phi(\widehat\alpha)&=&      \phi(\widehat\alpha+xu_n)-  \phi(\alpha_0)- [\phi(\widehat\alpha)-\phi(\alpha_0)] \cr
			&=&\frac{d\phi(\alpha_0)}{d\alpha}[\widehat\alpha-\alpha_0+xu_n] - \frac{d\phi(\alpha_0)}{d\alpha}[\widehat\alpha-\alpha_0]\cr
			&=&\langle v_n^*, \widehat\alpha-\alpha_0+xu_n\rangle-\langle v_n^*, \widehat\alpha-\alpha_0\rangle+o_P(\|v_n^*\|) n^{-1/2}\cr
			&=&\langle v_n^*, u_n\rangle x 
			= x \|v_n^*\|   .
			\end{eqnarray}
		
			Hence $  \widehat\phi(\widehat\alpha+xu_n)-\gamma_0= x \|v_n^*\|  +g_3  +b_1+b_2$. Also note that $g_3=O_P(\sigma n^{-1/2})$. Together $  (\widehat\phi(\widehat\alpha+xu_n)-\gamma_0)  ^2 \widehat\Sigma_2^{-1}   $ is bounded by
			$$
			[ x \|v_n^*\|  +g_3]^2\widehat\Sigma_2^{-1} +o_P(n^{-1})+O_P(b_1^2+b_2^2)+O_P(b_1+b_2)(\|v_n^*\| +\sigma)n^{-1/2}.
			$$
			By   Lemma \ref{ld.2}, $b_1=O_P(\nu_n) $. By   the proof of Theorem \ref{th4.2}, $b_2=O_P(\nu_n)= o_P(n^{-1/2} \sigma^{-1} )$. So the above is bounded by $(\sigma\geq \|v_n^*\|)$,
			$$ [ x \|v_n^*\|  +g_3]^2\widehat\Sigma_2^{-1} +o_P(n^{-1})+O_P(\nu_nn^{-1/2} )(\|v_n^*\| +\sigma) = [ x \|v_n^*\|  +g_3]^2\widehat\Sigma_2^{-1}+o_P(n^{-1}).$$

			(ii) $ (\widehat\phi(\widehat\alpha^R+xu_n)-\gamma_0)^2\widehat\Sigma_2^{-1} - (\widehat\phi(\widehat\alpha^R)-\gamma_0)^2\widehat\Sigma_2^{-1} $ equals
			$$
			\Delta_1:= (\widehat\phi(\widehat\alpha^R+xu_n)- \widehat\phi(\widehat\alpha^R))^2\widehat\Sigma_2^{-1}
			+2 (\widehat\phi(\widehat\alpha^R+xu_n)- \widehat\phi(\widehat\alpha^R))\widehat\Sigma_2^{-1} (\widehat\phi(\widehat\alpha^R)-\gamma_0).
			$$
			The same argument as in (\ref{eqd.7a}) yields $\widehat\phi(\widehat\alpha^R+xu_n)- \widehat\phi(\widehat\alpha^R)=x \|v_n^*\|  $.
			Meanwhile,
			\begin{eqnarray}
			\widehat\phi(\widehat\alpha^R)-\gamma_0&=&\widehat\phi(\alpha_0)  -\gamma_0
			+ \phi(\widehat\alpha^R)- \phi(\alpha_0)+ \underbrace{  [\widehat\phi(\widehat\alpha^R)-\widehat\phi(\alpha_0)]-[\phi(\widehat\alpha^R)- \phi(\alpha_0)]}_{=O_P(\nu_n) \text{ by Lemma \ref{ld.2}}}. \label{eqd.134}\cr\\
			\widehat\phi(\widehat\alpha^R+xu_n)- \widehat\phi(\widehat\alpha^R)&=&O_P(\mu_n) +  \phi(\widehat\alpha^R+xu_n)- \phi(\widehat\alpha^R)
			=\|v_n^*\|  x+O_P(\nu_n),\cr
			\phi(\widehat \alpha^R)- \phi(\alpha_0)&=&
			\frac{d\phi(\alpha_0)}{d\alpha}[\widehat\alpha^R-\alpha_0]  =\langle \widehat\alpha^R-\alpha_0,v_n^*\rangle    \cr
			&\leq& \|\widehat\alpha^R-\alpha_0\| \|v_n^*\|  
			\leq C\sqrt{Q(\widehat\alpha^R)}\|v_n^*\|   \cr
			&\leq& O_P(\bar\delta_n\|v_n^*\|)   .
			\end{eqnarray}
			
			Hence  with $\widehat\phi(\alpha_0)-\gamma_0=O_P(n^{-1/2}\sigma)$, and $ \varpi _n\sigma= O(1)$
			\begin{eqnarray*}
				\Delta_1&=& x^2 \|v_n^*\| ^2\widehat\Sigma_2^{-1} +o_P(n^{-1})+2[x \|v_n^*\|  ]   [\widehat\phi(\alpha_0)  -\gamma_0
				+ \phi(\widehat\alpha^R)- \phi(\alpha_0)+O_P(\mu_n)] \widehat\Sigma_2^{-1}\cr
				&=& x^2 \|v_n^*\| ^2 \widehat\Sigma_2^{-1}+2x \|v_n^*\|   [\widehat\phi(\alpha_0)  -\gamma_0
				+ \phi(\widehat\alpha^R)- \phi(\alpha_0)]\widehat\Sigma_2^{-1}+o_P(n^{-1})
				\cr
				&&
				+ O_P (\mu_n)n^{-1/2} \|v_n^*\| \cr
				&=& x^2 \|v_n^*\| ^2 \widehat\Sigma_2^{-1}+2x \|v_n^*\|   [\widehat\phi(\alpha_0)  -\gamma_0
				+ \phi(\widehat\alpha^R)- \phi(\alpha_0)]\widehat\Sigma_2^{-1}+ o_P(n^{-1}) ,
			\end{eqnarray*}

			(iii)
			Define
			\begin{eqnarray*}
				z_1&:=&(\widehat\phi(\widehat \alpha^R)-\gamma_0)^2 \widehat\Sigma_2^{-1}\cr
				z_2&:=&[\widehat\phi(\alpha_0)  -\gamma_0
				+ \phi(\widehat\alpha^R)- \phi(\alpha_0)]^2\widehat\Sigma_2^{-1} \cr
				z_3&:=&[ \phi_n(\alpha_0)  -\gamma_0
				+ \phi(\widehat\alpha^R)- \phi(\alpha_0)]^2\widehat\Sigma_2^{-1}.
			\end{eqnarray*}

			First the proof for bounding (\ref{eqc.3afda})  can be simplified to yield
			\begin{eqnarray*}
				&&    |\widehat\phi(\alpha_0)- \phi_n(\alpha_0)|\leq \frac{1}{n}\sum_{t=1}^n  ( \widehat\Gamma_t -\Gamma(X_t)) \rho(Y_{t+1},   \alpha_0)\cr
				&=& \frac{	1}{n}   \bar b_n( \alpha_0)'  (\widehat\Sigma_n^{-1}-\Sigma_n^{-1})\rho_n( \alpha_0) + O_P(\delta_n^\eta) [\sup_x \|\widehat\Sigma(x)-\Sigma(x)\| + \sqrt{\frac{k_n}{n}}]  .\cr
				&& \sqrt{z_2}+ \sqrt{z_3} = O_P(n^{-1/2}\sigma+\bar\delta_n\|v_n^*\| )     .
			\end{eqnarray*}

			By (\ref{eqd.134}),  and with the assumption $(\nu_n+p_n)\bar\delta_n\|v_n^*\| =o_P(n^{-1}),$ \begin{eqnarray*}
				|z_1-z_2|&\leq&	O_P(\nu_n^2)+ O_P( \mu_n)  |\widehat\phi(\alpha_0)  -\gamma_0
				+ \phi(\widehat\alpha^R)- \phi(\alpha_0)|
				\cr
				&=&o_P(n^{-1}) + O_P( \nu_n) (n^{-1/2}\sigma+   \bar\delta_n\|v_n^*\|  )
				=o_P(n^{-1}).\cr
				|z_2-z_3|&\leq& O_P(1)   |\widehat\phi(\alpha_0)- \phi_n(\alpha_0)|( \sqrt{z_2}+ \sqrt{z_3} ) = o_P(n^{-1})+ O_P( p_n    \bar\delta_n\|v_n^*\| )=o_P(n^{-1}).
			\end{eqnarray*}
			Hence $z_1-z_3=o_P(n^{-1}). $

		\end{proof}
	}
	
	\section{Verifying conditions for RL, NPIV and NPQIV in Section \protect\ref{sec:examples}}

	\subsection{{\protect\small Reinforcement learning model: proof of Proposition \protect\ref%
			{ass6.1rl}}}
	
	{\small
	\begin{proof}
Recall that $Q^{\pi}$ denotes the true $Q$-function.  	Let $X_t= (S_t, A_t)$. We have $$m(X_t, h)=\mathbb E(R_t|X_t)- h(S_t,A_t) + \gamma \mathbb E\left[\int_{x \in \mathcal A} \pi(x|S_{t+1}) h(S_{t+1}, x)\bigg{|}S_t, A_t \right]\mathrm{d} x$$
	In addition, for $	\frac{dm}{dh}[v] $ defined in  (\ref{eq6.2mdq}),
	 $
	\|v\|^2:= \mathbb{E}\left( 	\frac{dm}{dh}[v] \right)^2
	\Sigma(X_t)^{-1} .
 $ 

	   \textit{Verifying Assumption \ref{ass2.1}. } 
	 The Bellman equation implies $m(X_t, Q^{\pi})=0$ so for all $h\in\mathcal H_n$,  $m(X_t, h)= m(X_t, h)- m(X_t, Q^{\pi})$. Hence
	  $m(X_t, h)= 	\frac{dm}{dh}[h-Q^{\pi}] .$ 
	   $$
	   \|h- Q^{\pi}\|^2= \mathbb{E}\left( 	\frac{dm}{dh}[h-Q^{\pi}] \right)^2
	\Sigma(X_t)^{-1} = \mathbb E m(X_t, h)^2\Sigma(X_t)^{-1}.
	   $$
	   This shows condition (i). 
	   For condition (ii), it is also easy to see:
	   $$
	   \mathbb{E }m(X_t,\pi_n\alpha_0)^2 \Sigma(X_t)^{-1} = \mathbb{E}\left( 	\frac{dm}{dh}[\pi_n Q^{\pi}-Q^{\pi}] \right)^2
	\Sigma(X_t)^{-1}  = \|\pi_n Q^{\pi}-Q^{\pi}\|^2.
	   $$

	    \textit{Verifying Assumption \ref{ass3.1}. }  
	    For condition (i), let $T_t =\Psi_j(X_t)^2+1$. 
	    Also,
	$$
	\rho(Y_{t+1}, h)=R_t- h(S_t,A_t) + \gamma K(h),\quad K(h)=\int_{x \in \mathcal A} \pi(x|S_{t+1}) h(S_{t+1}, x) \mathrm{d} x.
	$$
	and
	$
	\epsilon(S_t, h_1)- 	\epsilon(S_t, h_2)
	=\gamma K(h_1)-\gamma K(h_2)-\gamma\mathbb E[K(h_1)-K(h_2)|S_t, A_t].
	$
	Now 
	$$
	|K(h_1)- K(h_2)| 
	\leq \|h_1-h_2\|_{\infty,\omega}  M(S_{t+1}),\quad M(S_{t+1}):=\int\pi(x|S_{t+1}) (1+x^2+S_{t+1}^2)^{\omega/2}dx.
	$$
	    Uniform in $j$, with $\mathbb E\max_{j\leq k_n} \Psi_j(X_t)^4<\infty$, and $\mathbb E M(S_{t+1})^4<\infty$,
	    $$
	    \mathbb  E T_t\sup_{ \|h_1-h
			\|_{\infty,\omega}< \delta} | \epsilon( S_t, h_1)- \epsilon( S_t, h )|
		^2
		\leq 4\gamma^2\sup_{ \|h_1-h
			\|_{\infty,\omega}< \delta} \|h_1-h\|_{\infty,\omega}^2C\leq C\delta^2.
	    $$
	    For (ii), Let $T_1:=\max_{j\leq k_n}\Psi_j(X_t)^2$. 
	    Also  $\mathbb E R_t^4<C$, $\mathbb E T_1 R_t^2<C$.   Since $\sup_{h\in\mathcal H_n}\|h\|_{\infty,\omega}^2 <C$,  $\mathbb{E} T_1\sup_{h\in\mathcal{H}_n}
	h(S_t, A_t)^2\leq  \mathbb E   T_1   (1+|S_t|^2+|A_t|^2)^{\omega}  \|h\|_{\infty,\omega}^2 <C.$ Also
	$
	\mathbb E T_1\sup_hK(h)^2
	\leq \mathbb E M(S_{t+1})^2\sup_h\|h\|_{\infty,\omega}^2
	$. Hence $\mathbb{E} T_1\sup_{h\in\mathcal{H}_n}
		\rho(Y_{t+1}, h )^2\leq C.$

	    For (iii), the pathwise derivative of $m$ is given by (\ref{eq6.2mdq}), for $C:= \left[\mathbb E(1+|S_t|^2+|A_t|^2)^{\omega}+\mathbb E M(S_{t+1})^2\right]$,
	    $$
	    \|h- Q^{\pi}\|^2\leq C\mathbb E[h-Q^\pi]^2
	    +C\mathbb E |\mathbb E(K(h)-K(Q^\pi)|S_t, A_t)|^2\leq C\|h-Q^\pi\|_{\infty,\omega}
	    $$

	      \textit{Verifying Assumption \ref{ass4.2}.} We note that for any $h\in\mathcal H_n\cup\{Q^{\pi}\}$, $$ \frac{d
			m(X_t,h)}{dh}[u_n]=\gamma \int_{x \in \mathcal A} \mathbb E\left[\pi(x|S_{t+1}) u_n(S_{t+1}, x) | S_t, A_t\right]\mathrm{d} x
	- u_n(S_t, A_t),$$
	which does not depend on $h$. Also,  for any $h, \tau, v,$ because of the linearity, 
$	\frac{d^2}{d\tau^2} m(X_t, h+ \tau v)=0$.
\\
For condition (i), let $r:=2+\zeta$, $a:= | \rho(Y_{t+1}, Q^{\pi})|^{2+\zeta}$, $b:=\left| \frac{d
			m(X_t, Q^{\pi})}{d h}[u_n] \right | $ then we have \\ $b\leq |\gamma \mathbb E_t\mathbb E^{\pi} u_n(S_{t+1},A)|+ |u_n(S_t, A_t)|$ where $\mathbb E_t=\mathbb E(.|S_t, A_t)$ and $\mathbb E^{\pi}$ is with respect to the distribution $\pi(.|S_{t+1})$ for $A$.  
			Let $d_{\pi}:=\mathbb E   \mathbb E^{\pi} |u_n(S_{t+1},A)|^{2r}$ and $d:=\mathbb E|u_n(S_t, A_t)|^{2r}$.
			Then 
			$$
			\mathbb E a^2 \leq C+ \mathbb E |R_t|^{4+2\zeta}<C.
			$$
			Also, $d+d_\pi\leq C$ because $\mathbb E   \mathbb E^{\pi} |u_n(S_{t+1},A)|^{2r}+ \mathbb E |v_n^*(S_t, A_t)|^{2r}\leq \|v_n^*\|^{2r}$.
		Hence
	$$
	\mathbb{E }| \rho(Y_{t+1}, Q^{\pi})|^{2+\zeta}\left| \frac{d
			m(X_t, Q^{\pi})}{d h}[u_n] \right |^{2+\zeta}+\mathbb{E }| \rho(Y_{t+1}
	,  Q^{\pi})|^{2+\zeta}	\leq C(\mathbb E a^2)^{1/2} \left[ d^{1/2}+  d_{\pi}^{1/2}+1\right]<C.
			$$

	      Conditions (ii)(iii)(iv) are trivially satisfied because of the linearity.
	\\
	For condition (v), let $T_2=\max_{j\leq k_n}\Psi_j(X_t)^2+1.$ Recall that for $h\in\mathcal C_n$, $h= h_1+ xu_n$ where $\|h_1-Q^{\pi}\|_{\infty,\omega}<C\delta_n$ and $|x|\leq Cn^{-1/2}$. Hence 
	\begin{eqnarray*}
&&\mathbb{E}T_2\sup_{h\in\mathcal{C}%
			_n} (\rho(Y_{t+1}, h)-\rho(Y_{t+1}, Q^{\pi}) )^2\leq C \mathbb{E}T_2\sup_{h\in\mathcal{C}%
			_n}|h(X_t)- Q^{\pi}(X_t)| ^2
		+\mathbb{E}T_2\sup_{h\in\mathcal{C}%
			_n} \gamma^2 [K(h)-K(Q^\pi)]^2\cr
			&\leq& \left[ C \mathbb{E}T_2 (1+\|X_t\|^2)^{\omega}  
			+\gamma^2 \mathbb{E}T_2 M(S_{t+1})^2\right]\sup_{h\in\mathcal{C}%
			_n}\|h- Q^\pi\|_{\infty,\omega}^2			\cr
			&\leq& O(\delta_n^2+n^{-1}) \leq C\delta_n^2.
	\end{eqnarray*}
	   
	\end{proof}
	 }
	
	\subsection{{\protect\small NPIV model: proof of Proposition \protect\ref%
			{prop5.1}}}
	
	{\small In this case $m(X_t,h)=\mathbb{E}(h_0(W_t)-h(W_t)|\sigma_t(\mathcal{X%
		}))$ and $\epsilon(S_t,\alpha)=U_t+h_0(W_t)-h(W_t)-\mathbb{E}%
		(h_0(W_t)-h(W_t)|\sigma_t(\mathcal{X}))$, $\frac{dm(X_t, \alpha)}{dh}[v] =
		\mathbb{E}(v(W_t)|\sigma_t(\mathcal{X})),$ and $\frac{d^2}{d\tau^2} m(X_t,
		h+ \tau v)=0$ because of the linearity. }
	
	{\small
		\begin{proof}
			We sequentially verify conditions in Assumptions \ref{ass2.1}, \ref{ass3.1}, \ref{ass4.2} and  \ref{a5.555}.

			\textit{Verifying Assumption \ref{ass2.1}  } This assumption follows immediately from
			$$
			\|\alpha_1-\alpha_2\|^2=\mathbb E\left( \mathbb E(h_1(W_t)-h_2(W_t)|\sigma_t(\mathcal X))\right)^2\Sigma(X_t)^{-1} =\mathbb E[ m(X_t, h_1)- m(X_t, h_2)]^2\Sigma(X_t)^{-1} .
			$$

			\textit{Verifying Assumption \ref{ass3.1} }  (i)  Uniformly  in $j\leq k_n$, for $M_t:=(1+|W_t|^2)^{\omega}$,  
			\begin{eqnarray*}
				&& \mathbb E[\Psi_j(X_t)^2+1] \sup_{ \|\alpha-\alpha_1\|_{\infty,\omega}<\delta}   |\epsilon(S_t,\alpha_1)-\epsilon(S_t,\alpha)|^2\cr
				&\leq&   \mathbb E[\Psi_j(X_t)^2+1] \sup_{ \|h-h_1\|_{\infty,\omega}<\delta}   [h_1(W_t)-h(W_t)-\mathbb E(h_1(W_t)-h(W_t)|\sigma_t(\mathcal X)) ]^2\cr
				&\leq&  4 \mathbb E[\Psi_j(X_t)^2+1] \left[ M_t+ \mathbb E(M_t|\sigma_t(\mathcal X)) \right]\sup_{ \|h-h_1\|_{\infty,\omega}<\delta}  \|h_1-h\|_{\infty,\omega}^2\leq C\delta^2
			\end{eqnarray*}
			given that $\mathbb E[\Psi_j(X_t)^2+1] M_t<\infty$.
			
			
		(ii) Suppose $\mathbb E\max_{j\leq k_n}\Psi_j(X_t)^2[(1+|W_t|^2)^{\omega}+U_t^2]<\infty$,
			\begin{eqnarray*}
				&&\mathbb E\max_{j\leq k_n}\Psi_j(X_t)^2 \sup_{\alpha\in\mathcal A_n}  \rho(Y_{t+1}, \alpha   )^2\cr
				&\leq&
				2 \mathbb E\max_{j\leq k_n}\Psi_j(X_t)^2 \sup_{\alpha\in\mathcal A_n}  [h_0(W_t)- h(W_t)]^2
				+2 \mathbb E\max_{j\leq k_n}\Psi_j(X_t)^2    U_t^2\leq C.
			\end{eqnarray*}
			
			(iii) We have
			$
			\|\alpha_1-\alpha_2\|^2\leq C\|\alpha_1-\alpha_2\|_{\infty,\omega}^2\mathbb E(1+|W_t|^2)^{\omega}.
			$
			
			\textit{Verifying Assumption \ref{ass4.2} }
			
			For (i)  we have
			$$
			\mathbb E | \rho(Y_{t+1}, \alpha_0)|^{2+\zeta}\left| \frac{d  m(X_t,\alpha_0)}{d\alpha}[u_n]   \right |^{2+\zeta}=  \mathbb E |  U_t|^{2+\zeta}|\mathbb E(v_n^*(W_t)|X_t)|^{2+\zeta} \|v_n^*\|^{-{(2+\zeta)}}<C .
			$$

			For (ii)(iii), we have  $\frac{d^2}{d\tau^2}  m(X_t,  h+ \tau v)=0$ for any $h$ and $v$ inside $\mathcal H_0\cup\mathcal H_n$ because of the linearity.
			For (iv), we have
			$
			\sup_{\alpha\in\mathcal C_n}\frac{1}{n}\sum_t[ \frac{dm(X_t,\alpha )}{d\alpha}[u_n]- \frac{dm(X_t,\alpha_0)}{d\alpha}[u_n]]^2 =0.
			$
			
			For (v),  let $A:=\max_{j\leq k_n}\Psi_j(X_t)^2+1$.  For  $h\in\mathcal C_n$, we know there is $h_n\in\mathcal H_n$ and $|x|\leq Cn^{-1/2}$ so that
			$$
			h= h_n+ xu_n,\quad \|h_n-h_0\|_{\infty,\omega} \leq C\delta_n.
			$$ Because $\mathbb E Au_n(W)^2<C$, hence
			\begin{eqnarray*}
				&&\mathbb E A \sup_{\alpha\in\mathcal C_n } (\rho(Y_{t+1}, h )-\rho(Y_{t+1}, \alpha_0)   )^2= \mathbb EA  \sup_{\alpha\in\mathcal C_n } (h(W_t)-h_0(W_t)   )^2 \cr
				&\leq&2\mathbb E A\sup_{\mathcal C_n} |h_n(W_t)- h_0(W_t)|^2
				+Cn^{-1}\mathbb E A u_n(W)^2\cr
				&\leq& C\delta_n^2+ Cn^{-1}\leq C\delta_n^2.
			\end{eqnarray*}

			\textit{Verifying Assumption \ref{a5.555}.}
			For notational simplicity, write $\Gamma_t=\Gamma(X_t)$, $\Sigma_t=\Sigma(X_t)$, $\widehat\Sigma_t=\widehat\Sigma(X_t)$ and $\rho_t=\rho(Y_{t+1},\alpha_0)$. Using $\widehat\Sigma_t^{-1}-\Sigma_t^{-1}=\widehat\Sigma_t^{-1}(\Sigma_t-\widehat\Sigma_t)\Sigma_t^{-1}$, the triangular inequality yields
			\begin{eqnarray*}
				&&\frac{1}{n}\sum_t \Gamma_t\Sigma_t (\widehat\Sigma_t^{-1}-\Sigma_t^{-1})\rho_t   \leq
				|\frac{1}{n}\sum_t \Gamma_t\Sigma_t (\widehat\Sigma_t^{-1}-\Sigma_t^{-1})  (\widehat\Sigma_t -\Sigma_t )\Sigma_t^{-1}\rho_t |\cr
				&&
				+|\frac{1}{n}\sum_t \Gamma_t   (\widehat\Sigma_t -\Sigma_t )\Sigma_t^{-1}\rho_t |\cr
				&\leq&|\frac{1}{n}\sum_t \Gamma_t   (\widehat\Sigma_t -\Sigma_t )\Sigma_t^{-1}\rho_t | +O_P(1)\frac{1}{n}\sum_t|(\widehat\Sigma_t-\Sigma_t)|^2 .
			\end{eqnarray*}
			Note  $\widehat\Sigma_t = \widehat A_n'\Psi_n(\Psi_n'\Psi_n)^{-1}\Psi(X_t)$ where $\widehat A_n$ is a $n\times 1$ vector of $\widehat\rho_t^2$.
			Also let $(A_n, \mathbb E(A_n|X), G_n, U_n) $ respectively be $n\times 1$ vectors of $(\rho_t^2, \Sigma_t, g_t, u_t)$
			where $g_t=\Gamma_t\Sigma_t^{-1}\rho_t$ and $u_t=\rho_t^2-\mathbb E(\rho_t^2|\sigma_t(\mathcal X))$.
			Let $J_t$ be  the $t$ th element of $(I-P_n)\mathbb E(A_n|X)$.

			We have $(\frac{1}{\sqrt{n}} \|\widehat A_n-A_n\| )^2 \leq C\frac{1}{n}\sum_t(\widehat \rho_t-\rho_t)^2\rho_t^2+ C\frac{1}{n}\sum_t(\widehat \rho_t-\rho_t)^4 =O_P(\delta_n^2)$. In addition, let $D$ be the diagonal matrix of $\Gamma_t\Sigma_t^{-1}$. Then $\frac{1}{\sqrt{n}} \|P_nG_n\|
			=O_P(\frac{1}{\sqrt{n}}\sqrt{\rho_n'DP_nD\rho_n}) =O_P(\sqrt{k_n/n}).
			$
			So we have the following decomposition
			\begin{eqnarray*}
				&&\frac{1}{n}\sum_t \Gamma_t   (\widehat\Sigma_t -\Sigma_t )\Sigma_t^{-1}\rho_t
				=\frac{1}{n}[\widehat A_n' P_n -\mathbb E(A_n|X)]G_n=a_1+a_2+a_3,\cr
				&&\frac{1}{n}\sum_t|(\widehat\Sigma_t-\Sigma_t)|^2=\frac{1}{n}\|P_n\widehat A_n-\mathbb E(A_n|X)\|^2\leq C(a_4+a_5+a_6)
				\cr
				a_1&=&\frac{1}{n}\mathbb E(A_n|X)' (P_n-I)  G_n =\frac{1}{n}\sum_{t} J_t\Gamma_t\Sigma_t^{-1}\rho_t=O_P(\frac{1}{\sqrt{n}})\sqrt{\mathbb EJ_t^2\Gamma_t^2\Sigma_t^{-1} }
				\cr
				&=&O_P(n^{-1/2}\sqrt{\mathbb EJ_t^2 }) =O_P(\sqrt{\varphi_n^2/n}). \cr
				a_2&=&\frac{1}{n}U_n' P_n  G_n \leq O_P(1 )\|\frac{1}{n}U_n' \Psi_n\| \|\frac{1}{n}\sum_t\Psi(X_t) \Gamma_t\Sigma_t^{-1}\rho_t\| =O_P(\frac{k_n}{n}) \cr
				a_3&=&  \frac{1}{n}[\widehat A_n-A_n]' P_n  G_n
				\leq\frac{1}{\sqrt{n}} \|\widehat A_n-A_n\| \frac{1}{\sqrt{n}} \|P_nG_n\|
				\leq O_P(\delta_n^2+ {\frac{k_n}{n}})=O_P(\delta_n^2)
				\cr
				a_4&=&\frac{1}{n}\| (I-P_n)\mathbb E(A_n|X)\|^2 = O_P(\varphi_n^2). \cr
				a_5&=&\frac{1}{n}\|  P_n U_n\|^2\leq O_P(1)\|\frac{1}{n}U_n' \Psi_n\|^2= O_P(\frac{k_n}{n}) \cr
				a_6&=&\frac{1}{n}\|  P_n (\widehat A_n-A_n)\|^2
				=O_P(\delta_n^2)
			\end{eqnarray*}
			Putting together,
			$\frac{1}{n}\sum_t \Gamma_t\Sigma_t (\widehat\Sigma_t^{-1}-\Sigma_t^{-1})\rho_t =  O_P(p_n) $
			where
			$
			p_n=\varphi_n^2+  \frac{k_n}{n}
			+  \delta_n^2\leq C\delta_n^2 . 
			$
			
		\end{proof}
	}
	
	\subsection{{\protect\small NPQIV model: proof of Proposition \protect\ref%
			{prop5.2}}}
	
	{\small
	}
	
	{\small
	}
	
	{\small
	}
	
	{\small In this model $m(X_t, \alpha)= P(U_t<h-h_0|\sigma_t(\mathcal{X}%
		))-\varpi$ where $U_t=Y_t- h_0(W_t)$. Suppose the conditional distribution
		of $U_t$ given $(X_t, W_t)$ is absolutely continuous with density function $%
		f_{U_t|\sigma_t(\mathcal{X}), W_t}(u)$. Then the derivative is defined as
		\begin{equation*}
		\frac{dm(X_t, \alpha)}{dh}[v] = \mathbb{E}(f_{U_t|\sigma_t(\mathcal{X}),
			W_t}( h(W_t)-h_0(W_t) )v(W_t)|\sigma_t(\mathcal{X})).
		\end{equation*}
	}
	
	{\small
		\begin{proof}
			
			\textit{Verifying Assumption \ref{ass2.1}. }
			Let
			\begin{eqnarray*}
				A_t( h)&:=&  \int_0^1 f_{U_t|\sigma_t(\mathcal X), W_t}\left(   x(h(W_t)- h_0(W_t)) \right)dx\cr
				B_t(v,h)&:=&\mathbb E  \left\{ A_t( v) [ h(W_t)- h_0(W_t)] |\sigma_t(\mathcal X)\right\}.
			\end{eqnarray*}
			Then
			$m(X_t, h)= B_t(h,h)$, $\mathbb Em(X_t, h)^2 \Sigma(X_t)^{-1}=\mathbb EB_t(h,h)^2\Sigma(X_t)^{-1}$ and $  \|\alpha-\alpha_0\|^2=\mathbb E  B_t(h_0,h)^2   \Sigma(X_t)^{-1}.$
			This assumption then follows from the condition that  $c_2\mathbb E  B_t(h ,h)^2   \Sigma(X_t)^{-1}\leq \mathbb E  B_t(h_0,h)^2   \Sigma(X_t)^{-1}\leq c_1 \mathbb E  B_t(h ,h)^2   \Sigma(X_t)^{-1}$ for all $\|h-h_0\|<\epsilon_0$.

			\textit{Verifying Assumption \ref{ass3.1}  }  (i) Let $A_j:=[\Psi_j(X_t)^2+1]$. Fix any $\alpha=h\in\mathcal A_n,$  
			\begin{eqnarray*}
				&& \mathbb E  A_j   \sup_{ \|\alpha-\alpha_1\|_{\infty,\omega}<\delta}   |\epsilon(S_t,\alpha_1)-\epsilon(S_t,\alpha)|^2\cr
				&\leq&
				2 \mathbb E  A_j   \sup_{ \|\alpha-\alpha_1\|_{\infty,\omega}<\delta}   |\rho(Y_{t+1},\alpha_1)-\rho(Y_{t+1},\alpha)|^2
				+2 \mathbb E  A_j   \sup_{ \|\alpha-\alpha_1\|_{\infty,\omega}<\delta}   |m(X_{t},\alpha_1)-m(X_{t},\alpha)|^2.      \end{eqnarray*}
			On one hand,
			\begin{eqnarray*}
				&&  \mathbb E  A_j   \sup_{ \|\alpha-\alpha_1\|_{\infty,\omega}<\delta}   |m(X_{t},\alpha_1)-m(_{t},\alpha)|^2\cr
				&\leq&
				2   \mathbb E  A_j   \sup_{ \|h-h_1\|_{\infty,\omega}<\delta}     P(h(W_t)- h_0(W_t)\leq U_t\leq h_1(W_t)- h_0(W_t)|X ) ^21\{h_1(W_t)>h(W_t)\}\cr
				&&+ 2   \mathbb E  A_j   \sup_{ \|h-h_1\|_{\infty,\omega}<\delta}    P(h_1(W_t)- h_0(W_t)\leq U_t\leq h(W_t)- h_0(W_t)|X ) ^21\{h(W_t)>h_1(W_t)\}\cr
				&\leq&   2   \mathbb E  A_j      \sup_uf_{U_t|\sigma_t(\mathcal X), W_t}(u)    ^2(1+|W_t|^2)^{\omega} \sup_{ \|h-h_1\|_{\infty,\omega}<\delta} \|h_1- h_2\|_{\infty,\omega} ^2\cr
				&\leq&  2\mathbb E  A_j \sup_uf_{U_t|\sigma_t(\mathcal X), W_t}(u) ^2(1+|W_t|^2)^{\omega}\delta^2 \leq C \delta^2.
			\end{eqnarray*}
			On the other hand,   for notational simplicity, write $a=h(W_t)- h_0(W_t)$, and $a_1=h_1(W_t)- h_0(W_t)$. Then $\|h-h_1\|_{\infty,\omega}<\delta$ implies $|a-a_1|\leq \delta(1+|W_t|^2)^{\omega/2}:=g_t(\delta)$. So 
			\begin{eqnarray*}
				&&   \mathbb E  A_j   \sup_{ \|\alpha-\alpha_1\|_{\infty,\omega}<\delta}   |\rho(Y_{t+1},\alpha_1)-\rho(Y_{t+1},\alpha)|^2
				\cr
				&\leq&
				\mathbb E  A_j   \sup_{ \|h-h_1\|_{\infty,\omega}<\delta}   1\{a\leq U_t\leq a_1\}1\{a_1>a\}
				+ \mathbb E  A_j   \sup_{ \|h-h_1\|_{\infty,\omega}<\delta} 1\{a_1\leq U_t\leq a\}1\{a>a_1\}\cr
				&\leq&
				\mathbb E A_j   \int    \sup_{h_1:  \|h-h_1\|_{\infty,\omega}<\delta}   1\{a\leq U_t\leq a_1\} f_{U_t|\sigma_t(\mathcal X), W_t}(u) du  1\{a_1>a\}
				\cr
				&& + \mathbb E  A_j   \int   \sup_{h_1:  \|h-h_1\|_{\infty,\omega}<\delta} 1\{a_1\leq U_t\leq a\}f_{U_t|\sigma_t(\mathcal X), W_t}(u)  du1\{a>a_1\}\cr
				&\leq&
				\mathbb E A_j   \int _a^{a+g_t(\delta)}    f_{U_t|\sigma_t(\mathcal X), W_t}(u) du  1\{a_1>a\}
				+ \mathbb E  A_j   \int  _{a-g_t(\delta)}^a  f_{U_t|\sigma_t(\mathcal X), W_t}(u)  du1\{a>a_1\}\cr
				&\leq&2 \sup_uf_{U_t|\sigma_t(\mathcal X), W_t}(u) \delta \mathbb EA_j(1+|W_t|^2)^{\omega/2}\leq C\delta.
			\end{eqnarray*}

	 (ii) We have
			$$
			\mathbb E\max_{j\leq k_n}\Psi_j(X_t)^2 \sup_{\alpha\in\mathcal A_n}  \rho(Y_{t+1}, \alpha   )^2\leq C  \mathbb E\max_{j\leq k_n}\Psi_j(X_t)^2 <C.
			$$
	(iii) Because 
	$	B_t(h_0,h)^2\leq \mathbb E  \left\{ A_t( h_0)^2 (1+W_t^2)^{\omega} |\sigma_t(\mathcal X)\right\}\|h-h_0\|_{\infty,\omega}^2,$ we have 
	$$
	\|h-h_0\|^2\leq\mathbb EB_t(h_0,h)^2\Sigma(X_t)^{-1}\leq \|h-h_0\|^2_{\infty,\omega}\mathbb EA_t( h_0)^2 (1+W_t^2)^{\omega}\Sigma(X_t)^{-1}.
	$$

			\textit{Verifying Assumption \ref{ass4.2} (i).}
			Trivially $  |\rho(y, h)|+|m(x, h)|\leq 4.$ Also
			$$
			\frac{dm(X_t, \alpha)}{dh}[u_n] = \mathbb E(f_{U_t|\sigma_t(\mathcal X), W_t}(0)u_n(W_t)|\sigma_t(\mathcal X))<C.
			$$
			So $\mathbb E | \rho(Y_{t+1}, \alpha_0)|^{2+\zeta}\left| \frac{d  m(X_t,\alpha_0)}{d\alpha}[u_n]   \right |^{2+\zeta}+\mathbb E | \rho(Y_{t+1}, \alpha_0)|^{2+\zeta}<C.$

			\textit{Verifying Assumption \ref{ass4.2} (ii).}   Let $f'_{U_t|\sigma_t(\mathcal X),W_t}$ denote the first derivative of $f_{U_t|\sigma_t(\mathcal X),W_t}$. We have
			$$
			\frac{d^2}{d\tau^2}  m(X_t, h+ \tau v)=\mathbb E[f'_{U_t|\sigma_t(\mathcal X),W_t} (h(W_t)-h_0(W_t)+\tau v(W_t)) v(W_t)^2|\sigma_t(\mathcal X)]
			$$
			Hence
			\begin{eqnarray*}
				&&\mathbb E\sup_{\alpha\in\mathcal C_n}\sup_{|\tau|\leq Cn^{-1/2}}\frac{1}{n}\sum_t\left[\frac{d^2}{d\tau^2}  m(X_t, \alpha+ \tau u_n)| \right]^2 \cr
				&\leq& \mathbb E\sup_{\alpha\in\mathcal C_n}\sup_x\sup_{|\tau|\leq Cn^{-1/2}} \mathbb E  \left[ f^{'2}_{U_t|\sigma_t(\mathcal X),W_t} (h(W_t)-h_0(W_t)+\tau v(W_t)) u_n(W_t)^4|\sigma_t(\mathcal X)=x  \right] \cr
				&\leq&\sup_{u, x,w} f^{'2}_{U_t,x,w} (u)  \mathbb E [u_n(W_t)^4|\sigma_t(\mathcal X)]<C.
			\end{eqnarray*}

			\textit{Verifying Assumption \ref{ass4.2} (iii).}
			\begin{eqnarray*}
				&&        \sup_{\tau\in(0,1)}\sup_{\alpha\in\mathcal C_n} \mathbb E  \left[ \frac{d ^2 }{d\tau^2} m(X_t,\alpha_0+\tau(\alpha-\alpha_0)) \right]^2 \cr
				&\leq&        \sup_{\tau\in(0,1)}\sup_{\alpha\in\mathcal C_n} \mathbb E  \left[    \mathbb E[f'_{U_t|\sigma_t(\mathcal X),W_t} ( \tau  (h-h_0))  (h-h_0)^2|\sigma_t(\mathcal X)]       \right]^2 \leq       \sup_{\alpha\in\mathcal C_n} \mathbb E  \left[    \mathbb E  (h-h_0)^2|\sigma_t(\mathcal X)       \right]^2  \cr
				&\leq&  \sup_{h\in\mathcal C_n}\sup_w|h(w)-h(w)|^4 \leq O(\delta_n^4)=o(n^{-1}).
			\end{eqnarray*}

			\textit{Verifying Assumption \ref{ass4.2} (iv).}      Let $g_1:=h(W_t)-h_0(W_t).$
			\begin{eqnarray*}
				&&                              k_n             \sup_{\alpha\in\mathcal C_n}\frac{1}{n}\sum_t[ \frac{dm(X_t,\alpha )}{d\alpha}[u_n]- \frac{dm(X_t,\alpha_0)}{d\alpha}[u_n]]^2\cr
				&    \leq   &     k_n         \sup_{\alpha\in\mathcal C_n}\frac{1}{n}\sum_t  [\mathbb E(f_{U_t|\sigma_t(\mathcal X), W_t}(g_1)-f_{U_t|\sigma_t(\mathcal X), W_t}( 0))u_n(W_t)|\sigma_t(\mathcal X)]  ^2\cr
				&    \leq   &    k_n        L  \sup_{\alpha\in\mathcal C_n}\frac{1}{n}\sum_t \mathbb E (g_1^2|\sigma_t(\mathcal X))   \mathbb E (u_n(W_t) ^2 |\sigma_t(\mathcal X))
				\cr
				&  \leq &  C      k_n       \sup_{\alpha\in\mathcal C_n}\frac{1}{n}\sum_t (\mathbb E   (h(W_t)-h_0(W_t))^2|\sigma_t(\mathcal X))
				=O(k_n\delta_n^2)
				= o_P( 1).
			\end{eqnarray*}

			\textit{Verifying Assumption \ref{ass4.2} (v).}     Let $A=\max_{j\leq k_n}\Psi_j(X_t)^2+1.$
			\begin{eqnarray*}
				&&  \mathbb E A \sup_{  h\in\mathcal C_n    } (\rho(Y_{t+1}, h)-\rho(Y_{t+1}, h_0)   )^2
				\cr
				&    \leq&   \mathbb E A \sup_{h\in\mathcal C_n }  1\{- |h-h_0|<U_t<|h-h_0|\} \leq   \mathbb E A   1\{- \sup_{h\in\mathcal C_n } |h-h_0|<U_t<\sup_{h\in\mathcal C_n }|h-h_0|\}\cr
				&=&  \mathbb E A  \int_{- \sup_{h\in\mathcal C_n } |h-h_0|} ^{ \sup_{h\in\mathcal C_n } |h-h_0|} f_{U_t|\sigma_t(\mathcal X), W_t}(u) du\cr
				&\leq& 2 \mathbb EA \sup_{ u} f_{u|\sigma_t(\mathcal X), W_t} (u) \sup_{\mathcal C_n }|h(W_t)- h_0(W_t) |\leq    O(\delta_n )\mathbb E A(1+W_t)^{\omega/2}.
			\end{eqnarray*}

			Finally, Assumption \ref{a5.555} is naturally satisfied in the NPQIV model  where $\widehat\Sigma(X_t)=\Sigma(X_t)=\varpi(1-\varpi)$.

		\end{proof}
	}

	\newpage

	\bibliographystyle{ims}
	\bibliography{liaoBib_newest}

\end{document}